%% file: EMNLP/arxiv.tex
\documentclass[11pt]{article}

\usepackage[preprint]{acl}

\usepackage{times}
\usepackage{latexsym}

\usepackage[T1]{fontenc}

\usepackage[utf8]{inputenc}

\usepackage{microtype}

\usepackage{inconsolata}

\usepackage{graphicx}
\usepackage[table]{xcolor}
\newcommand{\cmark}{\textcolor[RGB]{45,110,20}{\ding{51}}}
\newcommand{\xmark}{\textcolor[RGB]{200,0,0}{\ding{55}}}
\usepackage{enumitem}
\usepackage{subcaption}
\usepackage{graphicx}
\usepackage{multirow}
\usepackage{float}
\usepackage{tabularx}
\usepackage{booktabs}
\usepackage{fontawesome5}
\usepackage{graphicx}

\newcommand{\projectlogo}{%
  \raisebox{-0.15em}{%
    \includegraphics[height=1.1em]{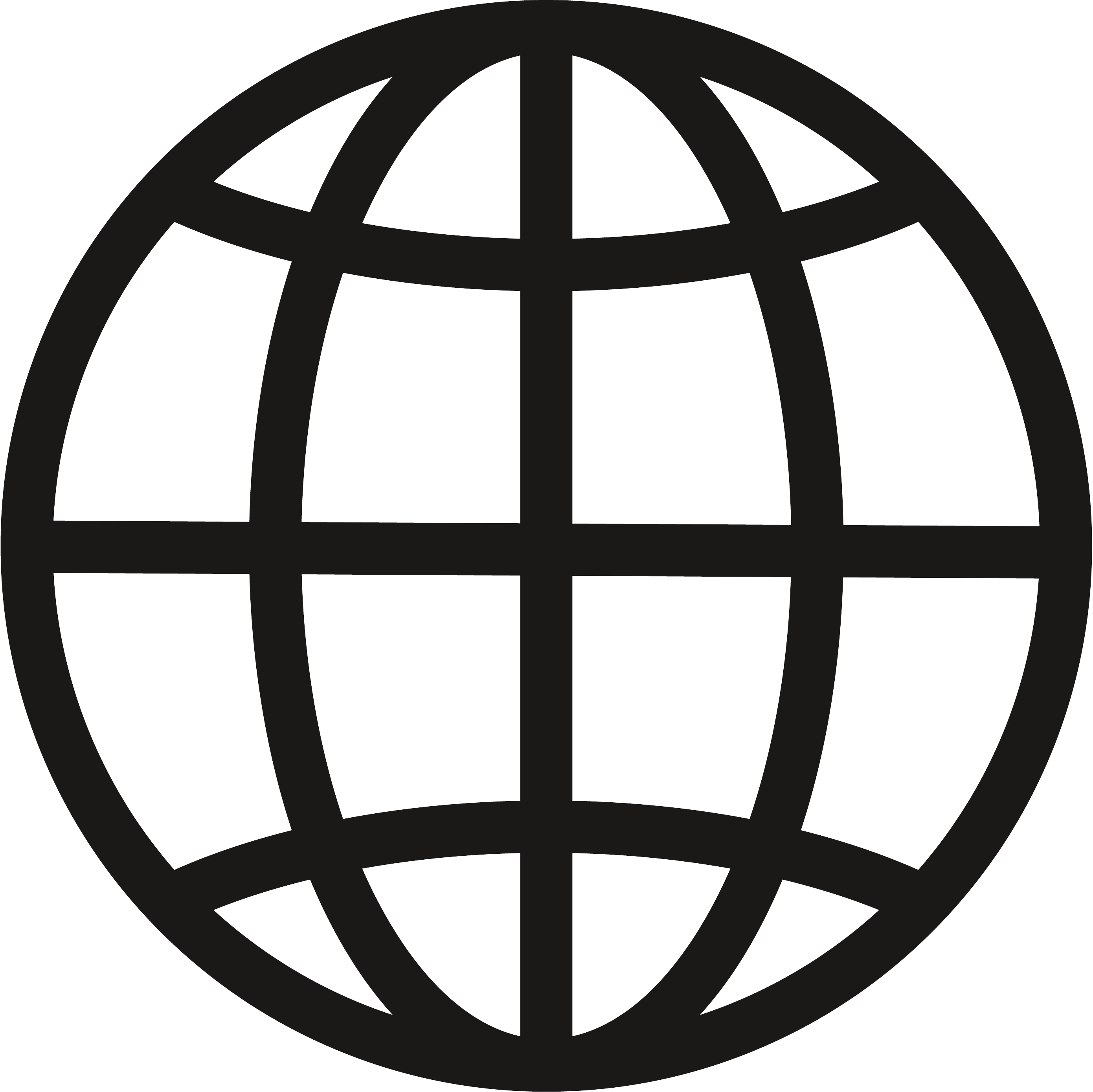}%
  }%
}

\newcommand{\codelogo}{%
  \raisebox{-0.15em}{%
    \includegraphics[height=1.1em]{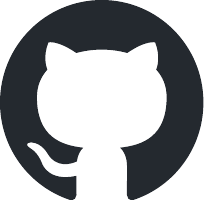}%
  }%
}

\newcommand{\hflogo}{%
  \raisebox{-0.15em}{%
    \includegraphics[height=1.1em]{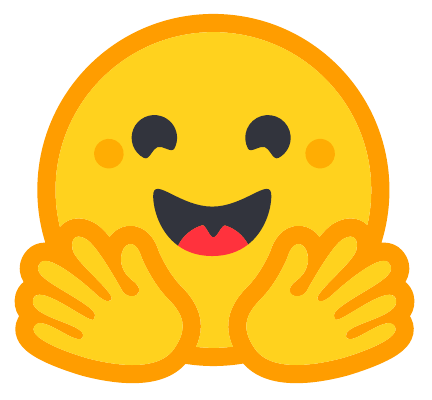}%
  }%
}
\title{RoboSPA: Can VLA Models Go Beyond Simple Scenes and Short-Horizon Tasks?}

\author{
  \textbf{Zhenxuan Fan\textsuperscript{1}},
  \textbf{Bo Zhang\textsuperscript{2}},
  \textbf{Yutong Lin\textsuperscript{1}},
  \textbf{Yuqian Yuan\textsuperscript{1}},
  \textbf{Juekai Lin\textsuperscript{1}},
  \textbf{Liang Liang\textsuperscript{1}}
  \\
  \textbf{Zhuoyi Huang\textsuperscript{3}},
  \textbf{Wenqiao Zhang\textsuperscript{1}\thanks{Corresponding author.}},
  \textbf{Juncheng Li\textsuperscript{1}\footnotemark[1]},
  \textbf{Siliang Tang\textsuperscript{1}},
  \textbf{Jun Xiao\textsuperscript{1}},
  \textbf{Yueting Zhuang\textsuperscript{1}}
  \\[0.5em]
  \textsuperscript{1}Zhejiang University
  \\
  \textsuperscript{2}University of Electronic Science and Technology of China
  \quad
  \textsuperscript{3}South China Normal University
  \\[0.3em]
  {\small
    \texttt{zxfan@zju.edu.cn}
    \quad
    \texttt{wenqiaozhang@zju.edu.cn}
  }
  \\[0.8em]
  {\small
    \href{https://fanzhenxuan.github.io/RoboSPA}{
      \textcolor{black}{\projectlogo\ Project Page}
    }
    \hspace{1.5em}
    \href{https://github.com/fanzhenxuan/RoboSPA}{
      \textcolor{black}{\codelogo\ Code}
    }
    \hspace{1.5em}
    \href{https://huggingface.co/datasets/zxfan/RoboSPA}{
      \textcolor{black}{\hflogo\ Data}
    }
  }
}

\begin{document}
\maketitle


\input{EMNLP/sections/0_abstract}
\input{EMNLP/sections/1_introduction}
\input{EMNLP/sections/2_related_work}
\input{EMNLP/sections/3_benchmark}
\input{EMNLP/sections/4_experiment}
\input{EMNLP/sections/5_conclusion}
\input{EMNLP/sections/6_limitation}
\input{EMNLP/sections/7_ethics_statement}
\input{EMNLP/sections/9_acknowledgments}




\bibliography{main}
\input{EMNLP/sections/8_appendix}

\end{document}

%% file: EMNLP/sections/0_abstract.tex
\begin{abstract}
Vision-Language-Action (VLA) models have shown promising progress in language-conditioned robotic manipulation. However, existing datasets and benchmarks mainly evaluate task completion under predefined settings, offering limited insight into model reasoning under increasing spatial and procedural complexity. We introduce \textbf{RoboSPA} (\textbf{Robo}t \textbf{S}patial-\textbf{P}rocedural \textbf{A}ssessment), a large-scale robotic manipulation dataset and benchmark for diagnosing embodied reasoning in VLA models. \texttt{RoboSPA} focuses on two core dimensions, Fine-Grained Spatial Reasoning and Long-Horizon Procedural Planning, covering 10 task categories and 56 base tasks. Each task is instantiated across five difficulty levels, yielding 280 variants with increasing spatial ambiguity and procedural complexity. We collect 527K trajectories across multiple embodiments and diverse scenes. Beyond binary success rate, \texttt{RoboSPA} introduces diagnostic metrics for more detailed evaluation. Experiments on representative VLA models show that current systems still struggle with complex spatial relations, precise low-level execution, and memory-intensive planning. These results establish \texttt{RoboSPA} as a challenging diagnostic benchmark for developing more capable, reliable, and generalizable embodied agents. Our data and code are available at \url{https://github.com/fanzhenxuan/RoboSPA}.
\end{abstract}

%% file: EMNLP/sections/1_introduction.tex
\section{Introduction}
Vision-Language-Action (VLA) models~\citep{pmlr-v229-zitkovich23a, pmlr-v270-kim25c,BlackK-RSS-25,nvidia2025gr00tn1openfoundation, intelligence2025pi05visionlanguageactionmodelopenworld,gao2026visualthinkvlavisualintermediatereasoning} have emerged as a promising paradigm for general-purpose embodied agents, integrating visual perception, language understanding, and action generation in a unified framework~\citep{ma2024survey,kawaharazuka2025vision}. 
With large-scale pretraining and multimodal alignment, these models show encouraging performance on manipulation tasks, especially in structured, short-horizon settings~\citep{shao2025large,zhong2025survey}.

However, deploying embodied agents in real-world scenarios requires capabilities beyond short-horizon instruction following in simple, well-structured scenes~\citep{liu2025aligning,zhang2025generative,wong2025survey,yuan2026eocbench,dang2026rynnbrainopenembodiedfoundation}. Everyday manipulation tasks pose three key challenges: 
(1) \textbf{Fine-grained target disambiguation}, requiring agents to identify the target among visually similar candidates through subtle spatial cues beyond category or color;
(2) \textbf{Temporally extended task execution}, requiring multi-step execution under temporal constraints and accumulated errors;
and (3) \textbf{Complexity-scalable embodied reasoning}, where increasing candidates, horizons, and environmental diversity amplify grounding and planning failures.
These challenges call for benchmarks that evaluate embodied reasoning under increasing spatial and procedural complexity.

\input{EMNLP/tables/datasets}
As shown in Table~\ref{tab:dataset_comparison}, existing robotic manipulation datasets and benchmarks leave several gaps for evaluating reasoning-oriented VLA models. 
First, fine-grained spatial reasoning is rarely evaluated explicitly, as prior works seldom test target disambiguation with subtle spatial cues. 
Second, long-horizon and step-level evaluation remain limited: LIBERO~\citep{liu2023libero} and RoboTwin 2.0~\citep{chen2026robotwin} mainly focus on short-horizon tasks, while RoboCasa~\citep{Nasiriany-RSS-24} and MIKASA-Robo~\citep{cherepanov2026memory} often lack step-level diagnosis. 
Third, most prior works lack controlled multi-level difficulty~\citep{Pumacay-RSS-24,fei25libero-plus,zhou2025liberoprorobustfairevaluation}, making it hard to analyze performance degradation as task complexity increases. Finally, dataset scale remains limited for broad reasoning evaluation. 
Most widely used datasets contain fewer than 200K trajectories, limiting task, embodiment, and scene coverage.

\begin{figure*}[t]
    \centering
    \includegraphics[width=0.95\textwidth]{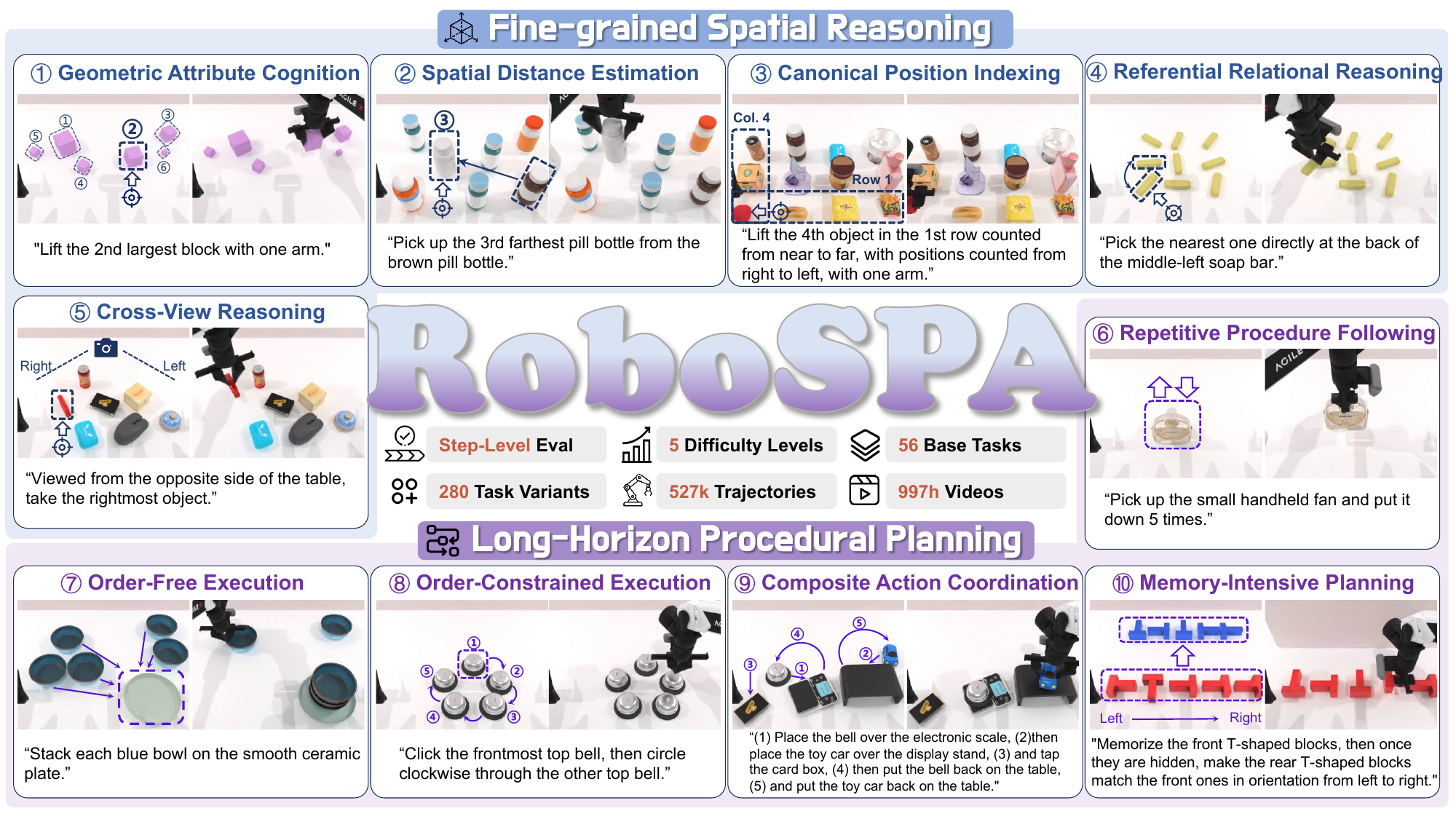}
    \caption{\textbf{Overview of \texttt{RoboSPA}}. \texttt{RoboSPA} centers on Fine-Grained Spatial Reasoning and Long-Horizon Procedural Planning, covering \textbf{10} task categories and \textbf{56} base tasks. Each task has \textbf{5} difficulty levels, yielding \textbf{280} variants. The dataset includes \textbf{527K} trajectories across multiple embodiments and scene settings, supporting step-level evaluation of VLA limitations in spatial reasoning, long-horizon execution, and scalable embodied reasoning.}
    \label{fig:introduction}
\end{figure*}

To address these limitations, we introduce \textbf{RoboSPA} (\textbf{Robo}t \textbf{S}patial-\textbf{P}rocedural \textbf{A}ssessment), a large-scale robotic manipulation dataset and benchmark. To systematically study embodied reasoning under increasing task complexity, \texttt{RoboSPA} is guided by three core design principles:

\begin{itemize}[leftmargin=12pt]

\item \textbf{Fine-Grained Spatial Reasoning.} 
As shown in Fig.~\ref{fig:introduction}, this component evaluates whether VLA models can ground instructions in complex spatial structures, such as geometric attributes, distances, cross-view cues, relations, and canonical indexing. This capability is essential for cluttered manipulation. To our knowledge, \texttt{RoboSPA} is the first VLA dataset to make fine-grained spatial reasoning a core evaluation dimension.

\item \textbf{Long-Horizon Procedural Planning.} 
This component evaluates whether VLA models can execute manipulation tasks with multi-step decisions. As shown in Fig.~\ref{fig:introduction}, it covers repetitive procedures, order-constrained and order-free execution, composite coordination, and memory-intensive planning. These capabilities are crucial for real-world manipulation requiring temporal consistency and stepwise progress.

\item \textbf{Multi-Level Hierarchical Evaluation.} 
This component measures how VLA model performance changes with increasing task complexity. Each base task is instantiated across five difficulty levels, enabling analysis of performance degradation under growing spatial complexity and action horizons. Beyond task-level success, \texttt{RoboSPA} reports step-level progress for detailed diagnosis of model failures.
\end{itemize}

Together, these designs provide a controlled foundation for large-scale VLA data collection and evaluation. \texttt{RoboSPA} provides demonstrations in clean and domain-randomized scenes, covering 56 base tasks across 10 capability categories and five difficulty levels, yielding 280 variants. Each task supports step-level evaluation beyond final success rates. Across five embodiments, \texttt{RoboSPA} contains 527K trajectories and 997 hours of videos, forming a comprehensive benchmark for embodied reasoning under increasing task complexity.

We evaluate representative VLA models~\citep{liu2025rdtb, agibotworldcontributors2025agibotworldcolosseolargescale, intelligence2025pi05visionlanguageactionmodelopenworld, zheng2026xvla} on \texttt{RoboSPA} and find that they struggle as task complexity increases. 
On the hardest tasks, all models achieve an average success rate below 25\%, with some tasks dropping to 0\%.
Diagnostic analyses reveal failures in target grounding, low-level manipulation, long-horizon tracking, and memory-based reasoning, highlighting \texttt{RoboSPA} as a challenging diagnostic testbed for embodied reasoning.

%% file: EMNLP/tables/datasets.tex
\begin{table*}[t]
\centering
\scriptsize
\setlength{\tabcolsep}{1.5pt}
\renewcommand{\arraystretch}{1.0}
\begin{tabular}{lccccccccc}
\toprule
\textbf{Name} &
\textbf{Categories} &
\textbf{Tasks} &
\textbf{Trajectories} &
\textbf{Embodiments} &
\textbf{Spatial} &
\textbf{Long-Horizon} &
\textbf{Step-level} &
\textbf{Multi-} &
\textbf{Scene} \\
& & & & &
\textbf{Reasoning} &
\textbf{Tasks} &
\textbf{Evaluation} &
\textbf{Difficulty} &
\textbf{Diversity} \\
\midrule
BC-Z~\cite{jang2022bc}                         & 3   & 100+    & 26K          & 1  & \xmark & \xmark & \xmark & \xmark & \xmark \\
RT-1~\cite{Brohan-RSS-23}                      & 8   & 700+    & 130K         & 1  & \xmark & \cmark & \xmark & \xmark & \cmark \\
BridgeData V2~\cite{pmlr-v229-walke23a}        & --  & 13      & 60.1K        & 1  & \xmark & \xmark & \xmark & \xmark & \cmark \\
Open X-Embodiment~\cite{10611477}              & 527 & 160,266 & 1.4M         & 22 & \xmark & \xmark & \xmark & \xmark & \cmark \\
DROID~\cite{khazatsky2024droid}                & --  & 86      & 76K          & 1  & \xmark & \xmark & \xmark & \xmark & \cmark \\
\midrule
RLBench~\cite{james2020rlbench} 
                                                          & --  & 100     & --           & 1  & \xmark & \cmark & \xmark & \xmark & \xmark \\
CALVIN~\cite{9788026}                           & --  & 34      & N/A$^\dagger$& 1  & \xmark & \cmark & \cmark & \xmark & \cmark \\
LIBERO~\cite{liu2023libero}                     & 4   & 130     & 6.5K         & 1  & \xmark & \xmark & \xmark & \xmark & \cmark \\
ManiSkill2~\cite{gumaniskill2}                  & 4   & 20      & N/A$^\ddagger$ & 1  & \xmark & \xmark & \xmark & \xmark & \cmark \\
RoboCasa~\cite{Nasiriany-RSS-24}                & --  & 100     & $\sim$100K   & 1  & \xmark & \cmark & \xmark & \xmark & \cmark \\
SimplerEnv~\cite{pmlr-v270-li25c}                  & --  & 8       & --           & 2  & \xmark & \xmark & \xmark & \xmark & \cmark \\
MIKASA-Robo~\cite{cherepanov2026memory}         & 12  & 32      & 32K          & 1  & \xmark & \cmark & \xmark & \cmark & \xmark \\
RoboTwin 2.0~\cite{chen2026robotwin}            & --  & 50      & $\leq$137.5K & 5  & \xmark & \xmark & \xmark & \xmark & \cmark \\
LIBERO-Pro~\cite{zhou2025liberoprorobustfairevaluation} 
                                                          & 4   & 40      & --           & 1  & \xmark & \xmark & \xmark & \xmark & \cmark \\
RMBench~\cite{chen2026rmbenchmemorydependentroboticmanipulation} 
                                                          & 2   & 9       & 450          & 1  & \xmark & \cmark & \xmark & \xmark & \xmark \\
\midrule
\textbf{\texttt{RoboSPA} (Ours)}                            & 10  & 280     & 527K         & 5  & \cmark & \cmark & \cmark & \cmark & \cmark \\
\bottomrule
\end{tabular}
\caption{
\textbf{Comparison of representative robotic manipulation datasets and benchmarks with \texttt{RoboSPA}.} The upper and lower parts summarize datasets and benchmarks, respectively. $^\dagger$CALVIN reports approximately 24 hours of teleoperated demonstrations. 
$^\ddagger$ManiSkill2 reports over 4M demonstration frames.
}
\label{tab:dataset_comparison}
\end{table*}

%% file: EMNLP/sections/2_related_work.tex
\section{Related Work}
\label{sec:related_work}
\paragraph{Vision-Language-Action Models.}
Large language models (LLMs)~\citep{grattafiori2024llama3herdmodels,yang2025qwen3technicalreport,cao2026draftthinkinglearningefficientreasoning,fan2026ctrlcotdualgranularitychainofthoughtcompression} and multimodal large language models (MLLMs)~\citep{liu2023visual,bai2025qwen3vltechnicalreport,11094265,lin2025healthgpt,DBLP:conf/aaai/WangFTZYLJZSWX26} have demonstrated strong capabilities for understanding and reasoning over language and multimodal inputs, respectively.
Building on these advances, VLA models map visual observations and language instructions to robot actions through a unified perception-to-control framework.
Early methods, such as RT-1~\citep{Brohan-RSS-23}, RT-2~\citep{pmlr-v229-zitkovich23a}, and OpenVLA~\citep{pmlr-v270-kim25c}, cast action prediction as autoregressive discrete token generation.
Subsequent VLA models largely adopt diffusion or flow-matching policies for continuous action generation, as in RDT~\citep{liu2025rdtb} and $\pi_0$~\citep{BlackK-RSS-25}.
To improve cross-task generalization, VLA models such as $\pi_{0.5}$~\citep{intelligence2025pi05visionlanguageactionmodelopenworld} and GR00T N1~\citep{nvidia2025gr00tn1openfoundation} adopt hierarchical policies to bridge high-level language understanding with low-level motor execution, while other works enhance action and embodiment generalization through implicit action modeling and cross-embodiment data~\citep{agibotworldcontributors2025agibotworldcolosseolargescale,zheng2026xvla}.
From a different perspective, another line of work improves VLA models through spatial guidance, memory mechanisms, and future-frame prediction, further enhancing their capability and usability~\citep{chen2025internvlam1spatiallyguidedvisionlanguageaction,shi2026memoryvla,torne2026memmultiscaleembodiedmemory,zhang2025dreamvla,Bi_2026_CVPR}.

\begin{figure*}[t]
\centering
\begin{minipage}{0.4\textwidth}
    \centering
    \includegraphics[width=\linewidth]{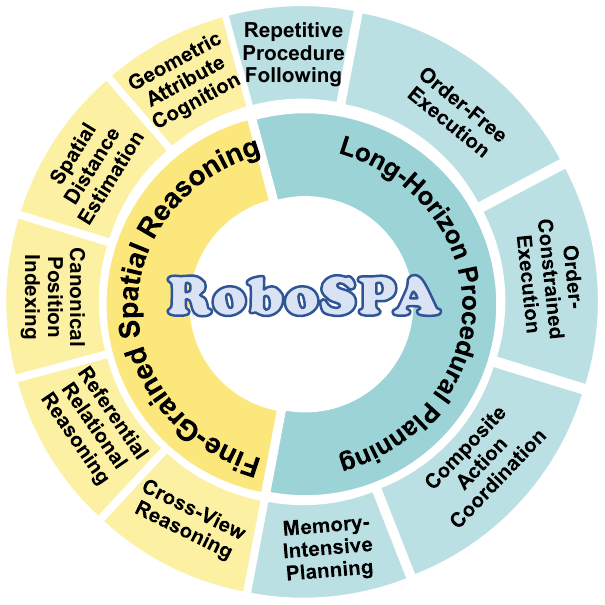}
    {\footnotesize \caption*{(a) Capability taxonomy}}
\end{minipage}
\hfill
\begin{minipage}{0.56\textwidth}
    \centering
    \includegraphics[width=\linewidth]{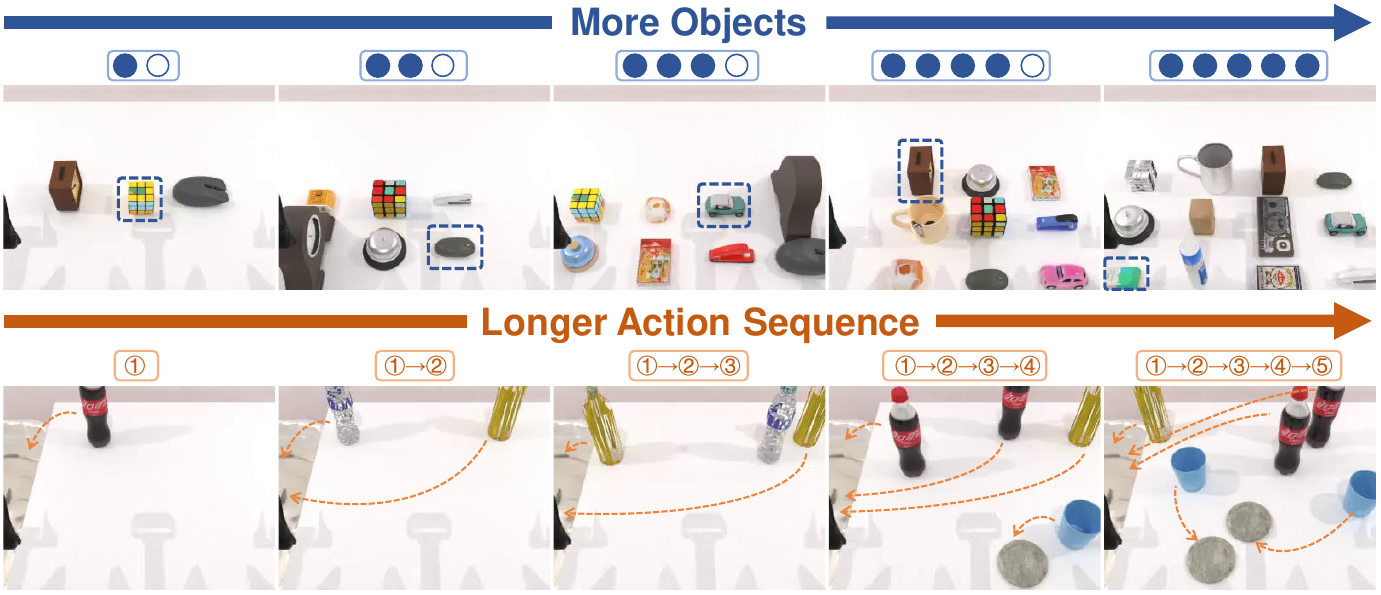}
    {\footnotesize \caption*{(b) Multi-difficulty task design}}
    
    \vspace{0.8em}
    
    \includegraphics[width=\linewidth]{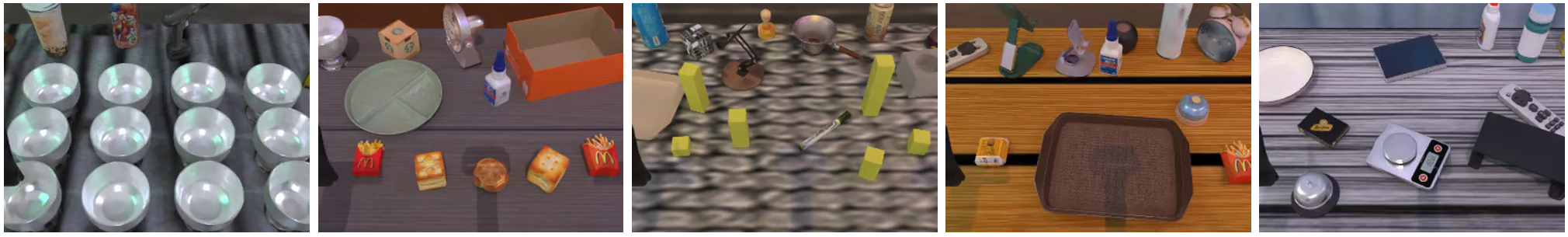}
    {\footnotesize \caption*{(c) Domain-randomized scenes}}
\end{minipage}

\caption{\textbf{Design of \texttt{RoboSPA}.} (a) Capability taxonomy with two dimensions and 10 categories. (b) Multi-difficulty task design through increasing object counts or action-sequence length. (c) Domain-randomized scenes with diverse layouts, distractors, lighting, textures, and tabletops.}
\label{fig:overview}
\end{figure*}

\paragraph{Robotic Manipulation Datasets and Benchmarks.}
The rapid development of VLA models has also accelerated the development of robotic manipulation datasets and evaluation benchmarks.
Large-scale datasets, including BridgeData V2~\citep{pmlr-v229-walke23a}, Open X-Embodiment~\citep{10611477}, DROID~\citep{khazatsky2024droid}, and AgiBot World~\citep{agibotworldcontributors2025agibotworldcolosseolargescale}, expand real-world, cross-embodiment manipulation coverage. 
Benchmark suites such as RLBench~\citep{james2020rlbench}, CALVIN~\citep{9788026}, ManiSkill2~\citep{gumaniskill2}, and LIBERO~\citep{liu2023libero} provide structured settings for evaluating diverse skills, language-conditioned control, and sequential manipulation. 
Recent benchmarks further examine scene and capability generalization, sim-to-real transfer, bimanual manipulation, and memory-dependent tasks~\citep{Pumacay-RSS-24,Nasiriany-RSS-24,pmlr-v270-li25c,sedlacek2025realm,garcia2025gembench,Zhang_2025_ICCV,zhou2025liberoprorobustfairevaluation,zhang2025vlaarena,chen2026robotwin,cherepanov2026memory,chen2026rmbenchmemorydependentroboticmanipulation}. However, they rarely systematically evaluate fine-grained spatial reasoning and long-horizon procedural planning with step-level diagnostics. \texttt{RoboSPA} addresses this gap with tasks specifically designed around these two capabilities.

%% file: EMNLP/sections/3_benchmark.tex
\section{RoboSPA}
\label{headings}
\subsection{Overview}
As shown in Fig.~\ref{fig:introduction}, we introduce \texttt{RoboSPA}, a large-scale robotic manipulation dataset and benchmark for evaluating reasoning-oriented VLA models. \texttt{RoboSPA} is built using the SAPIEN~\citep{9156706} simulator and RoboTwin 2.0~\citep{chen2026robotwin} framework. Each task is instantiated across multiple difficulty levels and scene settings.
\subsection{Benchmark Construction}
\subsubsection{Capability Taxonomy}

To systematically evaluate embodied reasoning in VLA models, we build a hierarchical capability taxonomy instead of treating manipulation tasks as isolated instances. As shown in Fig.~\ref{fig:overview}(a), it includes two core dimensions: Fine-Grained Spatial Reasoning and Long-Horizon Procedural Planning.

\paragraph{Fine-Grained Spatial Reasoning.}

This dimension assesses instruction grounding in complex spatial configurations. 
Tasks require selecting the correct target through fine-grained spatial reasoning in complex scenes. It includes five categories:

\begin{itemize}[leftmargin=*, itemsep=0.3em]
   
    \item \textbf{Geometric Attribute Cognition (GAC):} Evaluates whether the model can identify and manipulate objects by geometric or shape-related attributes beyond category-level recognition.
    
    \item \textbf{Spatial Distance Estimation (SDE):} Measures whether the model can compare distances among objects or to a reference entity, reasoning about proximity, remoteness, and relative distance.
    
    \item \textbf{Canonical Position Indexing (CPI):} Evaluates whether the model can identify objects by row-column indices under different counting directions and spatial scanning orders.

    \item \textbf{Referential Relational Reasoning (RRR):} Assesses whether the model can locate targets through directional relations to reference objects, covering basic and compositional positions.
    
    \item \textbf{Cross-View Reasoning (CVR):} Examines whether the model can interpret spatial instructions from non-egocentric viewpoints by transforming spatial references across perspectives.
    
\end{itemize}

\begin{figure*}[t]
    \centering

    \begin{subfigure}[t]{0.24\textwidth}
        \centering
        \includegraphics[width=\linewidth]{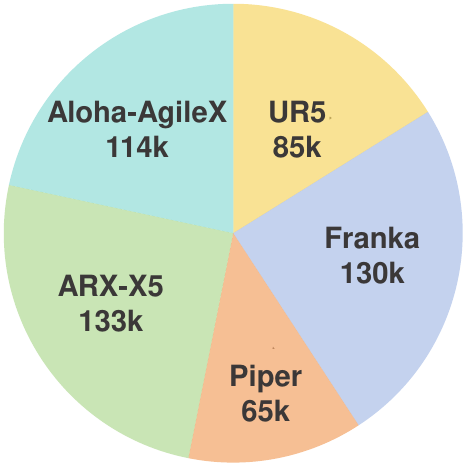}
        \caption{Embodiment Distribution}
        \label{fig:stat_a}
    \end{subfigure}
    \hfill
    \begin{subfigure}[t]{0.24\textwidth}
        \centering
        \includegraphics[width=\linewidth]{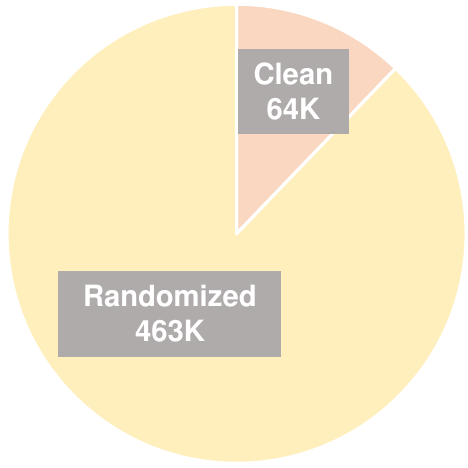}
        \caption{Scene Distribution}
        \label{fig:stat_b}
    \end{subfigure}
    \hfill
    \begin{subfigure}[t]{0.23\textwidth}
        \centering
        \includegraphics[width=\linewidth]{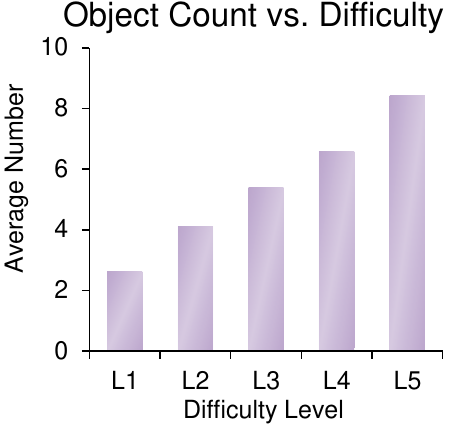}
        \caption{Average Object Count}
        \label{fig:stat_d}
    \end{subfigure}
    \hfill
    \begin{subfigure}[t]{0.24\textwidth}
        \centering
        \includegraphics[width=\linewidth]{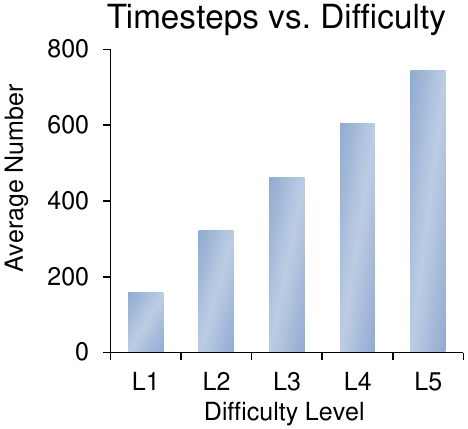}
        \caption{Average Timesteps}
        \label{fig:stat_c}
    \end{subfigure}

\caption{\textbf{Dataset statistics of \texttt{RoboSPA}.} (a) Trajectory distribution across five embodiments. (b) Trajectories across clean and domain-randomized scenes. (c) Average object count by difficulty for spatial reasoning tasks. (d) Average trajectory length by difficulty for long-horizon planning tasks.}
    \label{fig:statistics}
\end{figure*}
\paragraph{Long-Horizon Procedural Planning.}
This dimension evaluates a model's ability to complete extended manipulation tasks with diverse actions, multiple subgoals, states, and procedural constraints. It also consists of five categories:
\begin{itemize}[leftmargin=*, itemsep=0.3em]

    \item \textbf{Repetitive Procedure Following (RPF):} Assesses whether the model can repeat a specified operation the required number of times while tracking progress and stopping correctly.
    
    \item \textbf{Order-Free Execution (OFE):} Measures whether the model can complete multiple subgoals in flexible order, covering all targets without omission or unnecessary repetition.
    

    \item \textbf{Order-Constrained Execution (OCE):} Evaluates whether the model can execute subgoals in a specified order, as correct actions in the wrong order can still fail.

    \item \textbf{Composite Action Coordination (CAC):} Examines whether the model can coordinate heterogeneous manipulation skills across multi-stage procedures with different action types.
    
    \item \textbf{Memory-Intensive Planning (MIP):} Tests whether the model can retain information and use it to guide later actions when cues must be recalled or are no longer available.

\end{itemize}

\subsubsection{Task Suite Construction}

\paragraph{Task Construction.}
Based on the proposed taxonomy, we design each task to instantiate a specific reasoning requirement. Tasks are defined by expert code specifying the scene, execution procedure, and success condition. We reuse part of RoboTwin 2.0~\citep{chen2026robotwin} action primitives.
Ten trained graduate students design the tasks, with each student responsible for one capability category. 
Each task is reviewed by two additional students for quality assurance.
\paragraph{Hierarchical Difficulty Design.}

To measure model performance under increasing reasoning burden, we instantiate each task across five difficulty levels. 
As shown in Fig.~\ref{fig:overview}(b), difficulty is task-specific: spatial tasks increase candidate objects, while procedural tasks extend action sequences with more intermediate steps.
\paragraph{Instruction Design.}

For each task, we use GPT-5.2~\citep{openai2025gpt52} to generate 60 non-overlapping instruction templates with diverse linguistic forms, using 50 for training and 10 for testing. These templates are instantiated with task-specific object descriptions and manually checked to ensure diversity under controlled semantics.

\subsubsection{Data Collection}
After task design, we execute expert code in simulation to collect trajectories in both clean and domain-randomized scenes. Clean scenes contain only task-relevant objects, while domain-randomized scenes vary clutter, textures, lighting, and tabletop configurations, as shown in Fig.~\ref{fig:overview}(c), enabling evaluation of both task competence and scene robustness. We collect data across five embodiments: Aloha-AgileX, ARX-X5, Piper, Franka, and UR5.

\subsubsection{Evaluation Metrics}
To enable more diagnostic evaluation beyond \textit{Success Rate} ($\mathcal{SR}$), we further introduce two dimension-specific complementary metrics.

For fine-grained spatial reasoning tasks, we introduce \textit{Object-Normalized Target Accuracy} ($\mathcal{ONTA}$), inspired by Cohen's kappa~\citep{cohen1960coefficient}, to fairly compare spatial reasoning ability under varying scene complexity.
$\mathcal{ONTA}$ removes the object-count-induced chance baseline:
\begin{equation}
\mathcal{ONTA}(n)
=
\left(
\frac{\mathcal{SR}(n)-1/n}{1-1/n}
\right)
\times 100 .
\label{eq:onta}
\end{equation}
where \(n\) is the number of objects. 
$\mathcal{ONTA}$ is 100 for perfect target selection, 0 for random guessing, and negative for below-chance performance.

\input{EMNLP/tables/main_table}
\begin{figure*}[t]
    \centering
    \includegraphics[width=\textwidth]{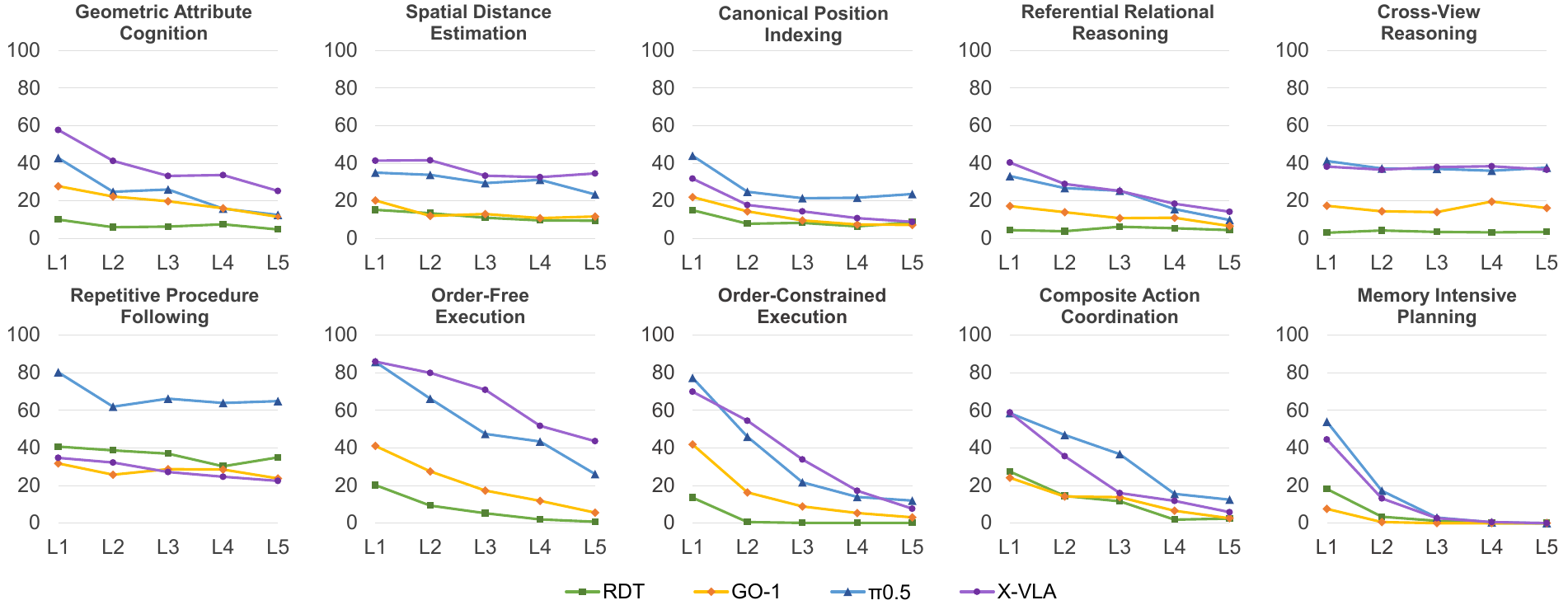}
    \caption{
    Success rate trends of four baseline VLA models across 10 task categories over five difficulty levels.}
    \label{fig:category_success_rates}
\end{figure*}

For long-horizon procedural planning tasks, we report \textit{Progress Score} ($\mathcal{PS}$) to capture partial completion. 
It is computed on a 0--100 scale:
\begin{equation}
\mathcal{PS}
=
\left(
\frac{1}{N}\sum_{i=1}^{N}\frac{c_i}{T}
\right)
\times 100 .
\label{eq:progress}
\end{equation}
where \(T\) is the total number of subtasks, and \(c_i\) is the number completed in episode \(i\).

\subsection{Benchmark Statistics}
\texttt{RoboSPA} contains \textbf{56} base tasks across two capability dimensions. Each base task is instantiated at \textbf{5} difficulty levels, yielding \textbf{280} task variants. 
In total, we collect approximately \textbf{527K} trajectories, corresponding to over \textbf{997} hours of interaction videos and \textbf{108M} timesteps.
More dataset statistics are provided in the Appendix~\ref{app:dataset_details}.
Fig.~\ref{fig:statistics}(a) presents the trajectory distribution across the \textbf{5} robotic embodiments, while Fig.~\ref{fig:statistics}(b) shows the trajectory distribution across clean and domain-randomized scenes.
Fig.~\ref{fig:statistics}(c) and Fig.~\ref{fig:statistics}(d) further illustrate the hierarchical difficulty design along the two dimensions, respectively.

%% file: EMNLP/tables/main_table.tex
\begin{table*}[t]
\centering
\caption{
\textbf{Performance comparison of four baseline VLA models on \texttt{RoboSPA}.} We use \textit{Success Rate} ($\mathcal{SR}$) as the evaluation metric. 
Drop denotes the L1-to-L5 decrease.
\colorbox{blue!15}{Blue} shading marks the best model for each metric.}

\label{tab:main_results}
\resizebox{\textwidth}{!}{
\begin{tabular}{lcccccccccccc}
\toprule
\multirow{2}{*}{Task}
& \multicolumn{3}{c}{RDT}
& \multicolumn{3}{c}{GO-1}
& \multicolumn{3}{c}{$\pi_{0.5}$}
& \multicolumn{3}{c}{X-VLA} \\
\cmidrule(lr){2-4}
\cmidrule(lr){5-7}
\cmidrule(lr){8-10}
\cmidrule(lr){11-13}
& L1 & L5 & Drop
& L1 & L5 & Drop
& L1 & L5 & Drop
& L1 & L5 & Drop \\
\midrule

\multicolumn{13}{c}{\textbf{Fine-Grained Spatial Reasoning}} \\
\midrule
Geometric Attribute Cognition 
& 10.0 & 4.8 & 5.2 
& 27.8 & 11.5 & 16.3 
& 42.8 & 12.5 & 30.3 
& \cellcolor{blue!15}57.8 & \cellcolor{blue!15}25.3 & 32.5 \\
Spatial Distance Estimation 
& 15.2 & 9.4 & 5.8 
& 20.2 & 11.6 & 8.6 
& 35.0 & 23.4 & 11.6 
& \cellcolor{blue!15}41.4 & \cellcolor{blue!15}34.6 & 6.8 \\
Canonical Position Indexing 
& 15.0 & 8.6 & 6.4 
& 22.0 & 7.2 & 14.8 
& \cellcolor{blue!15}44.0 & \cellcolor{blue!15}23.6 & 20.4 
& 31.8 & 8.8 & 23.0 \\
Referential Relational Reasoning 
& 4.4 & 4.4 & 0.0 
& 17.2 & 6.6 & 10.6 
& 33.2 & 9.8 & 23.4 
& \cellcolor{blue!15}40.4 & \cellcolor{blue!15}14.2 & 28.2 \\

Cross-View Reasoning 
& 3.0 & 3.4 & -0.4 
& 17.4 & 16.2 & 1.2 
& \cellcolor{blue!15}41.2 & \cellcolor{blue!15}37.6 & 3.6 
& 38.2 & 36.6 & 1.6 \\

\hline
\textit{Average} 
& 9.5 & 6.1 & 3.4 
& 20.9 & 10.6 & 10.3 
& 39.2 & 21.4 & 17.8 
& \cellcolor{blue!15}41.9 & \cellcolor{blue!15}23.9 & 18.0 \\

\midrule
\multicolumn{13}{c}{\textbf{Long-Horizon Procedural Planning}} \\
\midrule
Repetitive Procedure Following 
& 40.8 & 35.0 & 5.8 
& 31.8 & 23.8 & 8.0 
& \cellcolor{blue!15}80.3 & \cellcolor{blue!15}65.0 & 15.3 
& 34.8 & 22.5 & 12.3 \\
Order-Free Execution 
& 20.3 & 0.8 & 19.5 
& 41.1 & 5.6 & 35.5 
& 85.8 & 26.1 & 59.7 
& \cellcolor{blue!15}86.0 & \cellcolor{blue!15}43.6 & 42.4 \\
Order-Constrained Execution 
& 13.6 & 0.1 & 13.5 
& 41.9 & 3.1 & 38.8 
& \cellcolor{blue!15}77.4 & \cellcolor{blue!15}12.0 & 65.4 
& 70.0 & 7.7 & 62.3 \\
Composite Action Coordination 
& 27.5 & 2.4 & 25.1 
& 24.2 & 2.6 & 21.6 
& 58.6 & \cellcolor{blue!15}12.5 & 46.1 
& \cellcolor{blue!15}58.9 & 5.9 & 53.0 \\
Memory-Intensive Planning 
& 18.2 & \cellcolor{blue!15}0.0 & 18.2 
& 7.6 & \cellcolor{blue!15}0.0 & 7.6 
& \cellcolor{blue!15}54.0 & \cellcolor{blue!15}0.0 & 54.0 
& 44.6 & \cellcolor{blue!15}0.0 & 44.6 \\
\hline
\textit{Average} 
& 24.1 & 7.7 & 16.4 
& 29.3 & 7.0 & 22.3 
& \cellcolor{blue!15}71.2 & \cellcolor{blue!15}23.1 & 48.1 
& 58.8 & 15.9 & 42.9 \\

\midrule
\multicolumn{13}{c}{\textbf{Overall Benchmark}} \\
\midrule
\textit{Overall Average} 
& 16.8 & 6.9 & 9.9 
& 25.1 & 8.8 & 16.3 
& \cellcolor{blue!15}55.2 & \cellcolor{blue!15}22.3 & 32.9 
& 50.4 & 19.9 & 30.5 \\
\bottomrule
\end{tabular}
}
\end{table*}

%% file: EMNLP/sections/4_experiment.tex
\section{Experiments}
\label{others}
\input{tables/table2}
\input{tables/table3}
\subsection{Experimental Setup}
We evaluate four representative VLA models, RDT~\citep{liu2025rdtb}, GO-1~\citep{agibotworldcontributors2025agibotworldcolosseolargescale}, $\pi_{0.5}$~\citep{intelligence2025pi05visionlanguageactionmodelopenworld}, and X-VLA~\citep{zheng2026xvla}, using the clean-scene data collected with the Aloha-AgileX embodiment for both training and evaluation.
All models are trained and evaluated under a single-task setting. For each base task, we train a separate model using data from all five difficulty levels and evaluate it independently at each level.
During evaluation, each model is tested with 100 rollout trials per task variant.

\subsection{Main Results}
Table~\ref{tab:main_results} reports the \textit{Success Rates} ($\mathcal{SR}$) of four representative VLA models. 
Overall, model performance declines from L1 to L5, indicating limited ability to reason under complex task settings.
\paragraph{Model Comparison.}
Among the evaluated models, $\pi_{0.5}$ performs best overall, but reaches only 22.3\% success at L5, followed by X-VLA with 19.9\%. 
Both drop by over 30 percentage points as difficulty increases. 
GO-1 and RDT perform worse, achieving only 8.8\% and 6.9\% success at L5, respectively. 
These results show that stronger models handle easier cases better, but all models still lack robust reasoning under complex conditions.

\paragraph{Fine-Grained Spatial Reasoning.}
X-VLA shows the best performance in this dimension, but its success rate still drops from 41.9\% at L1 to only 23.9\% at L5. This decline shows that increasing spatial complexity substantially weakens target grounding. 
Models struggle particularly with Canonical Position Indexing (CPI) and Referential Relational Reasoning (RRR), indicating that current VLA models remain unreliable when target selection depends on fine-grained spatial relations rather than simple object recognition.

\paragraph{Long-Horizon Procedural Planning.}
For Long-Horizon Procedural Planning, the performance drop is even more pronounced, with the best-performing model, $\pi_{0.5}$, decreasing from 71.2\% at L1 to 23.1\% at L5.
Models perform best on Repetitive Procedure Following (RPF), while the other four categories show substantially lower performance.
Notably, all models fail on the hardest Memory-Intensive Planning (MIP), revealing severe limitations in maintaining task-relevant memory over extended horizons.

Fig.~\ref{fig:category_success_rates} further shows that performance drops from L1 to L5, exposing failures in spatial discrimination and long-horizon tracking.

\begin{figure*}[t]
    \centering
    \includegraphics[width=\textwidth]{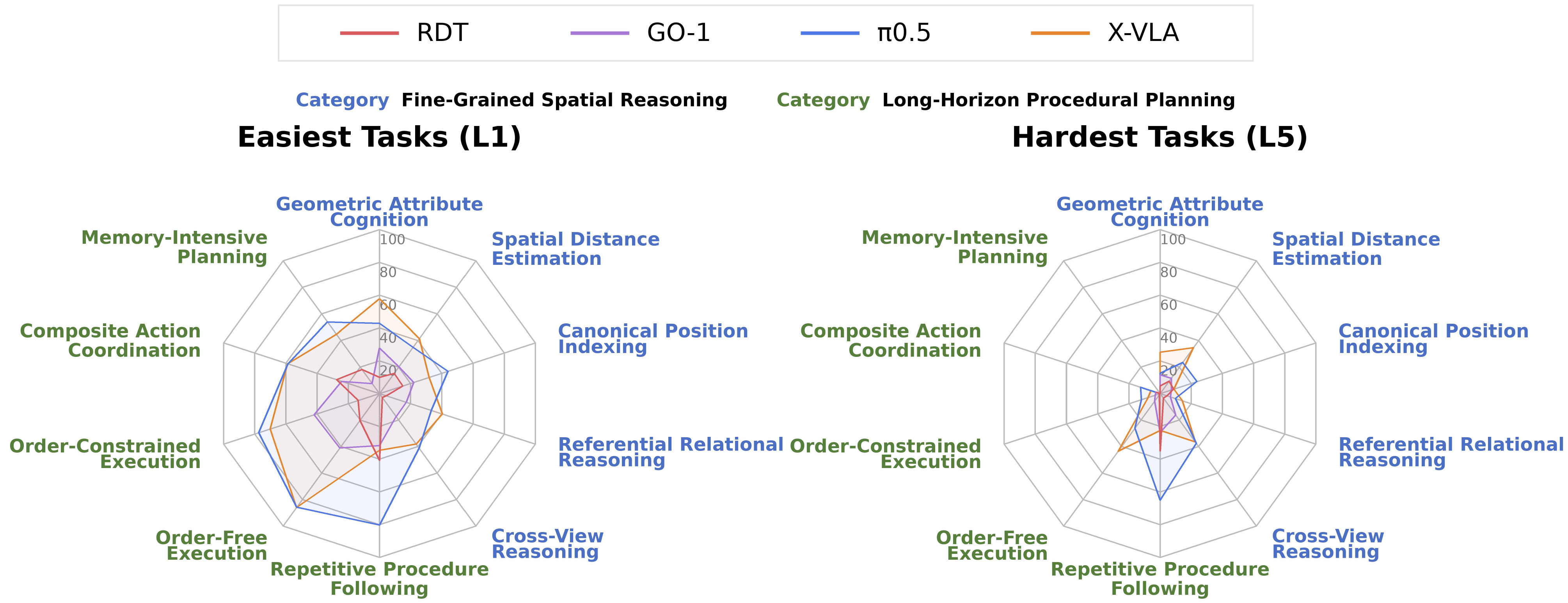}
    \caption{\textbf{Category-wise performance of four baseline VLA models on \texttt{RoboSPA}.} 
    \textbf{Left:} \textit{Success Rate} across 10 capability categories at L1. 
    \textbf{Right:} \textit{Success Rate} across 10 capability categories at L5.
}
    
    \label{fig:benchmark_radar}
\end{figure*}

\begin{figure*}[t]
    \centering
    \includegraphics[width=\textwidth]{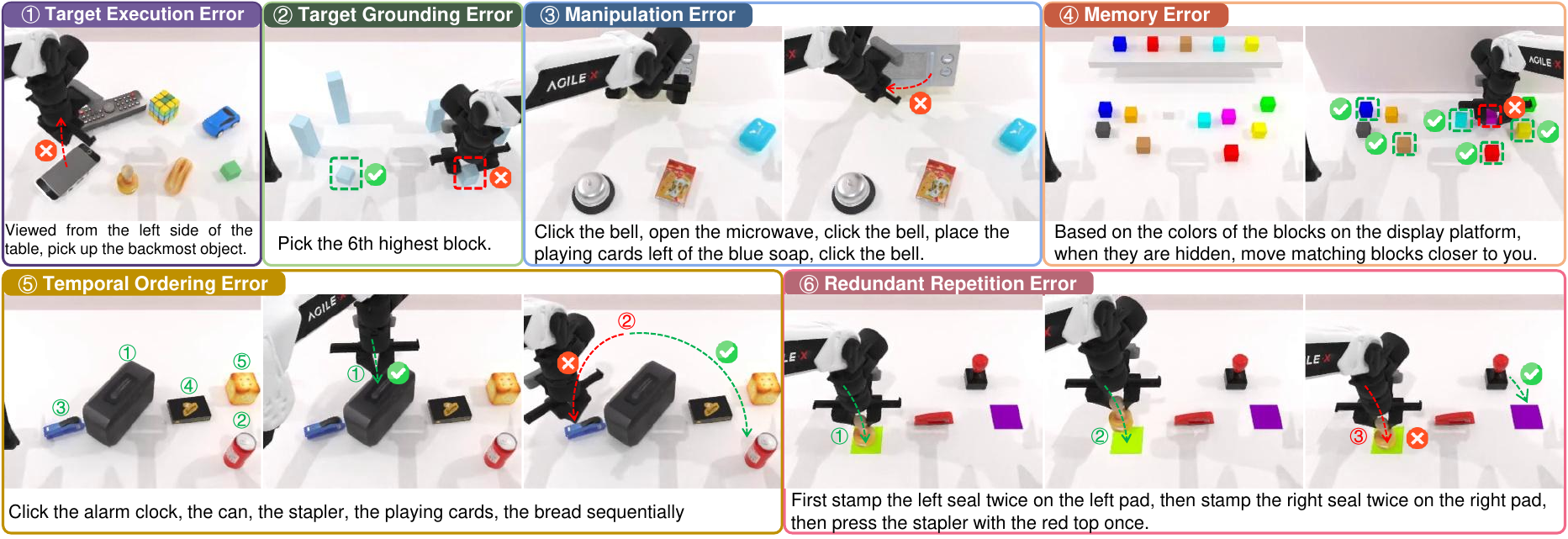}
\caption{
\textbf{Representative failure modes observed on \texttt{RoboSPA}.}
Failure modes vary across different \texttt{RoboSPA} task categories, showing that our benchmark can distinguish model weaknesses along different capability dimensions.
}
    \label{fig:fail_example}
\end{figure*}

\subsection{Diagnostic Evaluation}

\paragraph{Object-Normalized Target Accuracy.}
Table~\ref{tab:onta_spatial} compares models using the $\mathcal{ONTA}$ metric.
Overall, all models achieve low scores.
RDT and GO-1 obtain negative $\mathcal{ONTA}$ on most categories and difficulty levels, showing little spatial reasoning beyond random selection. 
$\pi_{0.5}$ and X-VLA show fewer negative scores and achieve better average $\mathcal{ONTA}$, but their overall scores remain close to the random-selection baseline. 
These results suggest that current VLA models still struggle to identify the correct target when success depends on fine-grained spatial cues.

\paragraph{Progress Score.}
Table~\ref{tab:long_horizon_completion} reports \textit{Progress Score} for long-horizon procedural planning tasks, providing a progress-based view beyond binary \textit{Success Rate}.
The results show that \textit{Progress Score} is generally higher than final \textit{Success Rate}, indicating that models can often execute part of a procedure but fail due to accumulated errors, incorrect ordering, or incomplete later-stage execution. $\pi_{0.5}$ achieves the strongest average completion rate, followed by X-VLA. 
Nevertheless, Memory-Intensive Planning (MIP) remains weak at higher difficulty levels, indicating that current models still have limited memory ability during long-horizon execution.

\subsection{Qualitative Results}
Fig.~\ref{fig:benchmark_radar} visualizes category-wise model performance across task difficulty. 
The radar plots show that model performance is highly uneven across categories and shrinks markedly under harder settings. 
This reveals persistent limitations in handling increasingly complex embodied reasoning tasks.

\subsection{Failure Analysis}
We further analyze model failures. 
Fig.~\ref{fig:fail_example} presents six representative failure modes, with quantitative details provided in the Appendix~\ref{app:detailed_error_analysis}.
\paragraph{Observed Failure Modes.}
For fine-grained spatial reasoning, target execution error denotes failed manipulation after correct grounding, while target grounding error denotes selecting a distractor. 
For long-horizon procedural planning, manipulation error refers to failed intermediate execution, memory error to forgetting earlier task-relevant information, temporal ordering error to executing subtasks in the wrong order, and redundant repetition error to repeating completed actions instead of progressing.

\paragraph{Implications for Future Models.}
These findings suggest several directions for future VLA models. 
Improving spatial performance requires stronger object-centric grounding and relation awareness, together with reliable low-level execution.
Long-horizon tasks require stronger progress tracking, memory, and execution monitoring to maintain task state. These abilities help models follow temporal constraints and reduce ordering errors, redundant repetitions, and cascading failures.

%% file: tables/table2.tex
\begin{table*}[t]
\centering
\caption{
Comparison of \textit{Object-Normalized Target Accuracy} ($\mathcal{ONTA}$) on fine-grained spatial reasoning tasks.
}
\label{tab:onta_spatial}
\resizebox{\textwidth}{!}{
\begin{tabular}{p{5.2cm}cccccccccccc}
\toprule
\multirow{2}{*}{Task}
& \multicolumn{3}{c}{RDT}
& \multicolumn{3}{c}{GO-1}
& \multicolumn{3}{c}{$\pi_{0.5}$}
& \multicolumn{3}{c}{X-VLA} \\
\cmidrule(lr){2-4}
\cmidrule(lr){5-7}
\cmidrule(lr){8-10}
\cmidrule(lr){11-13}
& L1 & L3 & L5
& L1 & L3 & L5
& L1 & L3 & L5
& L1 & L3 & L5 \\
\midrule
Geometric Attribute Cognition 
& -80.0 & -25.0 & -14.3
& -44.5 & -7.0 & -6.2
& -14.5 & 1.3 & -5.0
& \cellcolor{blue!15}15.5 & \cellcolor{blue!15}11.0 & \cellcolor{blue!15}10.3 \\

Spatial Distance Estimation 
& -27.2 & -9.1 & -4.6
& -19.7 & -6.7 & -2.1
& 2.5 & 13.6 & 11.6
& \cellcolor{blue!15}12.1 & \cellcolor{blue!15}18.2 & \cellcolor{blue!15}24.4 \\

Canonical Position Indexing 
& -27.5 & -4.9 & 0.3
& -17.0 & -3.3 & -1.2
& \cellcolor{blue!15}16.0 & \cellcolor{blue!15}10.2 & \cellcolor{blue!15}16.7
& -2.3 & 2.2 & 0.5 \\

Referential Relational Reasoning 
& -91.2 & -25.1 & -7.6
& -65.6 & -18.9 & -5.1
& -33.6 & \cellcolor{blue!15}0.5 & -1.5
& \cellcolor{blue!15}-19.2 & 0.3 & \cellcolor{blue!15}3.5 \\

Cross-View Reasoning 
& -45.5 & -28.8 & -28.8
& -23.9 & -14.7 & -11.7
& \cellcolor{blue!15}11.8 & 16.0 & \cellcolor{blue!15}16.8
& 7.3 & \cellcolor{blue!15}17.3 & 15.5 \\

\midrule
\textit{Average} 
& -54.3 & -18.6 & -11.0
& -34.1 & -10.1 & -5.3
& -3.6 & 8.3 & 7.7
& \cellcolor{blue!15}2.7 & \cellcolor{blue!15}9.8 & \cellcolor{blue!15}10.8 \\
\bottomrule
\end{tabular}
}
\end{table*}

%% file: tables/table3.tex
\begin{table*}[t]
\centering
\caption{
Performance comparison in terms of  \textit{
Progress Score} ($\mathcal{PS}$) on long-horizon procedural planning tasks.
}
\label{tab:long_horizon_completion}
\resizebox{\textwidth}{!}{
\begin{tabular}{lcccccccccccc}
\toprule
\multirow{2}{*}{Task}
& \multicolumn{3}{c}{RDT}
& \multicolumn{3}{c}{GO-1}
& \multicolumn{3}{c}{$\pi_{0.5}$}
& \multicolumn{3}{c}{X-VLA} \\
\cmidrule(lr){2-4}
\cmidrule(lr){5-7}
\cmidrule(lr){8-10}
\cmidrule(lr){11-13}
& L1 & L3 & L5
& L1 & L3 & L5
& L1 & L3 & L5
& L1 & L3 & L5 \\
\midrule

Repetitive Procedure Following 
& 40.8 & 46.3 & 44.5
& 31.8 & 35.0 & 32.0
& \cellcolor{blue!15}80.3 & \cellcolor{blue!15}78.7 & \cellcolor{blue!15}77.3
& 34.8 & 34.8 & 31.9 \\

Order-Free Execution 
& 20.3 & 33.0 & 24.9
& 41.1 & 39.7 & 25.7
& 85.8 & 75.0 & 62.1
& \cellcolor{blue!15}86.0 & \cellcolor{blue!15}84.4 & \cellcolor{blue!15}73.9 \\

Order-Constrained Execution 
& 13.6 & 25.3 & 14.7
& 41.9 & 24.1 & 14.8
& \cellcolor{blue!15}77.4 & 44.3 & \cellcolor{blue!15}28.0
& 70.0 & \cellcolor{blue!15}51.1 & 25.1 \\

Composite Action Coordination 
& 27.5 & 32.5 & 26.5
& 24.2 & 26.2 & 22.7
& 58.6 & \cellcolor{blue!15}60.5 & \cellcolor{blue!15}48.0
& \cellcolor{blue!15}58.9 & 49.0 & 40.1 \\

Memory-Intensive Planning 
& 18.2 & 5.0 & 3.6
& 7.6 & 4.9 & 6.0
& \cellcolor{blue!15}54.0 & \cellcolor{blue!15}18.7 & \cellcolor{blue!15}11.4
& 44.6 & 12.2 & 7.3 \\

\midrule
\textit{Average}
& 24.1 & 28.4 & 22.8
& 29.3 & 26.0 & 20.2
& \cellcolor{blue!15}71.2 & \cellcolor{blue!15}55.4 & \cellcolor{blue!15}45.4
& 58.8 & 46.3 & 35.7 \\
\bottomrule
\end{tabular}
}
\end{table*}

%% file: EMNLP/sections/5_conclusion.tex
\section{Conclusion}
We present \texttt{RoboSPA}, a reasoning-focused, large-scale robotic manipulation dataset and benchmark. Centered on Fine-Grained Spatial Reasoning and Long-Horizon Procedural Planning, \texttt{RoboSPA} covers ten task categories, five difficulty levels, multiple embodiments, and diverse scenes, with fine-grained metrics for spatial grounding and subtask completion. Experiments show that existing VLA models still struggle with complex spatial relations, precise execution, and memory-intensive planning, especially at higher difficulty levels. We hope \texttt{RoboSPA} can advance the development of more capable and reliable embodied agents.

%% file: EMNLP/sections/6_limitation.tex

\section*{Limitations}
Although our dataset covers diverse fine-grained spatial reasoning and long-horizon procedural planning tasks, several limitations remain. First, all tasks are constructed in simulation, and the sim-to-real gap may limit direct transfer to physical robots. Second, our benchmark focuses on tabletop manipulation, which may not fully capture the diversity and open-endedness of real-world environments. Third, while we include multiple robotic embodiments and domain-randomized scenes, broader settings such as deformable object manipulation, human-robot interaction, and open-ended task instructions remain underexplored.

%% file: EMNLP/sections/7_ethics_statement.tex
\section*{Ethics Statement}

\paragraph{Scope and Safety.}
\texttt{RoboSPA} is designed to diagnose and evaluate VLA models in simulated robotic manipulation environments, rather than to serve as a directly deployable real-world robotic system. Data collection is conducted entirely in simulation and does not involve real-world robot deployment. Since models evaluated or developed with \texttt{RoboSPA} may eventually be transferred to physical environments, such deployment should include appropriate safety constraints, human oversight, and task-specific validation.

\paragraph{Data Content and Privacy.}
\texttt{RoboSPA} does not involve human-subject experiments, personal data, or real-world private data, and thus poses no privacy risks to human participants. We checked the task instructions, object categories, scene assets, and collected demonstrations to ensure that the dataset does not contain personally identifying information or offensive, hateful, or explicit content. Therefore, no additional anonymization is required.

\paragraph{Artifact Licensing.}
\texttt{RoboSPA} builds upon third-party open-source resources. We use the SAPIEN simulator, which is released under the Apache License 2.0, and reuse code and 3D assets from RoboTwin 2.0 and RMBench, which are released under the MIT License. Baseline models and pretrained checkpoints used in our experiments are obtained from their official releases and used in accordance with their original license terms. These third-party artifacts are used consistently with their intended research purposes for simulation-based robotic manipulation, benchmark construction, data generation, and evaluation. We retain the original copyright and license notices of all third-party resources. The artifacts introduced by \texttt{RoboSPA}, including task definitions, benchmark code, evaluation scripts, and collected demonstration data, will be released under the MIT License.

\paragraph{Responsible Use.}
We acknowledge that simulator design, task selection, object categories, and embodiment choices may introduce biases and limit transferability, which we mitigate through multiple embodiments, diverse task categories, controlled difficulty levels, and both clean and domain-randomized scenes. We also encourage responsible use of \texttt{RoboSPA}, including efficient training, transparent reporting of computational costs, and consideration of environmental impact.

%% file: EMNLP/sections/9_acknowledgments.tex
\section*{Acknowledgments}
This work was supported by the National Key R\&D Program of China (2025ZD0123100), the NSFC (62272411), the Zhejiang NSF (LRG25F020001), and the Key R\&D Program of Zhejiang Province (2025C01030).

%% file: EMNLP/sections/8_appendix.tex
\appendix

\section*{Appendix}
\label{sec:appendix}

In this document, we provide additional details about \texttt{RoboSPA}. The appendix is organized as follows:

\begin{itemize}[leftmargin=*, itemsep=0.3em]
    \item \S\ref{app:detailed_error_analysis}: Detailed Failure Analysis. 
    \item \S\ref{app:evaluation_details}: Training and Evaluation Details.
    \item \S\ref{app:dataset_details}: Dataset Details.
    \item \S\ref{app:basic_grounding_tasks}: Basic Grounding Tasks.    
    \item \S\ref{app:task_list}: Task List and Descriptions.    
    \item \S\ref{app:additional_results}: Detailed Experimental Results.
    \item \S\ref{app:task_visualizations}: Task Visualizations.
\end{itemize}

\section{Detailed Failure Analysis}
\label{app:detailed_error_analysis}
\subsection{Analysis Setup}

\begin{figure*}[!t]
    \centering
    \includegraphics[width=0.95\textwidth]{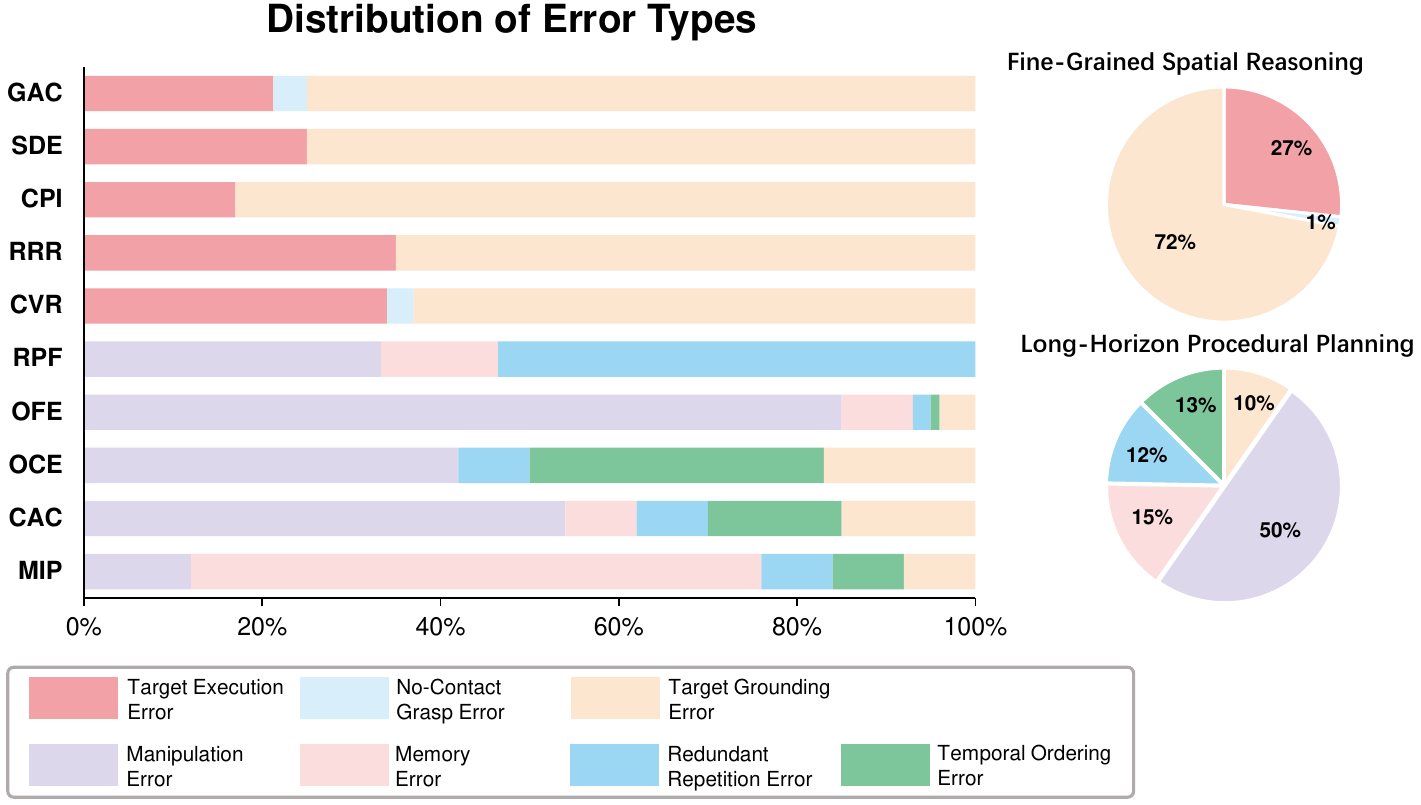}
    \caption{Distribution of error types for the $\pi_{0.5}$ model.}
    \label{fig:pi05_errors}

    \vspace{2mm}

    \includegraphics[width=0.95\textwidth]{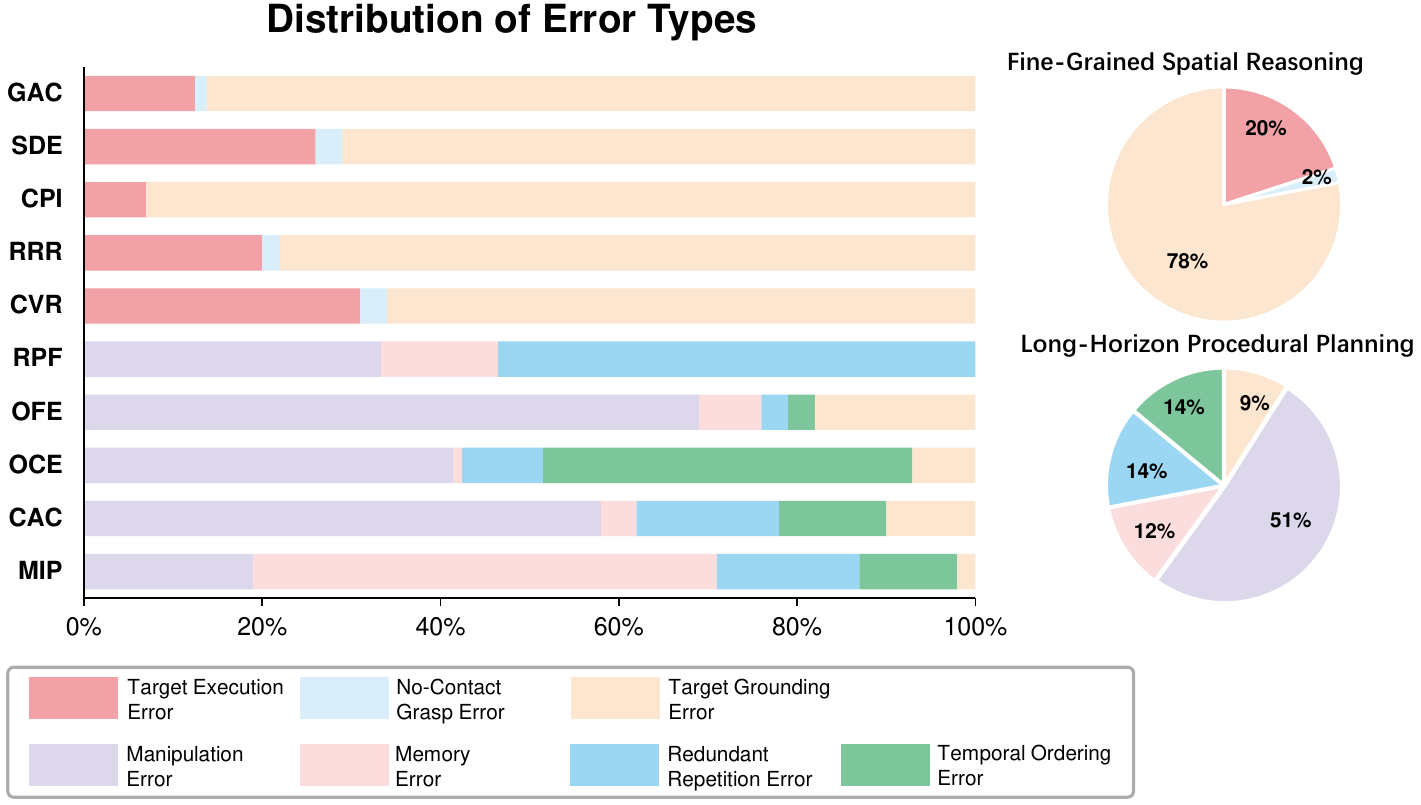}
    \caption{Distribution of error types for the X-VLA model.}
    \label{fig:xvla_errors}
\end{figure*}

We perform a detailed error analysis on the two best-performing models, $\pi_{0.5}$ and X-VLA. 
For each model, we analyze approximately 1,120 evaluation videos, sampling up to 20 error examples per L5 task; if fewer than 20 errors occur, we include all available examples. 
For fine-grained spatial reasoning tasks, we define three error types to distinguish target selection failures from execution failures:

\begin{itemize}[leftmargin=*, itemsep=0.3em]
    \item \textbf{Target Execution Error:} The model selects the correct target but fails to execute the grasp or pick action properly.
    \item \textbf{No-Contact Grasp Error:} The model attempts to grasp but fails to contact any object.
    \item \textbf{Target Grounding Error:} The model selects the wrong object instead of the intended target.
\end{itemize}

For long-horizon procedural planning tasks, we define five error types. One of them is the Target Grounding Error (as above), and the remaining four are:

\begin{itemize}[leftmargin=*, itemsep=0.3em]
    \item \textbf{Manipulation Error:} Low-level execution of a subtask fails (e.g., pick, pull, or move).
    \item \textbf{Memory Error:} The model fails to retain relevant information from previous steps, affecting subtask completion.
    \item \textbf{Temporal Ordering Error:} Subtasks are executed in an incorrect order.
    \item \textbf{Redundant Repetition Error:} The model repeats an already completed action instead of progressing to the next subgoal.
\end{itemize}

\subsection{Failure Analysis Results}
Figure~\ref{fig:pi05_errors} and Figure~\ref{fig:xvla_errors} show the distribution of error types for the $\pi_{0.5}$ and X-VLA models, respectively. 
Overall, the two models exhibit broadly similar error patterns across the two capability dimensions, although the exact proportions vary.

\input{EMNLP/tables/assets}
\paragraph{Fine-Grained Spatial Reasoning.} 
In these tasks, both models are primarily challenged by Target Grounding Errors, which account for over 70\% of all errors. 
The remaining errors are mostly Target Execution Errors, while No-Contact Grasp Errors occur very rarely. 
Among individual tasks, Canonical Position Indexing (CPI) exhibits the highest proportion of Target Grounding Errors, while Cross-View Reasoning (CVR) has a relatively larger share of Target Execution Errors compared with the other tasks.
This indicates that the main challenge lies in insufficient spatial reasoning, while low-level execution errors are comparatively rare.

\paragraph{Long-Horizon Procedural Planning.}
Within this dimension, errors are diverse. Overall, Manipulation Errors account for the largest share, approximately 50\%, while the remaining error types occur at roughly similar rates. 

Repetitive Procedure Following (RPF) primarily exhibits Redundant Repetition Errors, suggesting that models lack effective tracking of task progress and may decide the next action based solely on the current visual observation.
Order-Free Execution (OFE) and Composite Action Coordination (CAC) are dominated by Manipulation Errors, highlighting difficulties in low-level execution across multiple subgoals. 
Order-Constrained Execution (OCE) shows that, aside from Manipulation Errors, most remaining failures are Temporal Ordering Errors, indicating that models struggle to correctly sequence subtasks.
Memory-Intensive Planning (MIP) is primarily affected by Memory Errors, indicating limitations in retaining task-relevant information across extended action sequences. 

\paragraph{Summary.}
These task-specific error patterns demonstrate that different tasks in \texttt{RoboSPA} correspond to distinct failure modes, providing a benchmark with good task differentiation for diagnosing VLA model capabilities. 
They further suggest that future VLA models should strengthen spatial grounding, ensure precise low-level execution, implement effective progress tracking, and incorporate memory mechanisms to improve long-horizon tasks.

\section{Training and Evaluation Details}
\label{app:evaluation_details}
\input{EMNLP/tables/model_details}
\paragraph{Training.}
For each base task, we use 50 trajectories per difficulty level, resulting in 250 trajectories across five levels. For each task variant, we generate 60 templates, with 50 used for training and 10 held out for evaluation. These templates are instantiated with task-specific object descriptions to form 100 distinct training instructions and 100 non-overlapping evaluation instructions. During training, one instruction is randomly sampled from the corresponding task-variant instruction pool for each trajectory.

\begin{figure*}[!t]
\centering
\includegraphics[width=\textwidth]{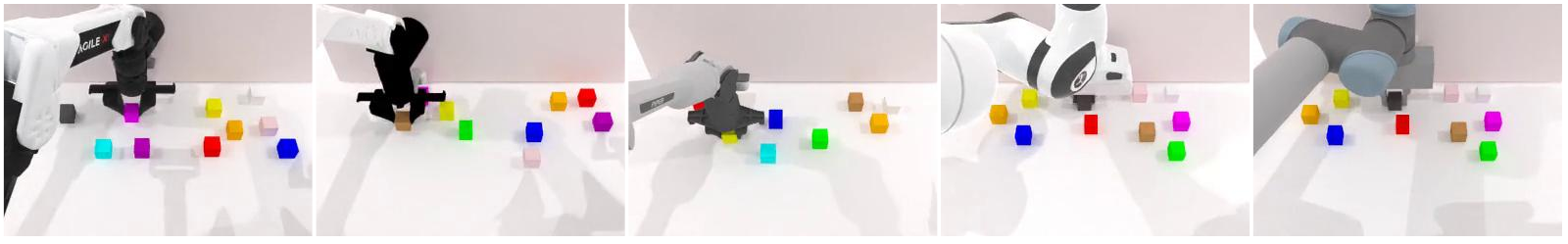}
\caption{
The five robotic embodiments supported by \texttt{RoboSPA}. From left to right: Aloha-AgileX, ARX-X5, Piper, Franka, and UR5.
}
\label{fig:multi_arms}
\end{figure*}

To ensure a fair and controlled comparison, all baseline models are initialized from their officially released pretrained weights and fully fine-tuned on our dataset, with all parameters updated during training. We follow their official implementations, recommended environments, preprocessing pipelines, and evaluation settings, with only the training data and evaluation tasks replaced by \texttt{RoboSPA}. We adopt a unified training configuration across all models, using a batch size of 16 and training for 20,000 steps. All experiments are conducted on NVIDIA RTX PRO 6000 GPUs under consistent hardware settings. Table~\ref{tab:model_size_budget} summarizes the model sizes and training compute budgets per base task, including the number of parameters, GPU configuration, training time, and GPU hours for each evaluated baseline. Across all 56 base tasks and four baseline models, the total training budget is approximately 1,568 GPU-hours.

\begin{figure*}[!t]
\centering
\includegraphics[width=\textwidth]{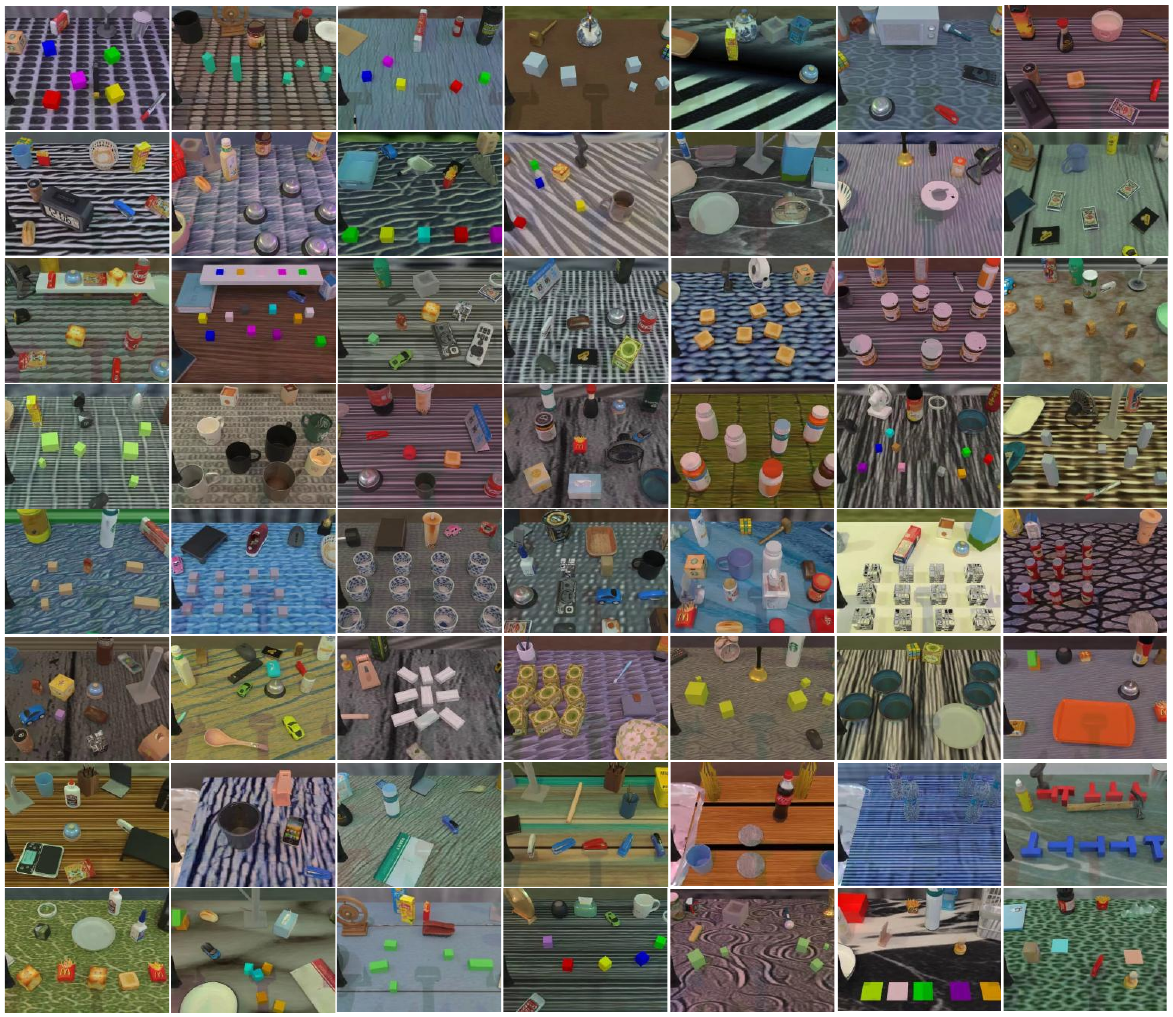}
\caption{Examples of domain-randomized scenes in \texttt{RoboSPA}.}
\label{fig:domain_randomized_scenes}
\end{figure*}

\paragraph{Evaluation.}
Each task variant is evaluated over 100 test trials, and the \textit{Success Rate} is computed accordingly. To assess generalization ability, all evaluation instructions are strictly disjoint from those used during training, ensuring no overlap between training and test instructions. In addition to binary success, each task script records the completion status of individual actions or subtasks during execution. This enables step-level fine-grained evaluation for long-horizon tasks, where a rollout may fail the final goal while still completing part of the required procedure.

\section{Dataset Details}
\label{app:dataset_details}

\subsection{Object Asset Details}
\label{app:object_asset_details}

The object assets used in \texttt{RoboSPA} are collected from the RoboTwin-OD asset library~\citep{chen2026robotwin} and RMBench~\citep{chen2026rmbenchmemorydependentroboticmanipulation}. In total, \texttt{RoboSPA} includes 120 object categories and 581 object variants, covering diverse daily-use objects with different appearances, geometries, and semantic types. These assets are used to instantiate task scenes across both fine-grained spatial reasoning and long-horizon procedural planning tasks, as summarized in Table~\ref{tab:asset_categories}.

\subsection{Multi-Embodiment Setting}
\label{app:multi-embodiment}

\texttt{RoboSPA} supports multiple robotic embodiments, including Aloha-AgileX, ARX-X5, Piper, Franka, and UR5, as shown in Figure~\ref{fig:multi_arms}. These embodiments share the same task semantics, language instructions, object configurations, and success conditions, while differing in morphology, workspace, camera viewpoint, and execution behavior. This design enables evaluation of VLA models under diverse robotic configurations and tests whether models can generalize beyond a single embodiment.

\subsection{Domain-Randomized Scenes}
\label{app:domain_randomization}

To evaluate robustness beyond clean environments, we collect additional trajectories under domain-randomized scenes, as shown in Figure~\ref{fig:domain_randomized_scenes}. The task semantics and success conditions remain unchanged, while the visual and environmental conditions are varied.

Compared with clean scenes, the randomized setting introduces four types of variations: background appearance, tabletop clutter, table height, and lighting conditions. Background randomization exposes models to diverse visual contexts, while tabletop clutter introduces distractor objects. Table-height perturbation adds mild geometric variation, and lighting randomization changes illumination conditions.


\section{Basic Grounding Tasks}
\label{app:basic_grounding_tasks}
We additionally evaluate three basic grounding variants as lower-level references for spatial reasoning. 
These variants match the object categories and candidate counts of the corresponding \texttt{RoboSPA} tasks, namely SDE, CVR, and CPI, respectively. 
Color-Based Grounding relies only on color cues, Egocentric Position Grounding uses robot-view positions without cross-view transformation, and Simple Ordinal Indexing fixes the counting directions to left-to-right and far-to-near, unlike CPI where indexing directions vary.

As shown in Table~\ref{tab:basic_spatial_tasks}, these tasks achieve relatively high \textit{Success Rate}s across difficulty levels, indicating that basic grounding tasks may be insufficient to expose the reasoning limitations of current VLA models.

\input{EMNLP/tables/v1}
In contrast, \texttt{RoboSPA} introduces fine-grained spatial reasoning settings, such as distance estimation, cross-view reasoning, relational grounding, and canonical position indexing with more complex spatial layouts. These tasks provide a more challenging and diagnostic evaluation of spatial reasoning ability.

\section{Task List and Descriptions}
\label{app:task_list}
Tables~\ref{tab:spatial_task_list} and~\ref{tab:long_horizon_task_list} provide the full list of base tasks in \texttt{RoboSPA} and their objectives. Each base task is instantiated into 5 difficulty variants, where fine-grained spatial reasoning tasks vary the number of candidate objects and long-horizon procedural planning tasks vary the number of procedural steps.

\section{Detailed Experimental Results}
\label{app:additional_results}

\subsection{Statistical Uncertainty}
\label{app:statistical_uncertainty}

We use a distinct seed for each evaluation episode, which determines the scene and object layout. Given the same checkpoint and seed, the simulated rollout is largely deterministic. To quantify the remaining finite-sample uncertainty, we report bootstrap 95\% confidence intervals for the L5 results in Table~\ref{tab:statistical_uncertainty}. The results show that the main performance trends remain unchanged after accounting for statistical uncertainty.

\subsection{Multi-Embodiment Evaluation}
\label{app:multi_embodiment}

To evaluate consistency across robotic platforms, we evaluate X-VLA on five embodiments. Due to computational resource constraints, we select one representative task from each of RoboSPA's ten capability categories. For each task and embodiment, the model is independently trained and evaluated using data from that embodiment. Tasks are denoted by their category abbreviation and index (e.g., GAC-1), with the index corresponding to their order within the category in Tables~\ref{tab:spatial_task_list} and~\ref{tab:long_horizon_task_list}.

As shown in Tables~\ref{tab:multi_embodiment_fgsr} and~\ref{tab:multi_embodiment_lhpp}, absolute performance varies across embodiments, but the L5 success rate is lower than the L1 success rate for every evaluated task and embodiment. This demonstrates that RoboSPA exhibits a consistent difficulty trend across robotic embodiments.

\subsection{Success Rates}
\label{app:detailed_per_task_results}

To provide a complete view of model performance, we report the detailed per-task \textit{Success Rate}s for all evaluated models across the five difficulty levels. Tables~\ref{tab:appendix-rdt-success}--\ref{tab:appendix-xvla-success} list the results of RDT, GO-1, $\pi_{0.5}$, and X-VLA on every task in \texttt{RoboSPA}. These tables complement the aggregated results in the main paper by showing how each model performs on individual task variants, making it easier to identify task-specific strengths and failure patterns. 

Overall, the detailed results show that models generally perform better in settings with fewer distractor objects and shorter action sequences. However, performance drops become much sharper as difficulty increases, especially for tasks involving fine-grained spatial disambiguation, strict ordering, compositional execution, or memory-intensive planning.

\subsection{Object-Normalized Target Accuracy}
\label{app:detailed_onta_results}

Tables~\ref{tab:appendix-rdt-onta}--\ref{tab:appendix-xvla-onta} report the \textit{Object-Normalized Target Accuracy} ($\mathcal{ONTA}$) of all evaluated models on fine-grained spatial reasoning tasks. These results help identify whether failures are caused by weak target disambiguation rather than low-level manipulation errors.

The $\mathcal{ONTA}$ results reveal clear differences in spatial grounding ability across models. RDT and GO-1 achieve poor scores, with many values below zero, indicating below-chance target selection on numerous spatial tasks. $\pi_{0.5}$ and X-VLA perform better, but their scores still remain low on many variants and often stay close to the random-guessing baseline. These results suggest that current VLA models still have substantial room for improvement in fine-grained spatial reasoning, especially under increasing spatial ambiguity and scene complexity.

\subsection{Progress Score}
\label{app:detailed_progress_results}

Tables~\ref{tab:appendix-rdt-step}--\ref{tab:appendix-xvla-step} report the \textit{Progress Score}s of all evaluated models on long-horizon procedural planning tasks. Unlike binary \textit{Success Rate}, the \textit{Progress Score} measures partial subtask completion during execution. These results provide a more fine-grained view of long-horizon failures by showing how much of the required procedure each model completes before failing.

The results highlight the limited long-horizon planning ability of current models. RDT and GO-1 obtain consistently low \textit{Progress Scores}, with most tasks scoring below 50, suggesting that they often fail to complete even half of the required procedure. $\pi_{0.5}$ and X-VLA achieve higher \textit{Progress Scores}, indicating better ability to complete intermediate subtasks. However, their scores still decline clearly as difficulty increases, showing that current VLA models remain far from robust long-horizon procedural planning.

\section{Task Visualizations}
\label{app:task_visualizations}

\begin{figure*}[t]
\centering

\includegraphics[width=\linewidth]{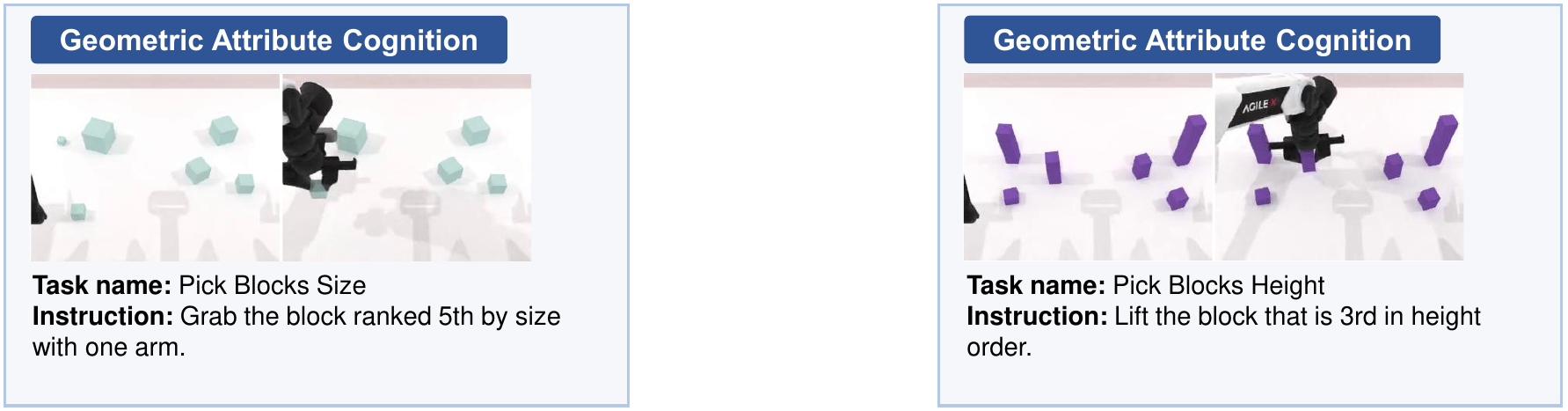}
\vspace{2pt}  

\includegraphics[width=\linewidth]{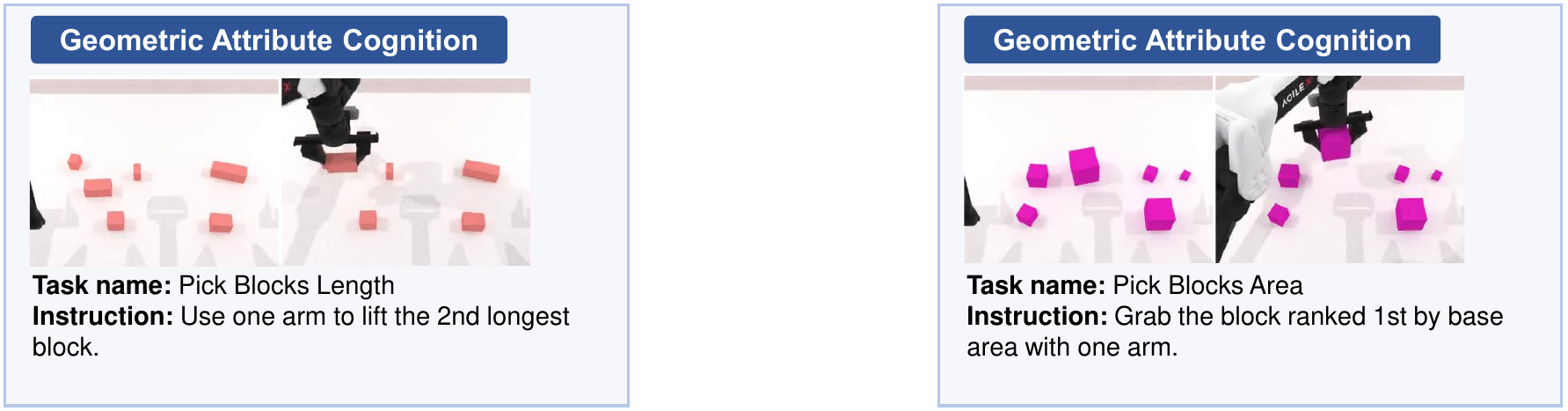}
\vspace{2pt}

\includegraphics[width=\linewidth]{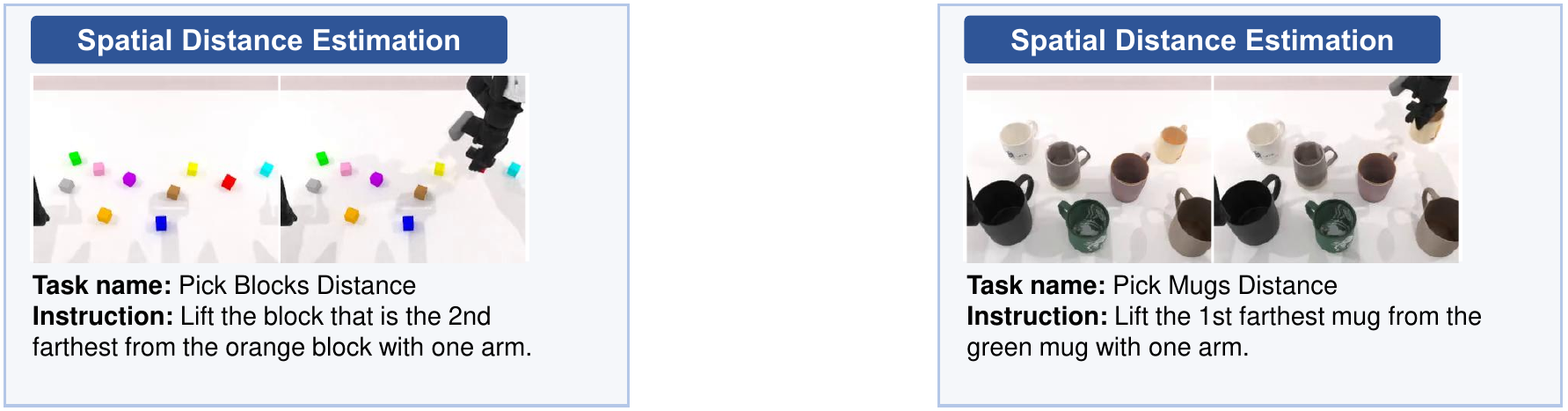}
\vspace{2pt}

\includegraphics[width=\linewidth]{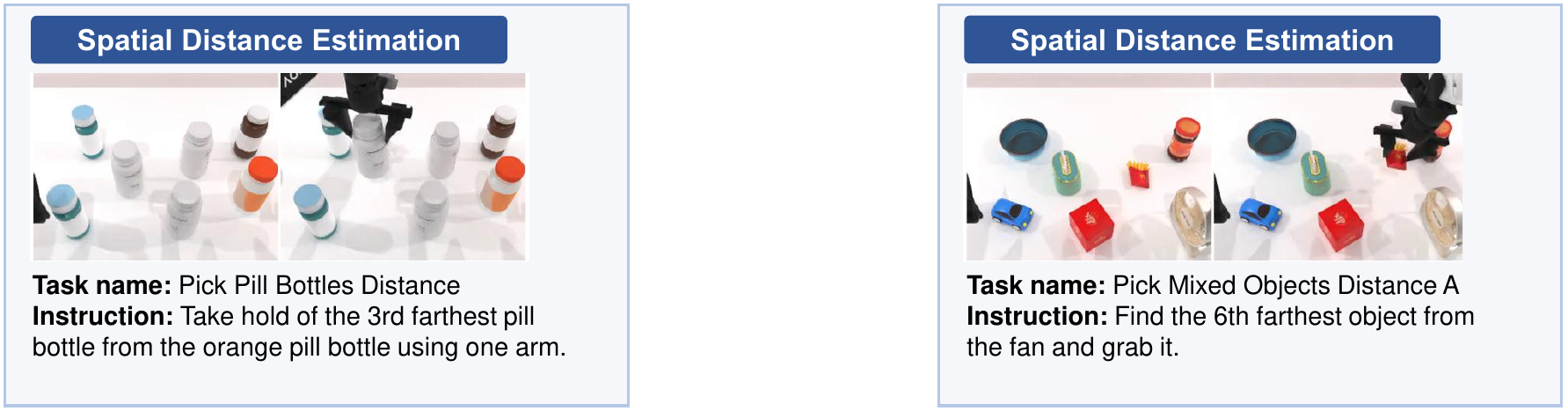}
\vspace{2pt}

\includegraphics[width=\linewidth]{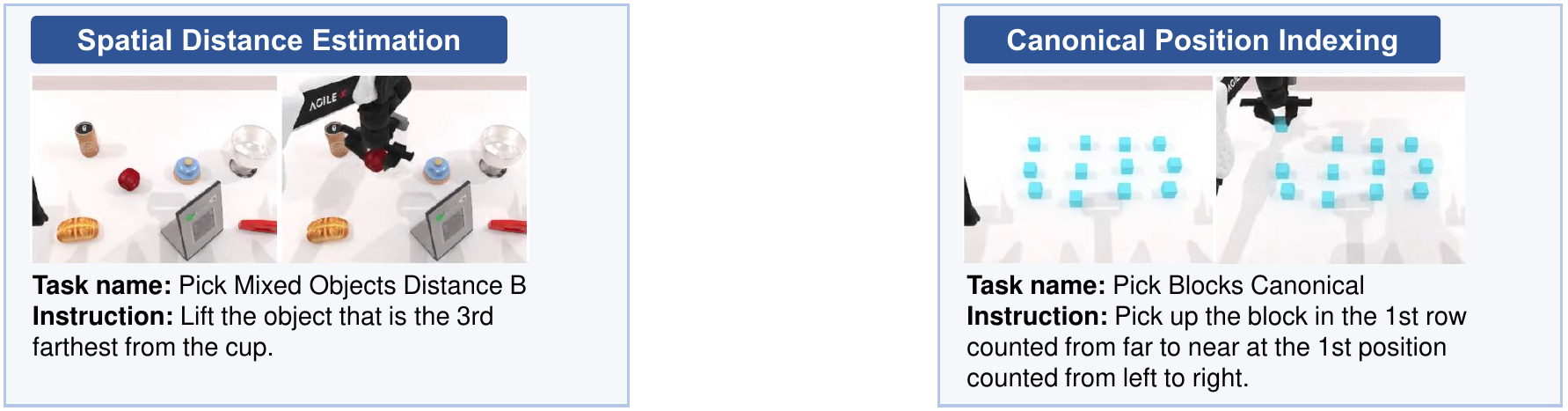}

\caption{Fine-grained spatial reasoning task examples in \texttt{RoboSPA} (Part 1).}
\label{fig:spatial_reasoning_1}
\end{figure*}

\begin{figure*}[t]
\centering

\includegraphics[width=\linewidth]{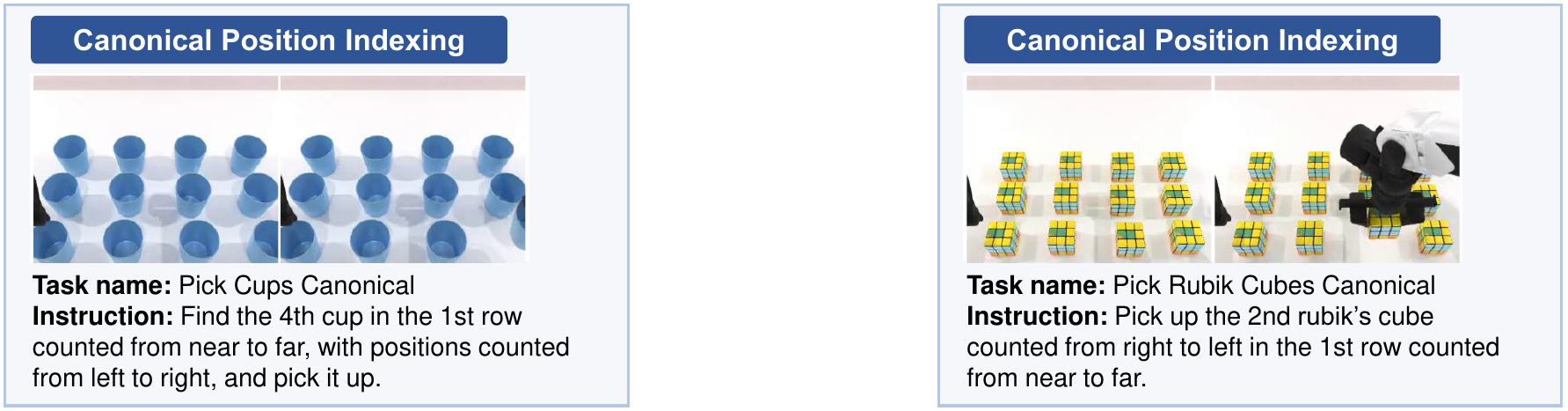}
\vspace{2pt}  

\includegraphics[width=\linewidth]{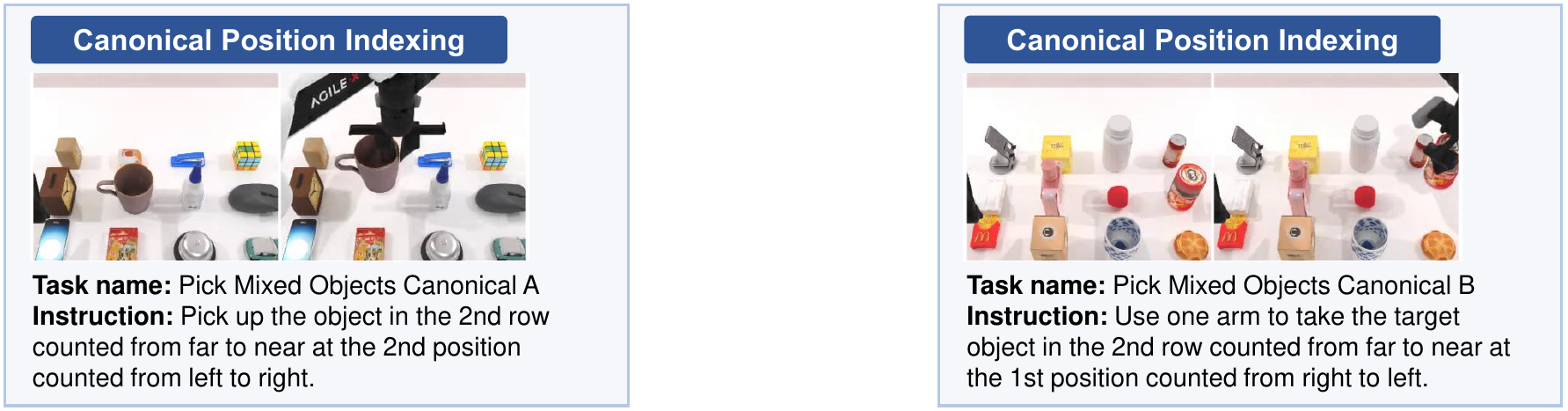}
\vspace{2pt}

\includegraphics[width=\linewidth]{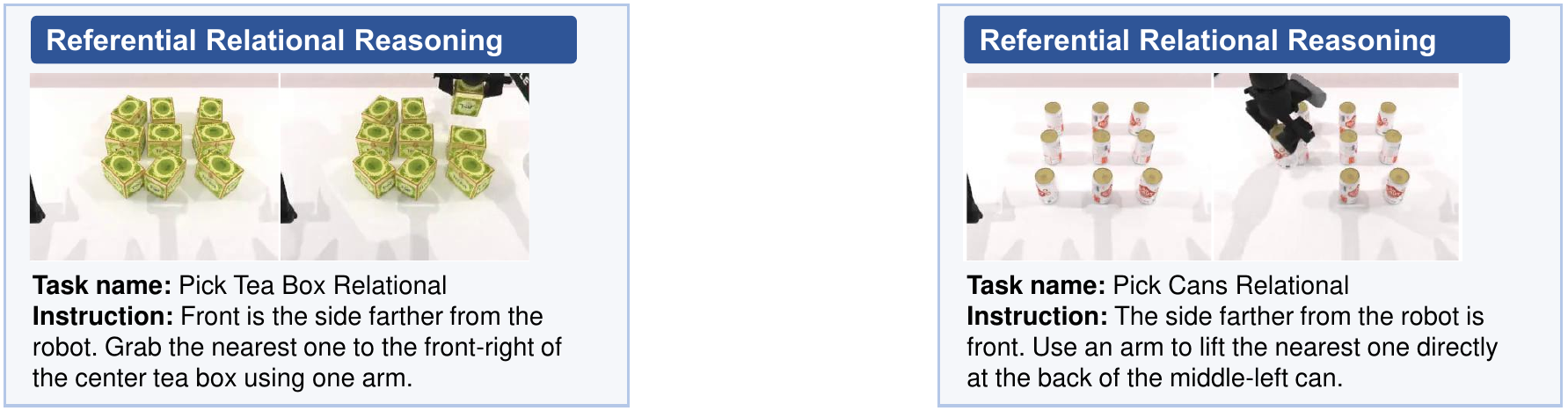}
\vspace{2pt}

\includegraphics[width=\linewidth]{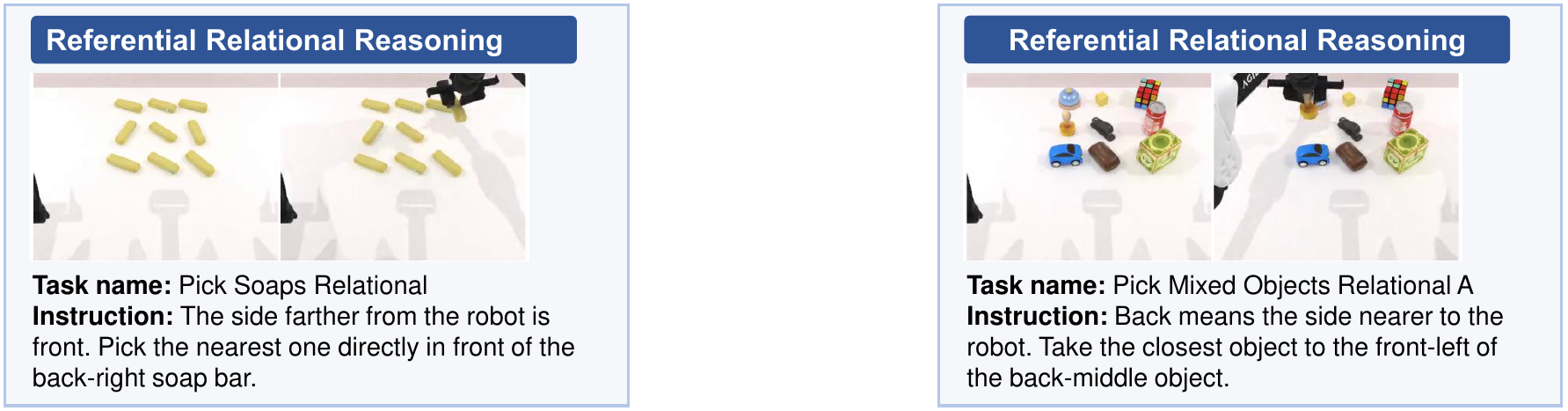}
\vspace{2pt}

\includegraphics[width=\linewidth]{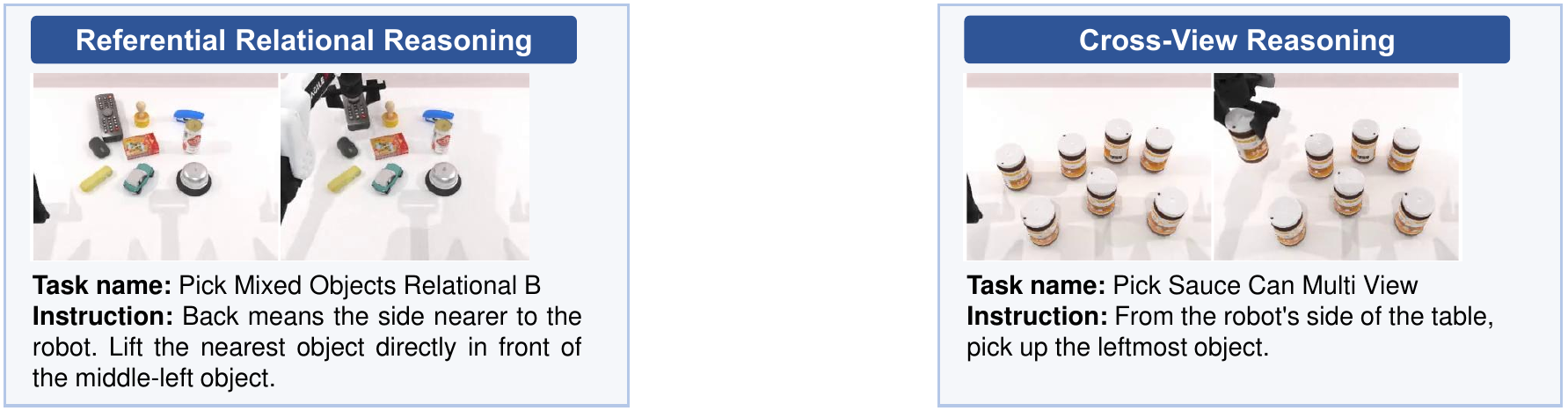}

\caption{Fine-grained spatial reasoning task examples in \texttt{RoboSPA} (Part 2).}
\label{fig:spatial_reasoning_2}
\end{figure*}

\begin{figure*}[t]
\centering

\includegraphics[width=\linewidth]{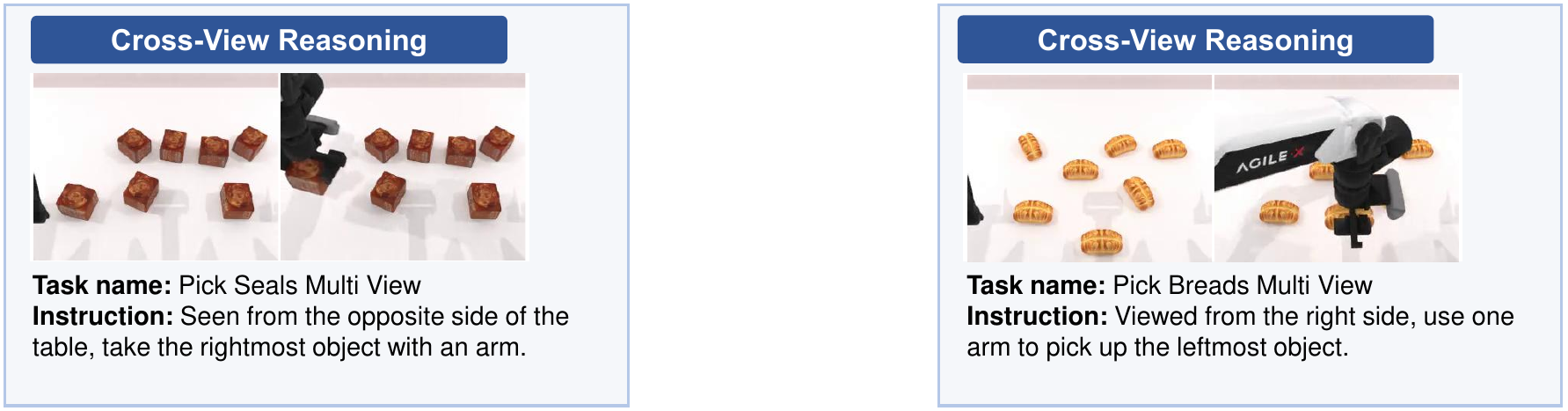}
\vspace{2pt}  

\includegraphics[width=\linewidth]{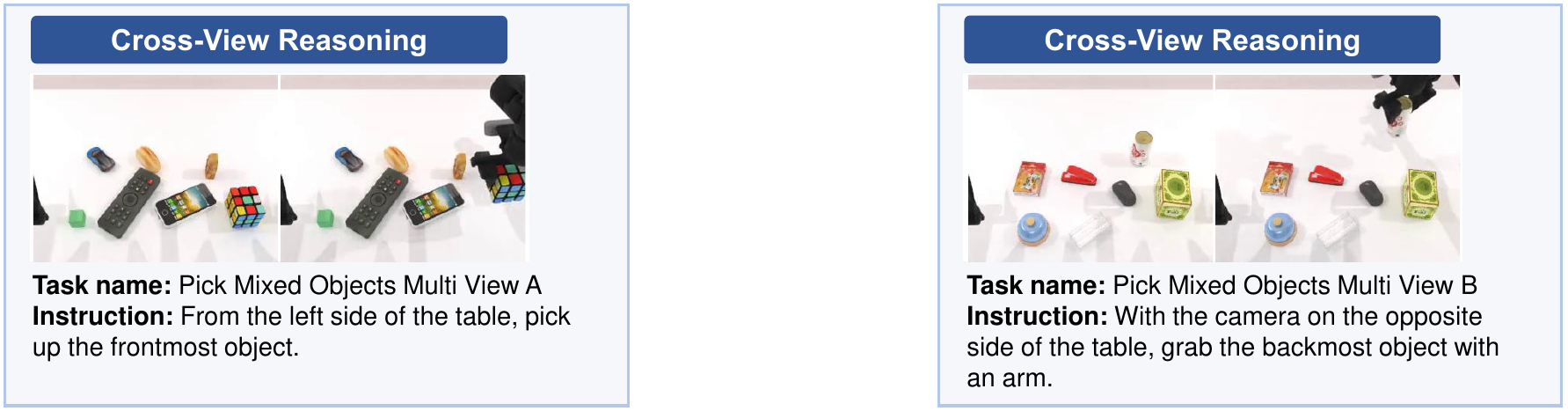}
\caption{Fine-grained spatial reasoning task examples in \texttt{RoboSPA} (Part 3).}
\label{fig:spatial_reasoning_3}
\end{figure*}

\begin{figure*}[t]
\centering

\includegraphics[width=\linewidth]{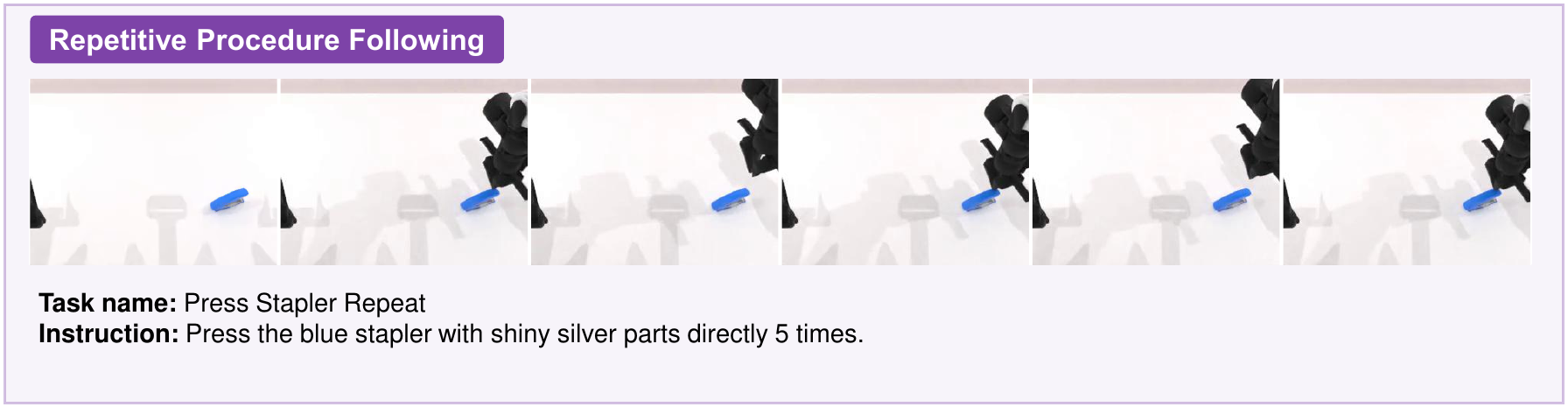}
\vspace{2pt}  

\includegraphics[width=\linewidth]{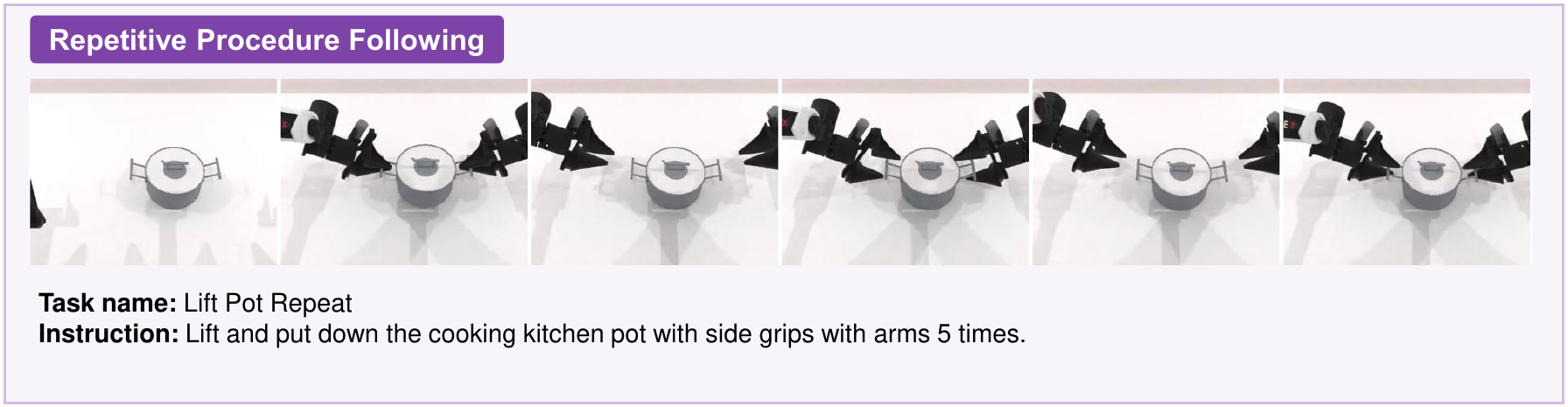}
\vspace{2pt}

\includegraphics[width=\linewidth]{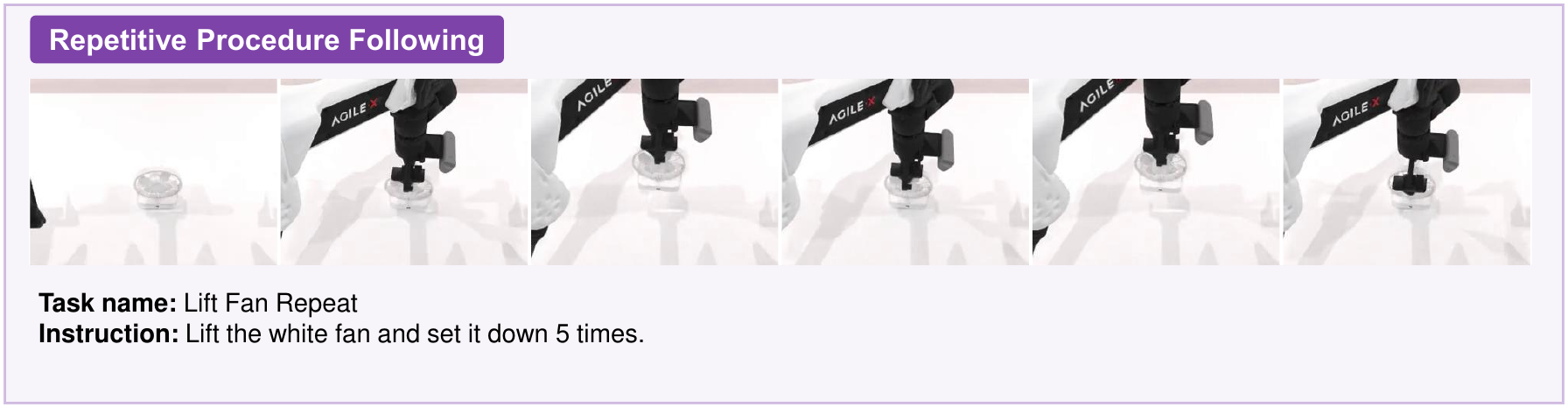}
\vspace{2pt}

\includegraphics[width=\linewidth]{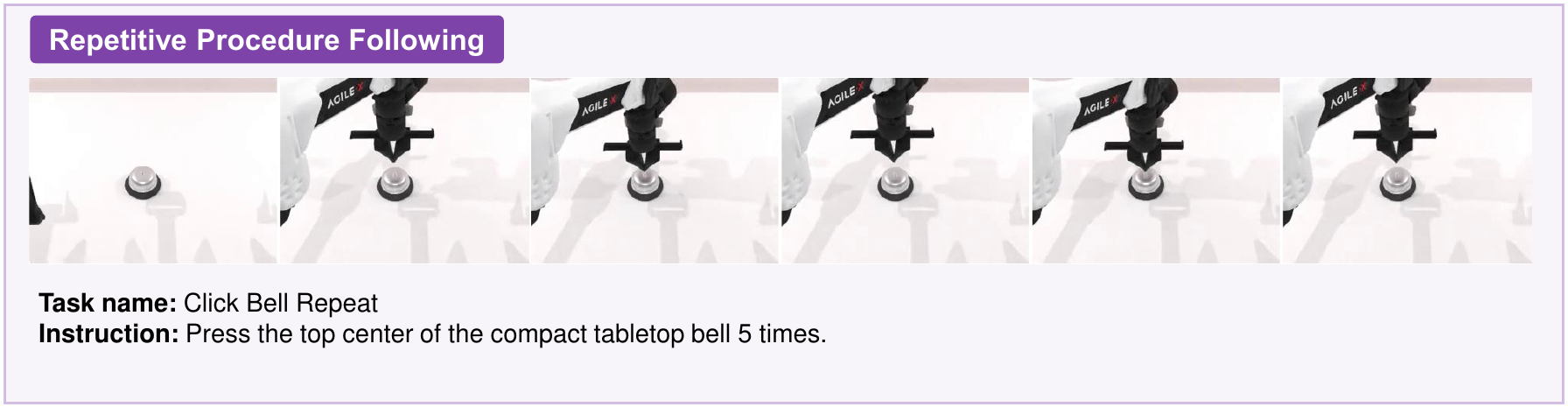}
\vspace{2pt}

\includegraphics[width=\linewidth]{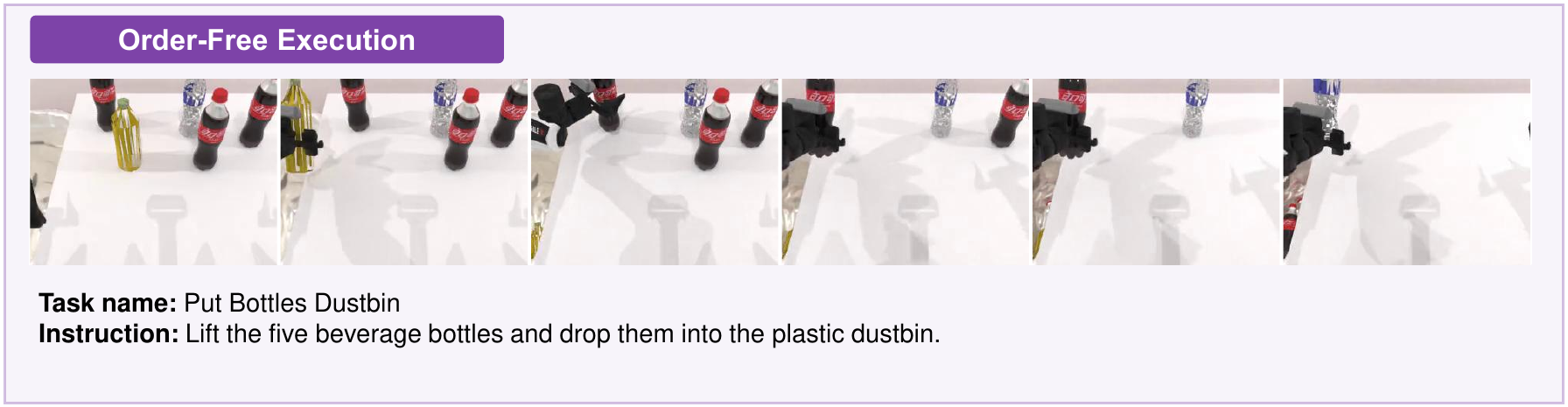}

\caption{Long-horizon procedural planning task examples in \texttt{RoboSPA} (Part 1).}
\label{fig:LH_pdf}
\end{figure*}

\begin{figure*}[t]
\centering

\includegraphics[width=\linewidth]{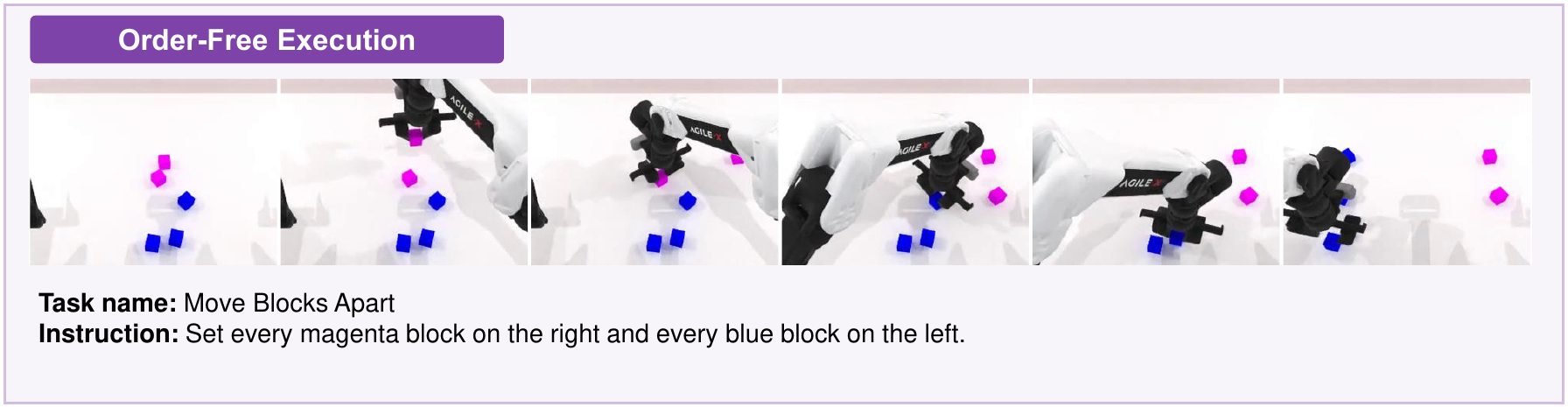}
\vspace{2pt}  

\includegraphics[width=\linewidth]{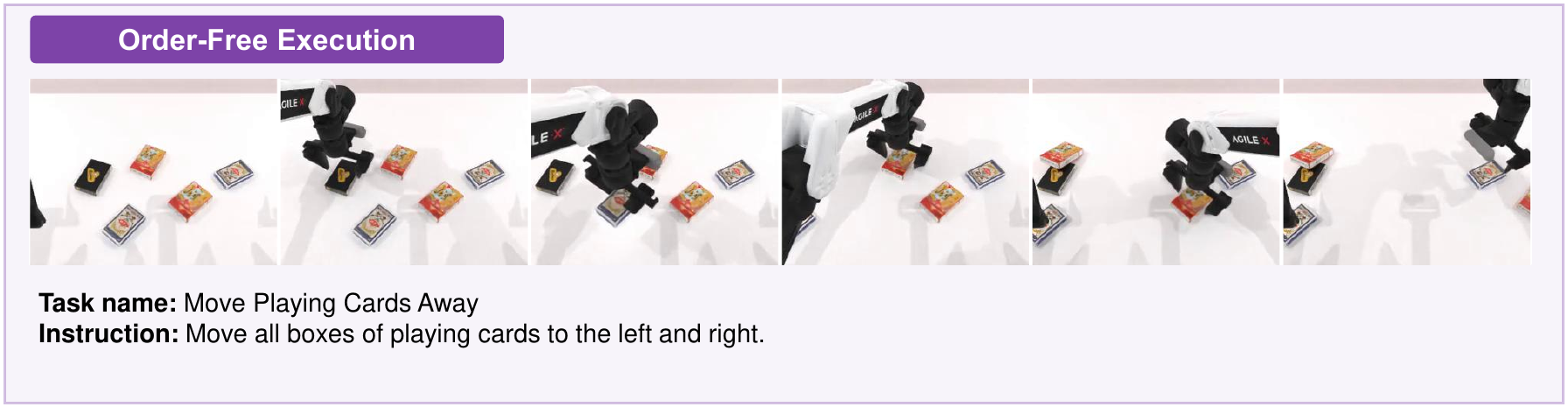}
\vspace{2pt}

\includegraphics[width=\linewidth]{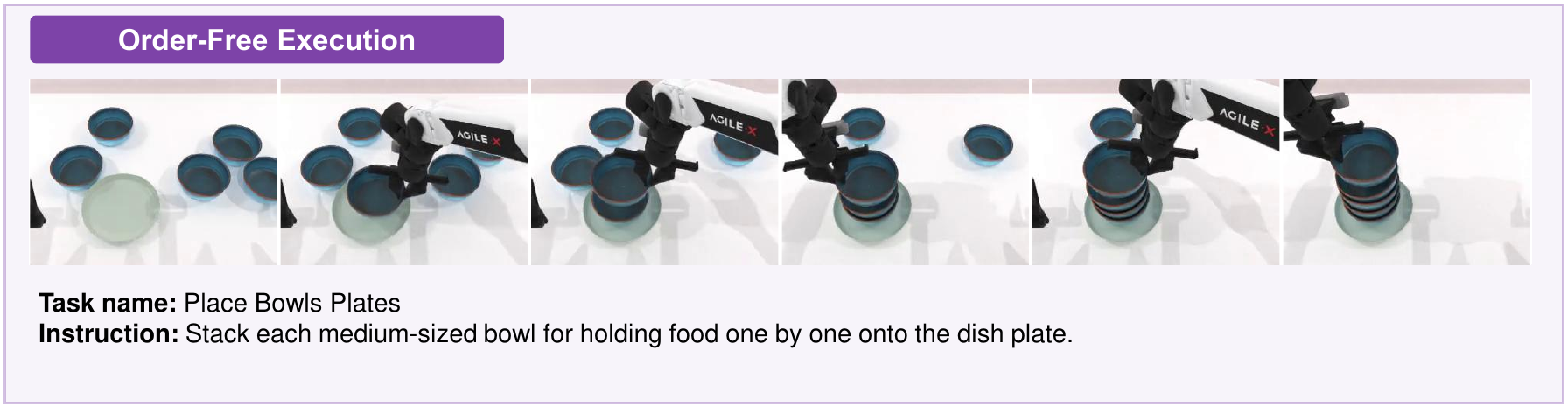}
\vspace{2pt}

\includegraphics[width=\linewidth]{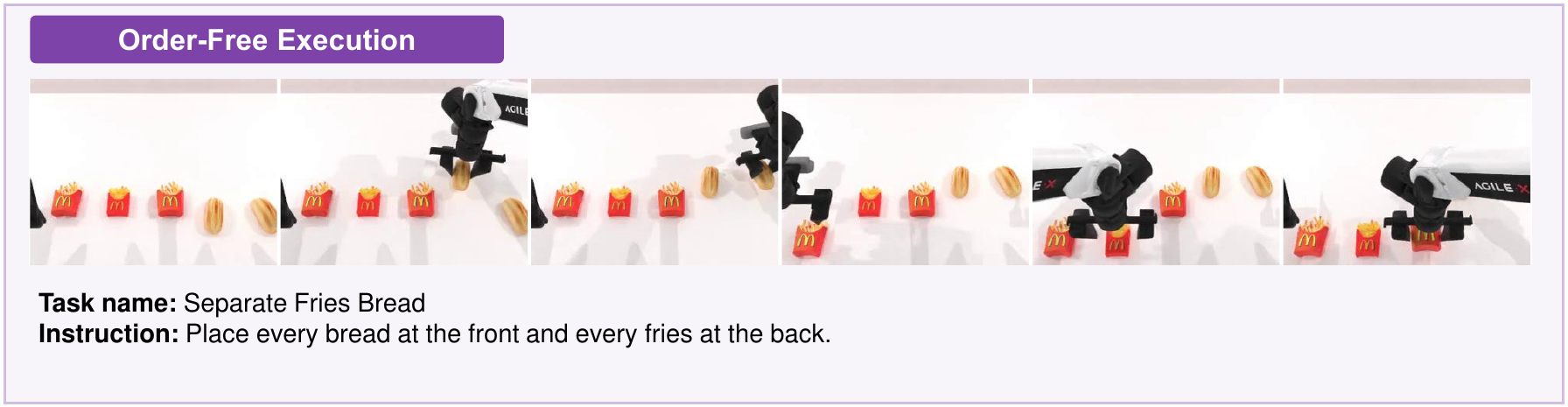}
\vspace{2pt}

\includegraphics[width=\linewidth]{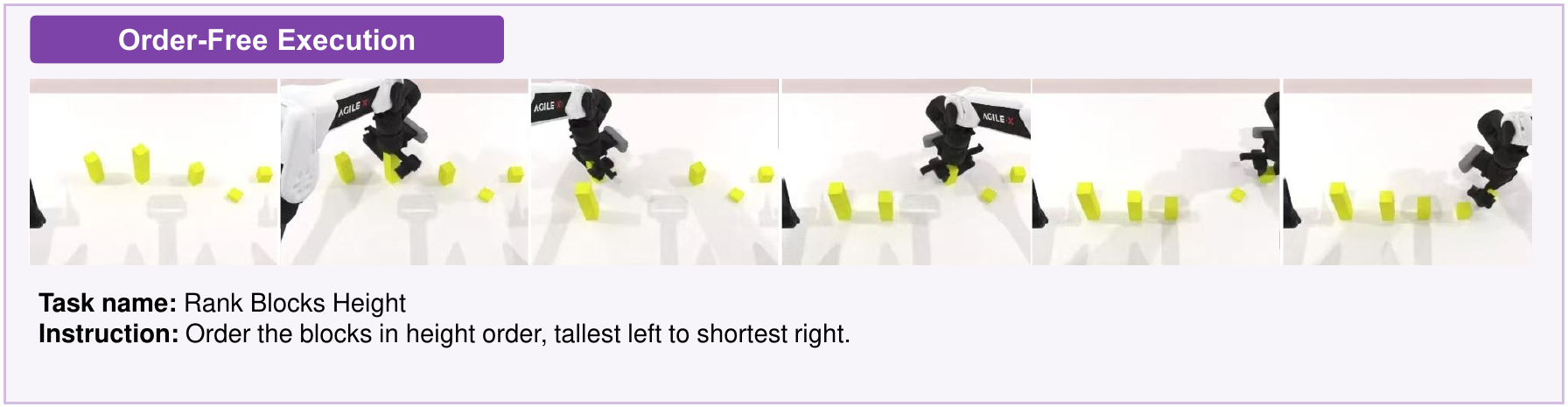}

\caption{Long-horizon procedural planning task examples in \texttt{RoboSPA} (Part 2).}
\label{fig:LH_2}
\end{figure*}

\begin{figure*}[t]
\centering

\includegraphics[width=\linewidth]{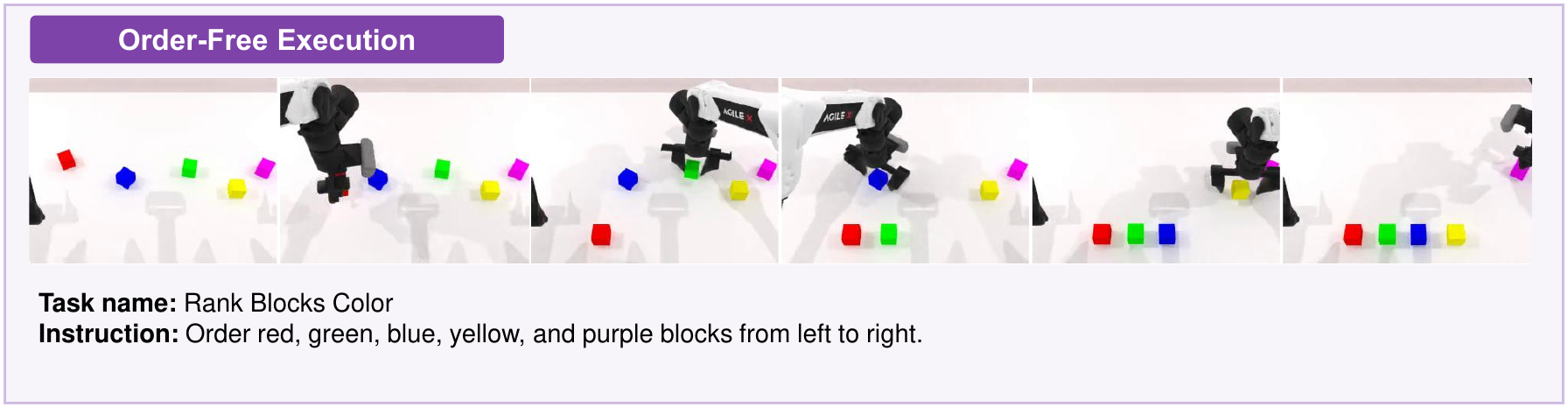}
\vspace{2pt}  

\includegraphics[width=\linewidth]{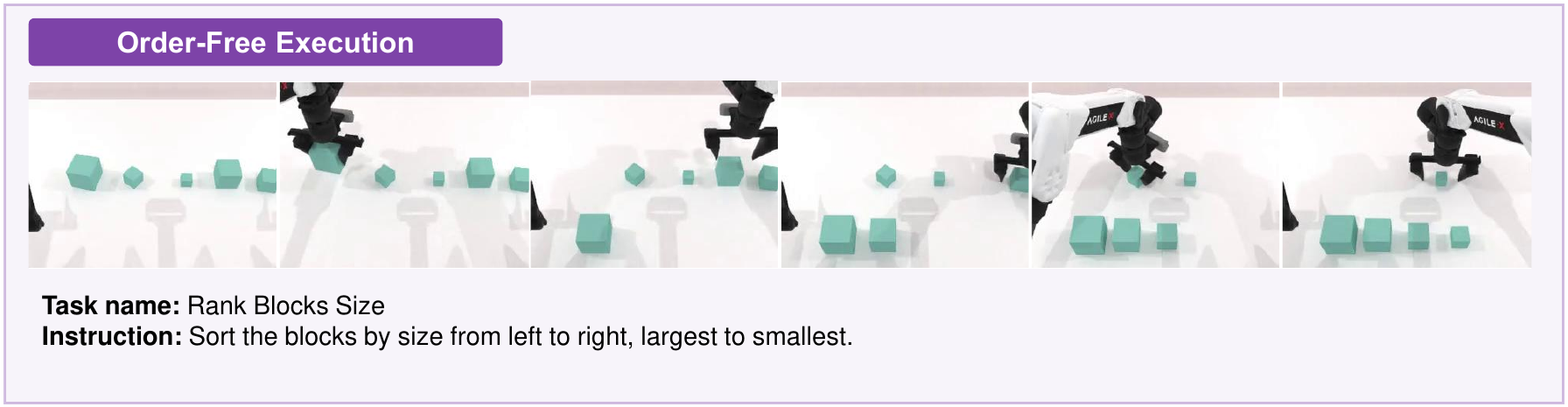}
\vspace{2pt}

\includegraphics[width=\linewidth]{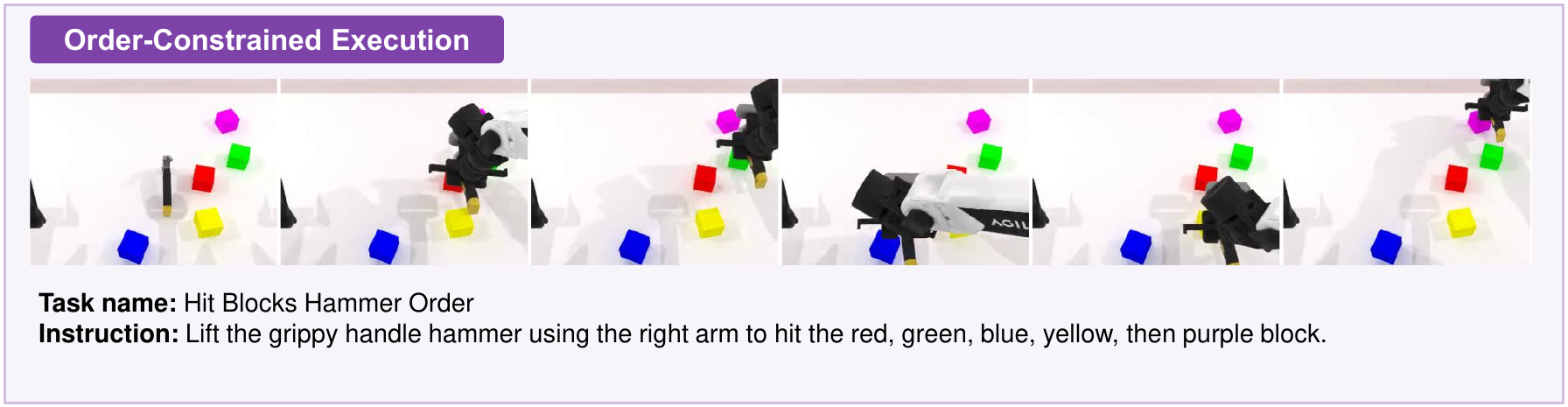}
\vspace{2pt}

\includegraphics[width=\linewidth]{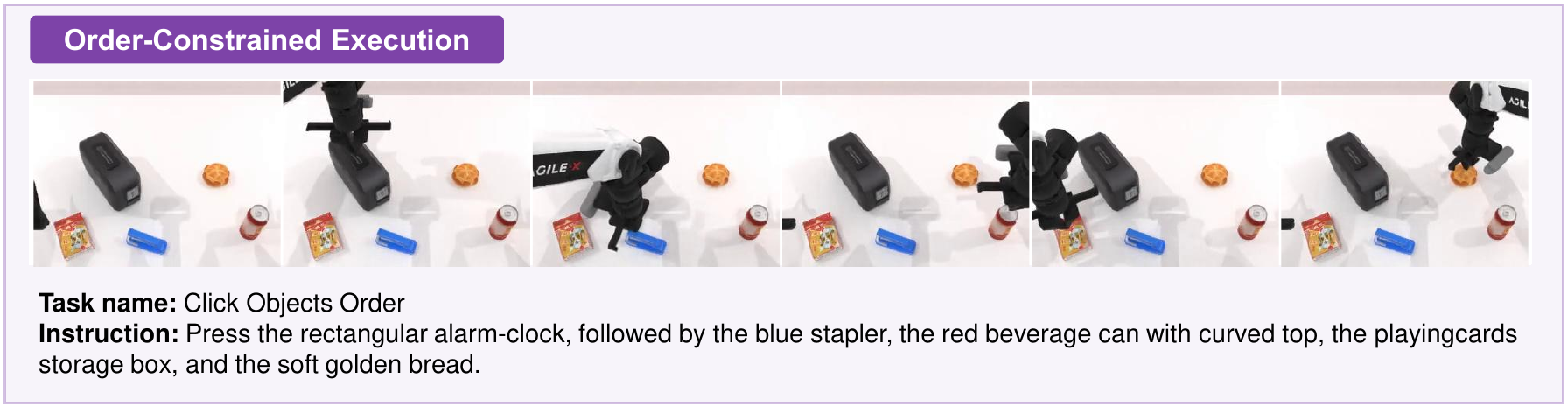}
\vspace{2pt}

\includegraphics[width=\linewidth]{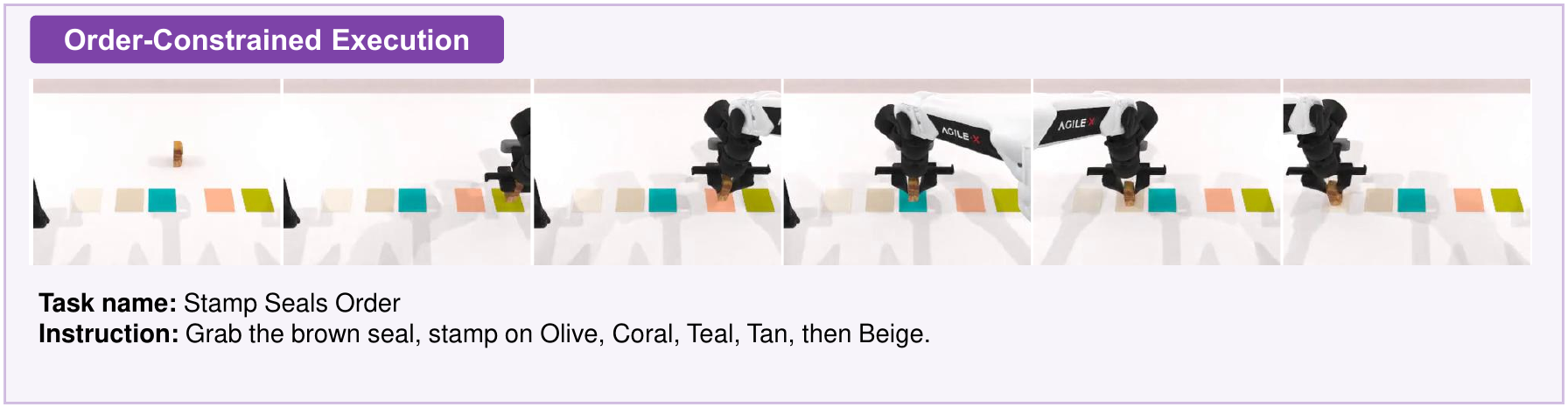}

\caption{Long-horizon procedural planning task examples in \texttt{RoboSPA} (Part 3).}
\label{fig:LH_3}
\end{figure*}

\begin{figure*}[t]
\centering

\includegraphics[width=\linewidth]{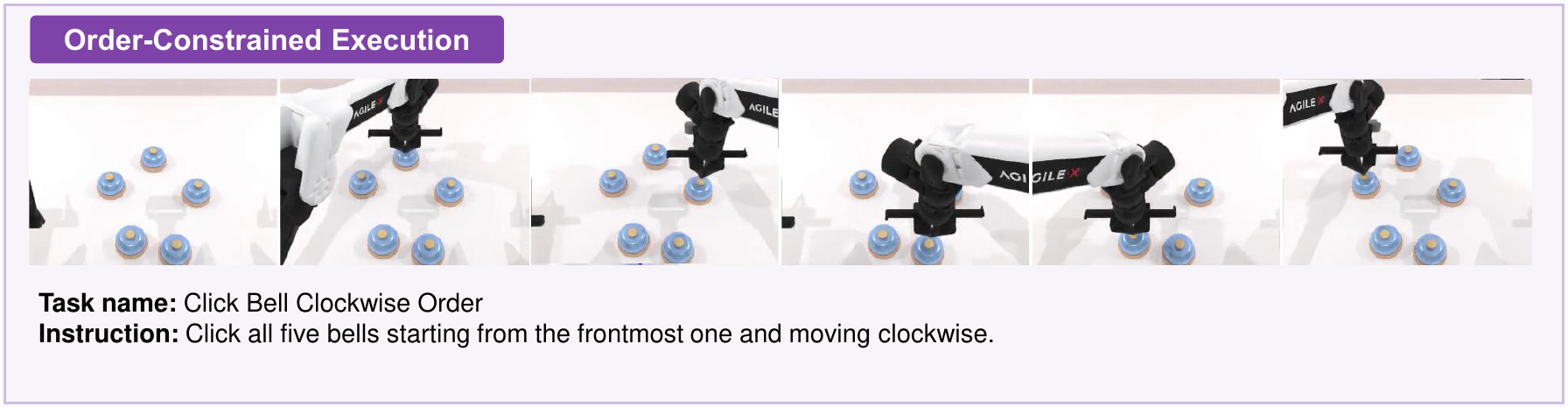}
\vspace{2pt}  

\includegraphics[width=\linewidth]{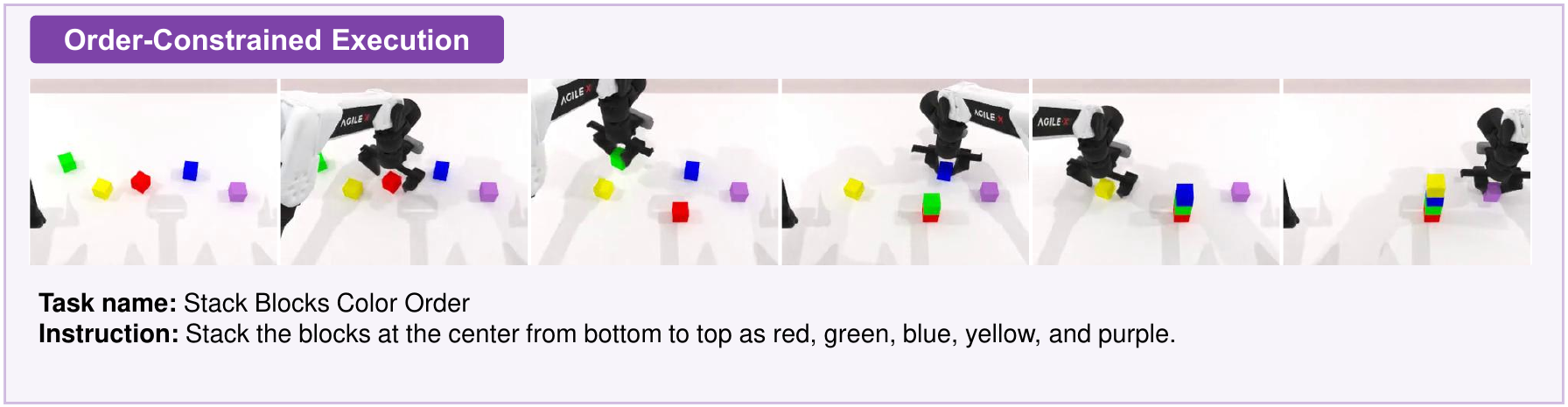}
\vspace{2pt}

\includegraphics[width=\linewidth]{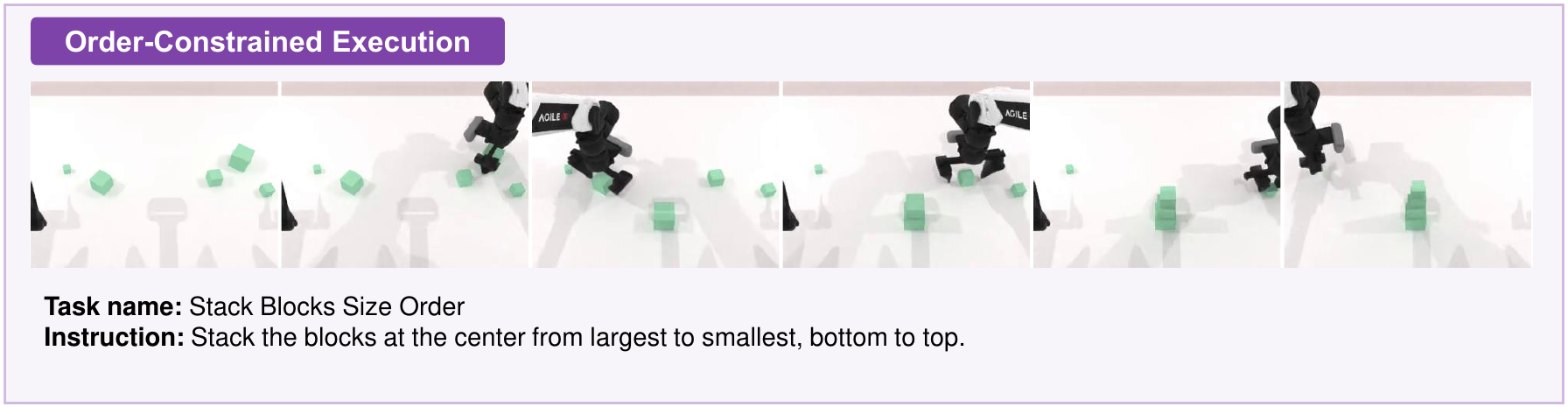}
\vspace{2pt}

\includegraphics[width=\linewidth]{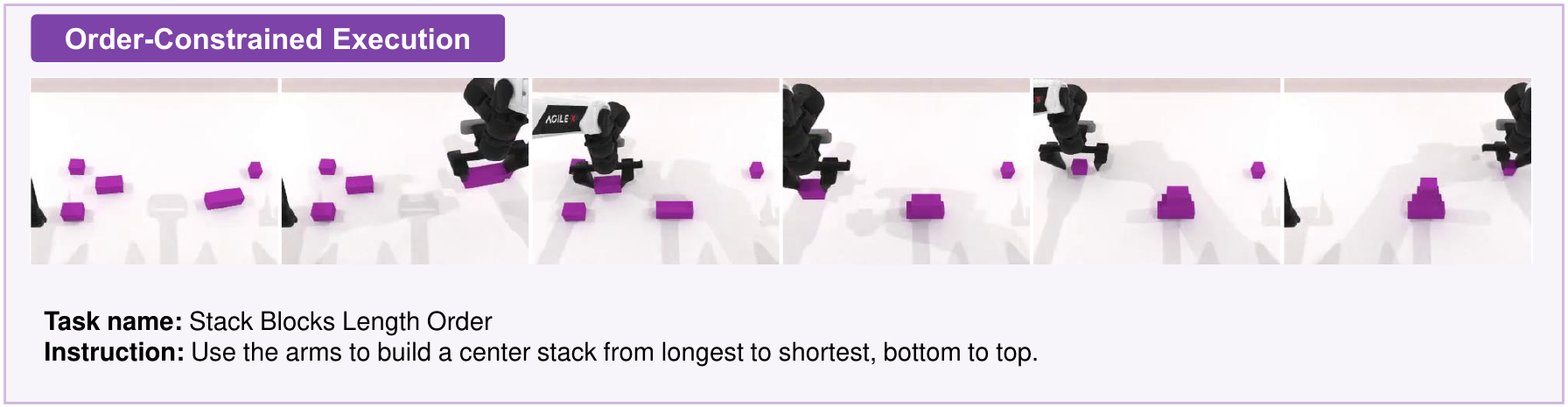}
\vspace{2pt}

\includegraphics[width=\linewidth]{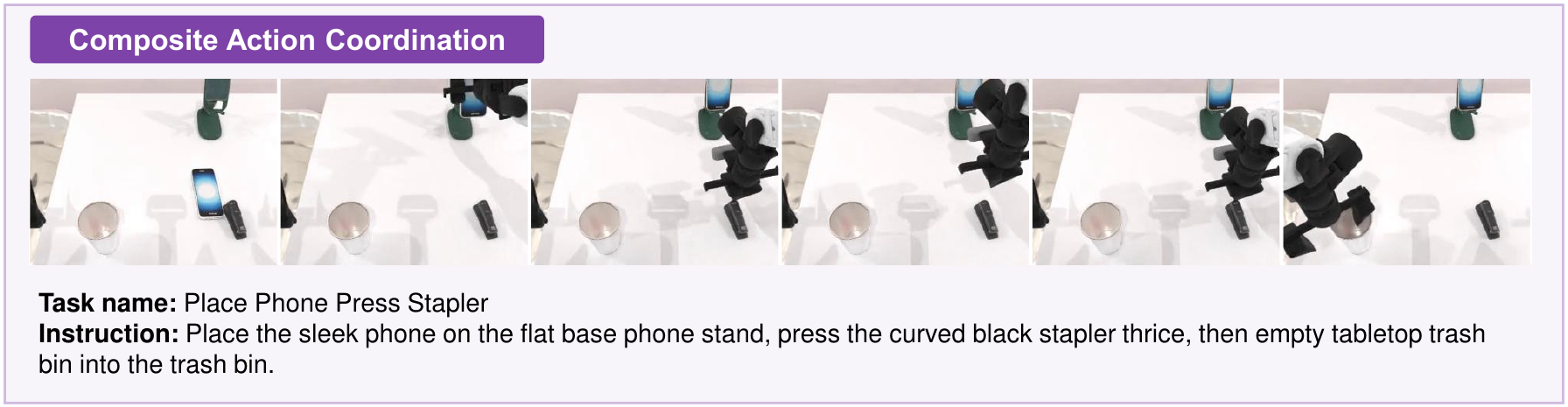}

\caption{Long-horizon procedural planning task examples in \texttt{RoboSPA} (Part 4).}
\label{fig:LH_4}
\end{figure*}

\begin{figure*}[t]
\centering

\includegraphics[width=\linewidth]{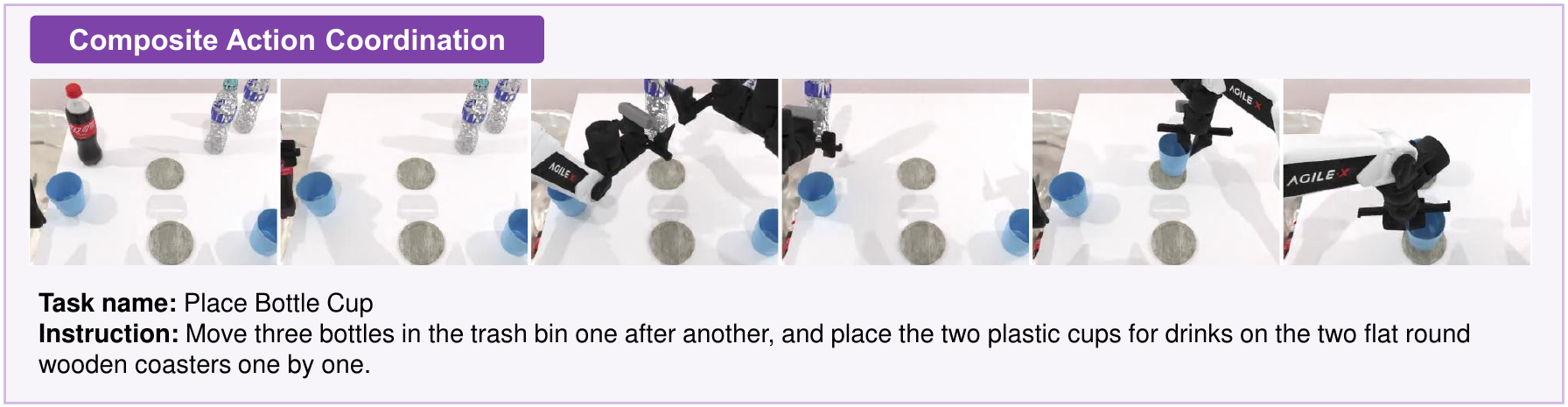}
\vspace{2pt}  

\includegraphics[width=\linewidth]{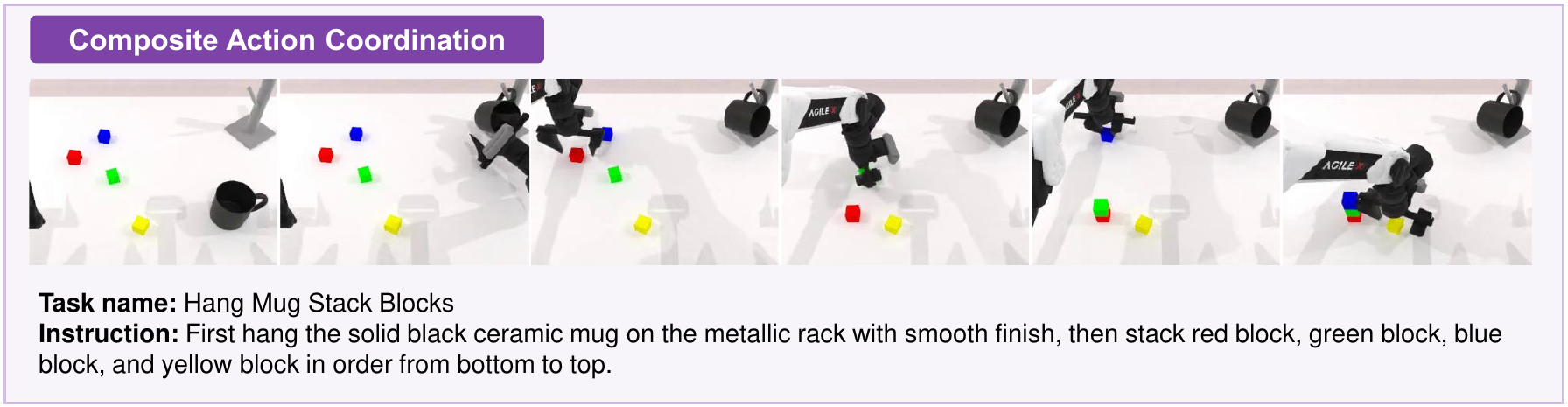}
\vspace{2pt}

\includegraphics[width=\linewidth]{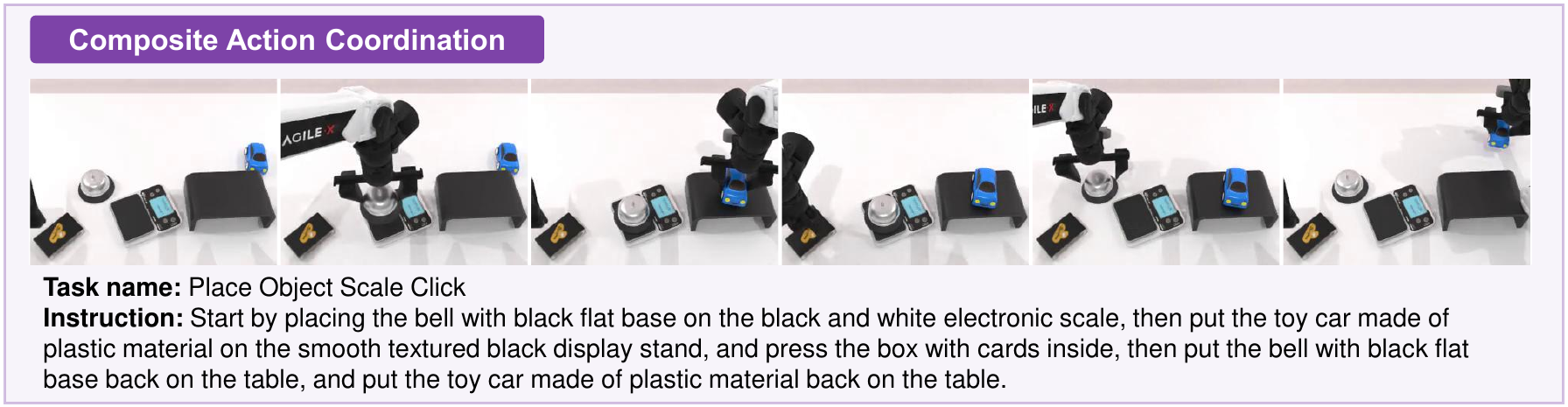}
\vspace{2pt}

\includegraphics[width=\linewidth]{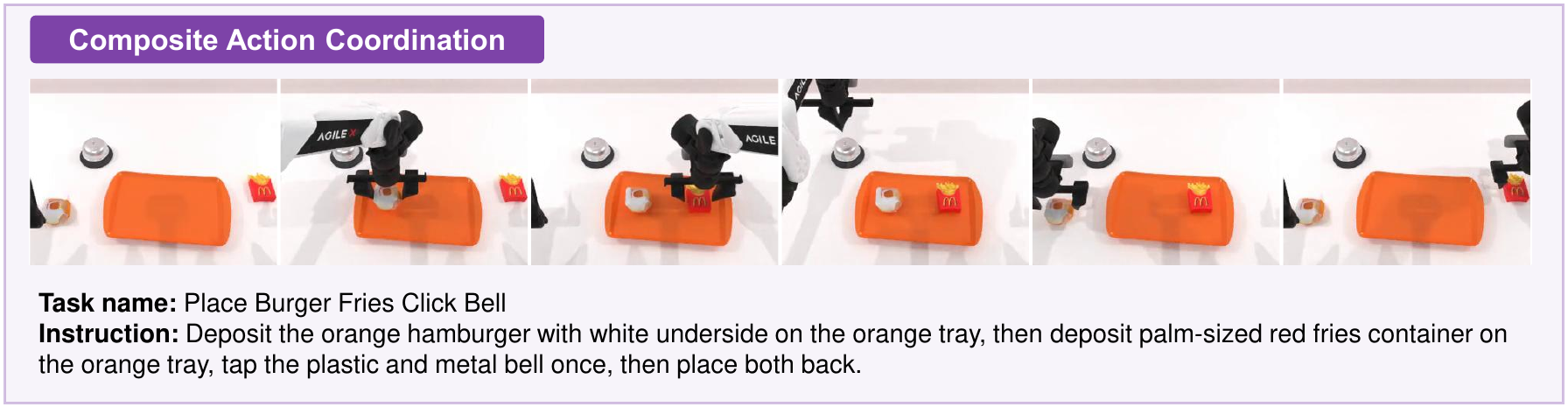}
\vspace{2pt}

\includegraphics[width=\linewidth]{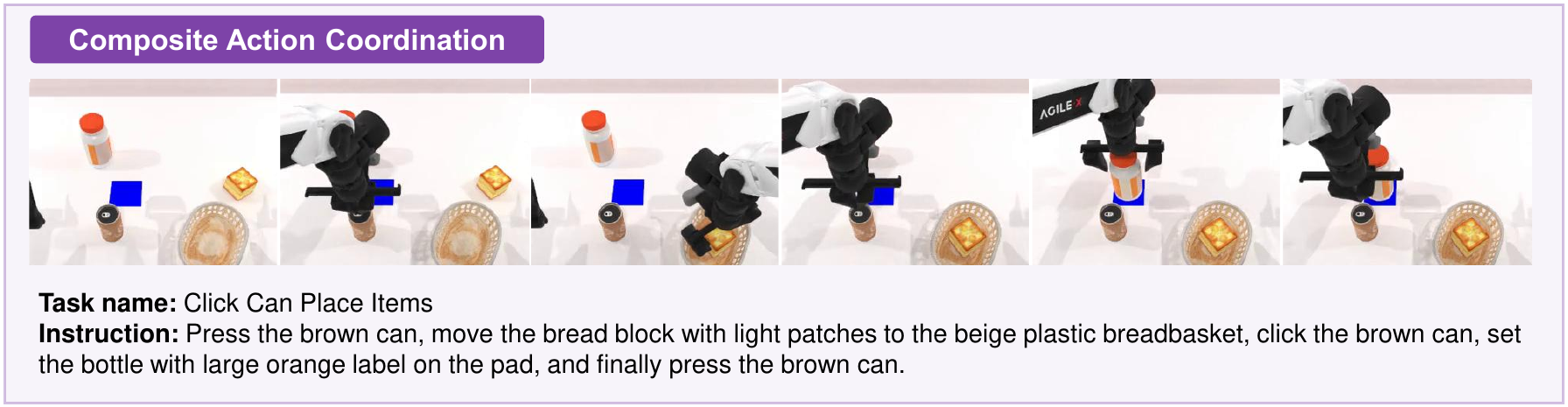}

\caption{Long-horizon procedural planning task examples in \texttt{RoboSPA} (Part 5).}
\label{fig:LH_5}
\end{figure*}

\begin{figure*}[t]
\centering

\includegraphics[width=\linewidth]{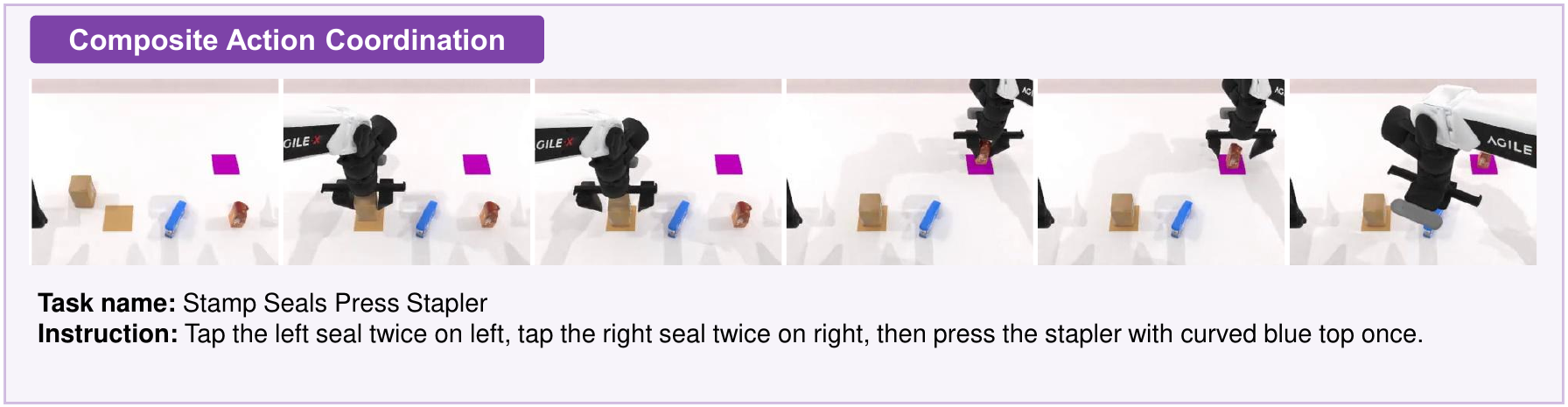}
\vspace{2pt}  

\includegraphics[width=\linewidth]{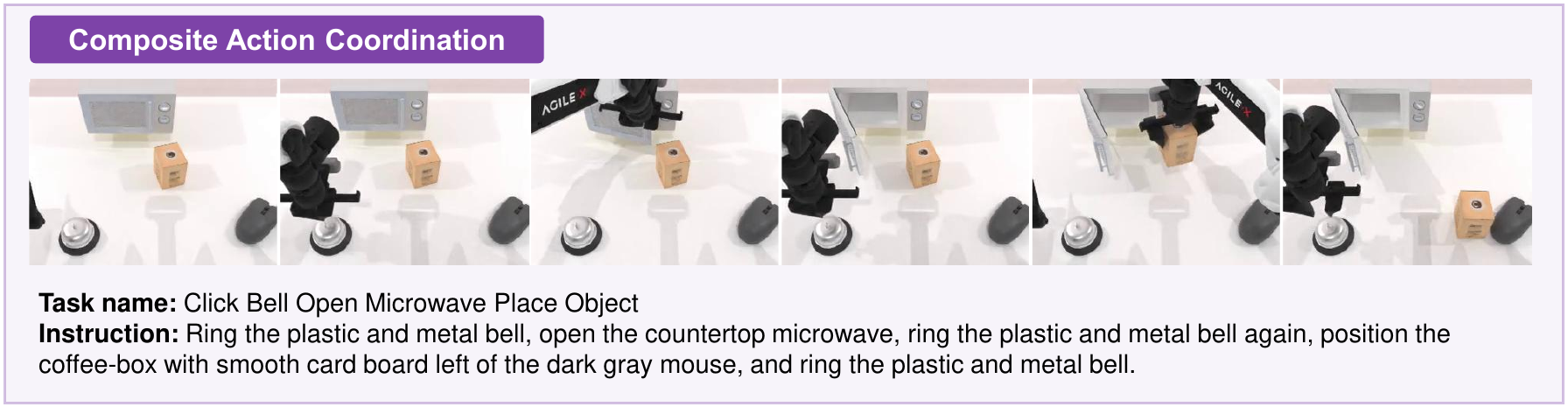}
\vspace{2pt}

\includegraphics[width=\linewidth]{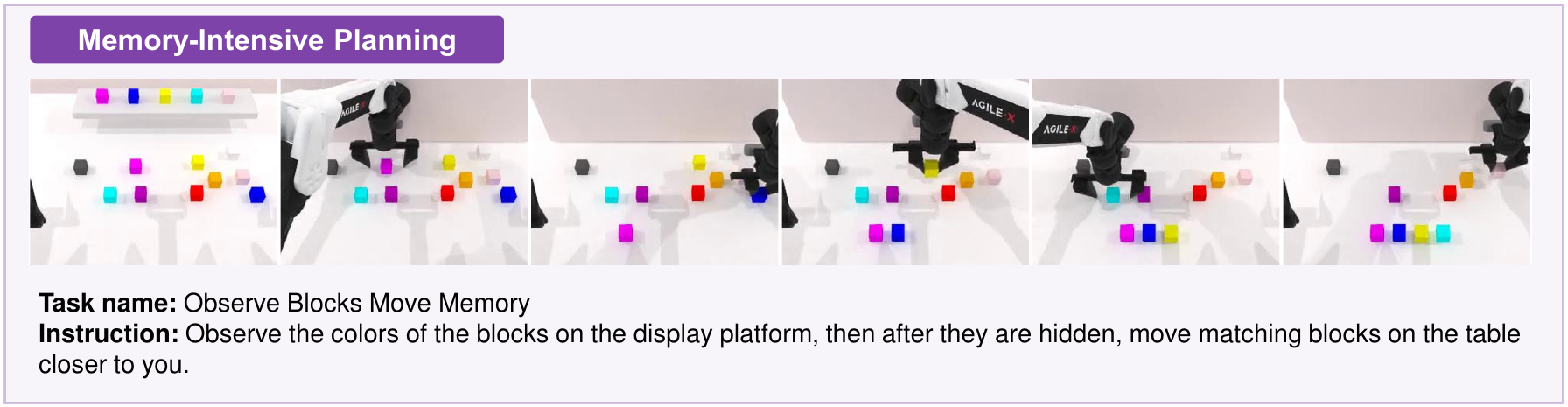}
\vspace{2pt}

\includegraphics[width=\linewidth]{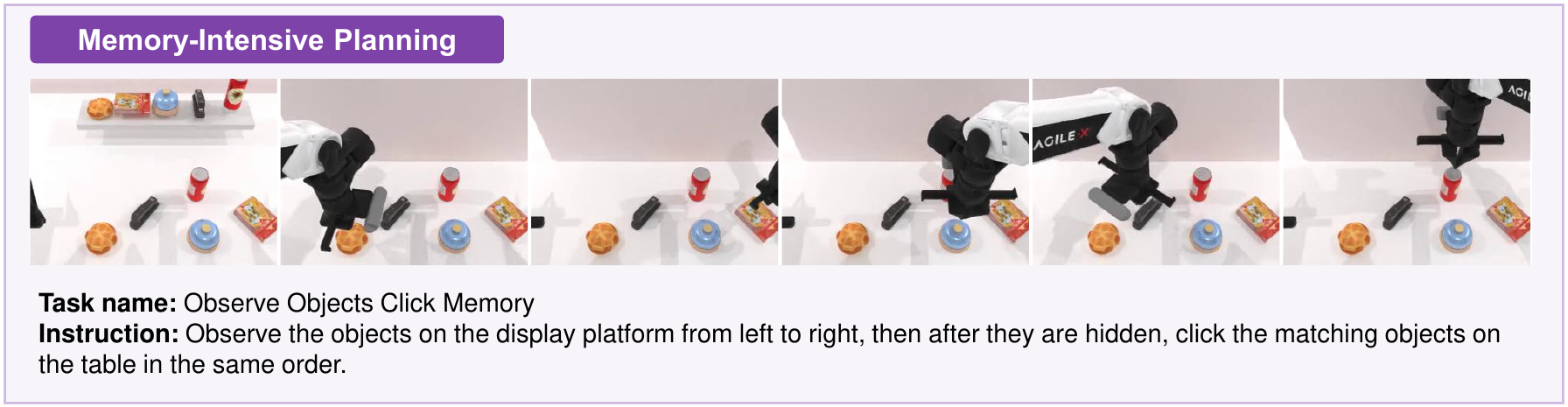}
\vspace{2pt}

\includegraphics[width=\linewidth]{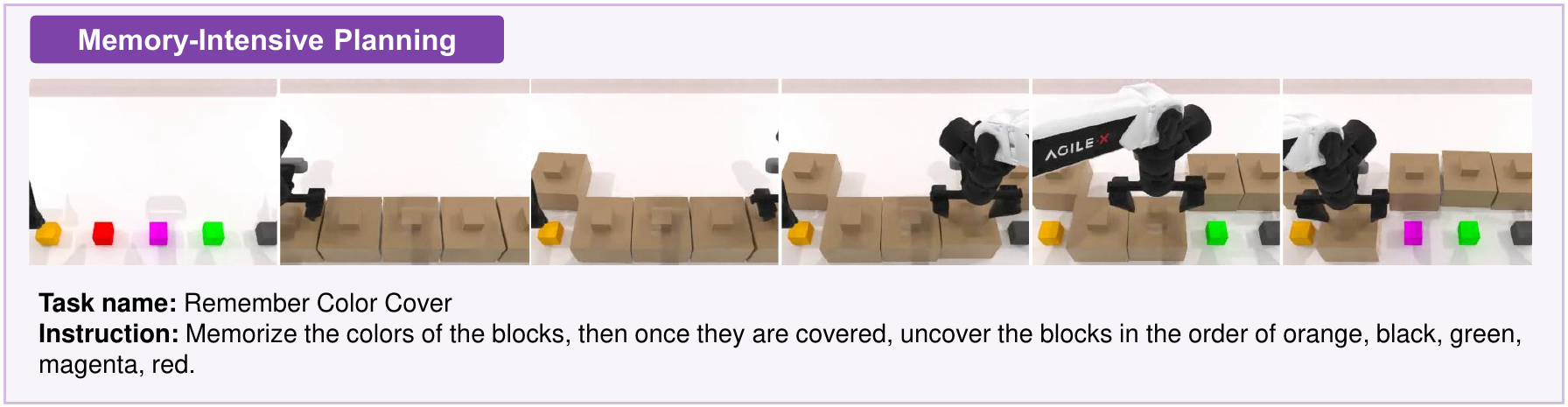}

\caption{Long-horizon procedural planning task examples in \texttt{RoboSPA} (Part 6).}
\label{fig:LH_6}
\end{figure*}

\begin{figure*}[t]
\centering

\includegraphics[width=\linewidth]{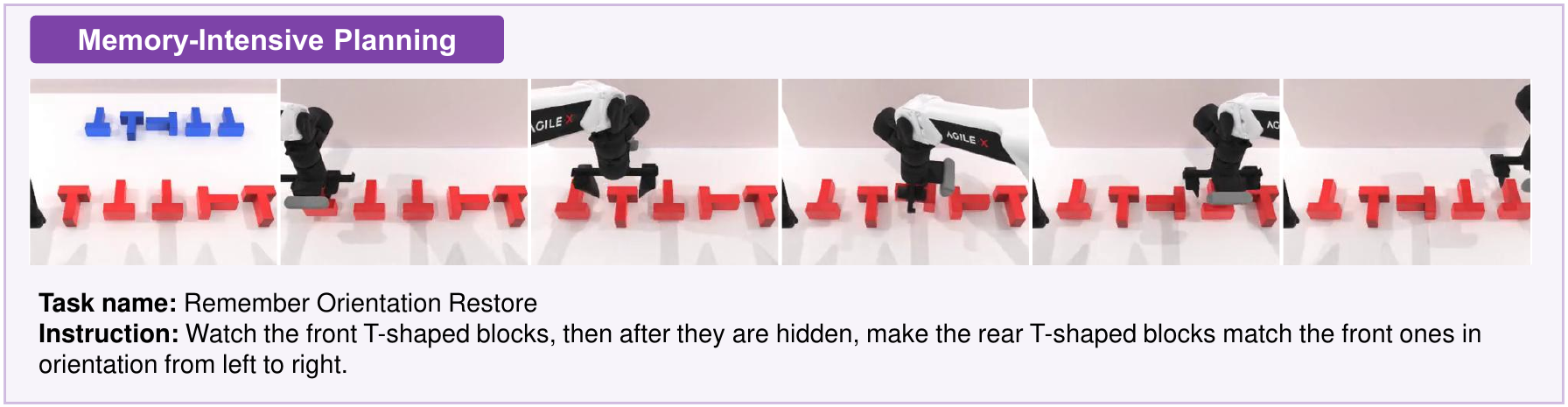}
\vspace{2pt}  

\includegraphics[width=\linewidth]{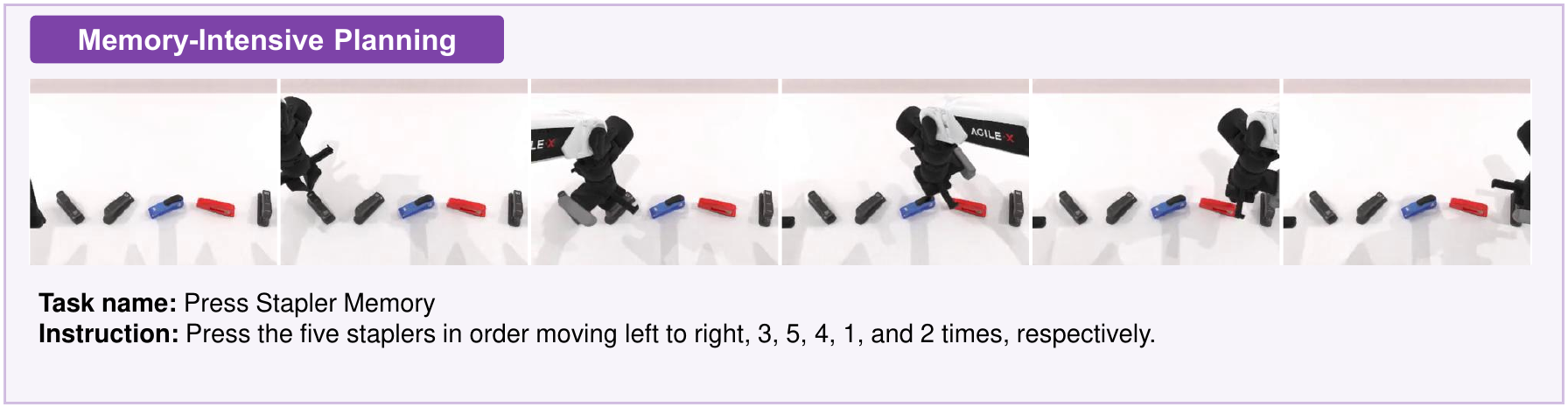}
\vspace{2pt}
\caption{Long-horizon procedural planning task examples in \texttt{RoboSPA} (Part 7).}
\label{fig:LH_7}
\end{figure*}

We provide visualizations for the most difficult variant (L5) of each task in \texttt{RoboSPA}, as shown in Figures~\ref{fig:spatial_reasoning_1}--\ref{fig:LH_7}.
 Each row shows representative rollout frames together with the task category, task name, and instruction, illustrating the scene layout, target objects, and required execution process.
\input{EMNLP/tables/all_tasks}
\input{EMNLP/tables/rebuttal2}

\input{EMNLP/tables/rebuttal}

\input{EMNLP/tables/model_results_SR}

\input{EMNLP/tables/model_results_ONTA}

\input{EMNLP/tables/model_results_PS}

%% file: EMNLP/tables/assets.tex
\begin{table*}[!t]
\centering
\caption{Asset categories and the number of variants used in RoboSPA.}
\label{tab:asset_categories}
\footnotesize
\setlength{\tabcolsep}{4pt}
\renewcommand{\arraystretch}{1.05}
\begin{tabular}{lclclclc}
\toprule
\textbf{Asset} & \textbf{\#Var.} &
\textbf{Asset} & \textbf{\#Var.} &
\textbf{Asset} & \textbf{\#Var.} &
\textbf{Asset} & \textbf{\#Var.} \\
\midrule
bottle & 27 & olive-oil & 5 & pencup & 7 & globe & 2 \\
bowl & 7 & drill & 7 & kitchenpot & 7 & trophy & 5 \\
cover & 1 & jam-jar & 5 & battery & 6 & notebook & 3 \\
plate & 1 & screwdriver & 1 & plasticbox & 11 & brush-pen & 6 \\
fluted-block & 2 & fork & 1 & tabletrashbin & 11 & rest & 4 \\
french-fries & 4 & knife & 1 & msg & 6 & glue & 6 \\
hamburg & 6 & apple & 2 & soy-sauce & 5 & cleaner & 4 \\
T\_block & 2 & cabinet & 1 & vinegar & 3 & screen & 4 \\
shoe-box & 1 & box & 1 & steamer & 3 & speaker & 6 \\
tray & 4 & milk-box & 4 & boxdrink & 4 & fan & 7 \\
kettle & 9 & mug & 13 & vagetable & 7 & seal & 6 \\
pen & 7 & rack & 1 & paymentsign & 4 & milk-tea & 6 \\
dustbin & 1 & shoe & 10 & can & 6 & roller & 3 \\
plant-pot & 5 & wooden\_box & 1 & electronicscale & 5 & fruit & 7 \\
dumbbell-rack & 4 & book & 2 & rubikscube & 3 & board & 5 \\
bookcase & 4 & microwave & 2 & displaystand & 5 & sauce-can & 5 \\
laptop & 11 & sand-clock & 3 & bread & 7 & skillet & 4 \\
oven & 4 & alarm-clock & 6 & breadbasket & 5 & soap & 4 \\
calculator & 6 & mouse & 3 & phone & 5 & block & 7 \\
microphone & 4 & stapler & 7 & phonestand & 7 & hydrating-oil & 4 \\
coaster & 1 & shampoo & 7 & remotecontrol & 7 & basket & 4 \\
hammer & 1 & bell & 2 & pillbottle & 5 & callbell & 6 \\
cup & 13 & candlestick & 5 & playingcards & 3 & tea-box & 6 \\
cup-with-liquid & 1 & dumbbell & 7 & smallshovel & 4 & coffee-box & 7 \\
tissue-box & 7 & teanet & 5 & brush & 4 & perfume & 4 \\
scanner & 5 & baguette & 2 & woodenmallet & 5 & keyboard & 4 \\
chips-tub & 4 & small-speaker & 3 & gong & 6 & whiteboard-eraser & 1 \\
pet-collar & 4 & switch & 8 & woodenblock & 5 & tooth-paste & 1 \\
table-tennis & 2 & toycar & 6 & waterer & 8 & mini-chalkboard & 1 \\
roll-paper & 4 & markpen & 6 & wineglass & 5 & plant & 1 \\
\bottomrule
\end{tabular}
\end{table*}

%% file: EMNLP/tables/model_details.tex
\begin{table}[t]
\centering
\caption{Model size and training compute budget per base task. All experiments were conducted on NVIDIA RTX PRO 6000 GPUs.}
\label{tab:model_size_budget}
\small
\resizebox{\columnwidth}{!}{
\begin{tabular}{lcccc}
\toprule
Model & Params. & \#GPUs & Time & GPU Hours. \\
\midrule
RDT & 1.2B & 1 & 2.5 h & 2.5 \\
GO-1 & 2.8B & 2 & 6 h & 12 \\
$\pi_{0.5}$ & 3.3B & 2 & 6 h & 12 \\
X-VLA & 0.9B & 1 & 1.5 h & 1.5 \\
\bottomrule
\end{tabular}
}
\end{table}

%% file: EMNLP/tables/v1.tex
\begin{table}[ht]
\centering
\caption{Performance on basic spatial task variants.}
\label{tab:basic_spatial_tasks}
\small
\setlength{\tabcolsep}{4pt}
\resizebox{\columnwidth}{!}{
\begin{tabular}{lccccc}
\toprule
\textbf{Task} & \textbf{L1} & \textbf{L2} & \textbf{L3} & \textbf{L4} & \textbf{L5} \\
\midrule
Color-Based Grounding & 80.6 & 77.8 & 79.8 & 71.2 & 74.0 \\
Egocentric Position Grounding & 67.8 & 69.4 & 74.8 & 72.2 & 72.4 \\
Simple Ordinal Indexing & 87.0 & 86.8 & 83.8 & 87.0 & 84.8 \\
\bottomrule
\end{tabular}
}
\end{table}

%% file: EMNLP/tables/all_tasks.tex
\begin{table*}[ht]
\centering
\caption{Task list and descriptions for fine-grained spatial reasoning. Category abbreviations: GAC = Geometric Attribute Cognition, SDE = Spatial Distance Estimation, CPI = Canonical Position Indexing, RRR = Referential Relational Reasoning, and CVR = Cross-View Reasoning.}

\label{tab:spatial_task_list}
\footnotesize
\setlength{\tabcolsep}{3.5pt}
\renewcommand{\arraystretch}{1.08}
\begin{tabularx}{\textwidth}{p{0.07\textwidth} p{0.30\textwidth} X}
\toprule
\textbf{Cat.} & \textbf{Task Name} & \textbf{Task Description} \\
\midrule
GAC & Pick Blocks Size & Pick the target block from multiple blocks based on the size. \\
GAC & Pick Blocks Height & Pick the target block from multiple blocks based on the height. \\
GAC & Pick Blocks Length & Pick the target block from multiple blocks based on the length. \\
GAC & Pick Blocks Area & Pick the target block from multiple blocks based on the base area. \\

SDE & Pick Blocks Distance & Pick the target block from multiple blocks using both color and relative distance cues. \\
SDE & Pick Mugs Distance & Pick the target mug from multiple mugs using both color and relative distance cues. \\
SDE & Pick Pill Bottles Distance & Pick the target pill bottle from multiple pill bottles using both color and relative distance cues. \\
SDE & Pick Mixed Objects Distance A & Pick the target object from mixed-object set A using both color and relative distance cues. \\
SDE & Pick Mixed Objects Distance B & Pick the target object from mixed-object set B using both color and relative distance cues. \\

CPI & Pick Blocks Canonical & Pick the target block according to its ordinal position among multiple blocks. \\
CPI & Pick Cups Canonical & Pick the target cup according to its ordinal position among multiple cups. \\
CPI & Pick Rubik Cubes Canonical & Pick the target Rubik cube according to its ordinal position among multiple Rubik cubes. \\
CPI & Pick Mixed Objects Canonical A & Pick the target object from mixed-object set A according to ordinal position. \\
CPI & Pick Mixed Objects Canonical B & Pick the target object from mixed-object set B according to ordinal position. \\

RRR & Pick Tea Box Relational & Pick the target tea box according to its relative position to another reference tea box. \\
RRR & Pick Cans Relational & Pick the target can according to its relative position to another reference can. \\
RRR & Pick Soaps Relational & Pick the target soap according to its relative position to another reference soap. \\
RRR & Pick Mixed Objects Relational A & Pick the target object in a mixed-object scene according to another reference object. \\
RRR & Pick Mixed Objects Relational B & Pick the target object in a mixed-object scene according to another reference object. \\

CVR & Pick Sauce Can Multi View & Pick up the target sauce can specified by its absolute position under a given viewpoint. \\
CVR & Pick Seals Multi View & Pick up the target seal specified by its absolute position under a given viewpoint. \\
CVR & Pick Breads Multi View & Pick up the target bread specified by its absolute position under a given viewpoint. \\
CVR & Pick Mixed Objects Multi View A & Pick up the target object from mixed-object group A specified by its absolute position under a given viewpoint. \\
CVR & Pick Mixed Objects Multi View B & Pick up the target object from mixed-object group B specified by its absolute position under a given viewpoint. \\

\bottomrule
\end{tabularx}
\end{table*}


\begin{table*}[!t]
\centering
\caption{Task list and descriptions for long-horizon procedural planning. Category abbreviations: RPF = Repetitive Procedure Following, OFE = Order-Free Execution, OCE = Order-Constrained Execution, CAC = Composite Action Coordination, and MIP = Memory-Intensive Planning.}
\label{tab:long_horizon_task_list}
\footnotesize
\setlength{\tabcolsep}{3.5pt}
\renewcommand{\arraystretch}{1.08}
\begin{tabularx}{\textwidth}{p{0.07\textwidth} p{0.30\textwidth} X}
\toprule
\textbf{Cat.} & \textbf{Task Name} & \textbf{Task Description} \\
\midrule

RPF & Press Stapler Repeat & Press the stapler repeatedly. \\
RPF & Lift Pot Repeat & Lift and put down the pot repeatedly. \\
RPF & Lift Fan Repeat & Lift the fan repeatedly. \\
RPF & Click Bell Repeat & Click the bell repeatedly. \\

OFE & Put Bottles Dustbin & Put multiple bottles into the dustbin without requiring a fixed execution order. \\
OFE & Move Blocks Apart & Move multiple blocks apart without requiring a fixed execution order. \\
OFE & Move Playing Cards Away & Move multiple playing cards away without requiring a fixed execution order. \\
OFE & Place Bowls Plates & Place multiple bowls on plates without requiring a fixed execution order. \\
OFE & Separate Fries Bread & Separate fries and bread into different areas without requiring a fixed execution order. \\
OFE & Rank Blocks Height & Arrange multiple blocks by height while allowing flexible intermediate execution order. \\
OFE & Rank Blocks Color & Arrange multiple blocks by color while allowing flexible intermediate execution order. \\
OFE & Rank Blocks Size & Arrange multiple blocks by size while allowing flexible intermediate execution order. \\

OCE & Hit Blocks Hammer Order & Hit multiple blocks with a hammer in the specified color order. \\
OCE & Click Objects Order & Click multiple target objects in the specified order. \\
OCE & Stamp Seals Order & Stamp multiple seals in the specified order. \\
OCE & Click Bell Clockwise Order & Click the bell and then execute the clockwise movement sequence in order. \\
OCE & Stack Blocks Color Order & Stack multiple blocks in the specified color order. \\
OCE & Stack Blocks Size Order & Stack multiple blocks in the specified size order. \\
OCE & Stack Blocks Length Order & Stack multiple blocks in the specified length order. \\

CAC & Place Phone Press Stapler & Execute a manipulation task involving phone placement, repeated pressing, and dumping. \\
CAC & Place Bottle Cup & Execute a manipulation task involving throwing bottles into a dustbin and placing cups on coasters. \\
CAC & Hang Mug Stack Blocks & Execute a manipulation task involving hanging a mug on a rack and stacking blocks from bottom to top. \\
CAC & Place Object Scale Click & Execute a manipulation task involving object placement, pressing, and returning objects to the table. \\
CAC & Place Burger Fries Click Bell & Execute a manipulation task involving placing a hamburger and fries on a tray, clicking a bell, and returning them. \\
CAC & Click Can Place Items & Execute a manipulation task involving interleaved can clicking with placing bread in a basket and a pill bottle on a pad. \\
CAC & Stamp Seals Press Stapler & Execute a manipulation task involving stamping left and right seals twice on their pads and pressing a stapler. \\
CAC & Click Bell Open Microwave Place Object & Execute a manipulation task involving interleaved bell clicking, opening a microwave, and placing an object left of another. \\

MIP & Observe Blocks Move Memory & First remember the colors of the display blocks, then move matching blocks on the table closer after the display blocks are hidden. \\
MIP & Observe Objects Click Memory & First remember the objects on the display platform from left to right, then click matching objects on the table in the same order after they are hidden. \\
MIP & Remember Color Cover & First remember the block colors, then uncover the blocks in the given color order after they are covered. \\
MIP & Remember Orientation Restore & First remember the orientations of the front T-shaped blocks, then adjust the rear T-shaped blocks to match them from left to right after they are hidden. \\
MIP & Press Stapler Memory & Remember how many times each stapler should be pressed and then press them. \\

\bottomrule
\end{tabularx}
\end{table*}

%% file: EMNLP/tables/rebuttal2.tex
\begin{table*}[t]
\centering
\caption{Bootstrap 95\% confidence intervals for the average L5 \textit{Success Rates} (\%).}
\label{tab:statistical_uncertainty}
\small
\begin{tabular*}{\textwidth}{@{\extracolsep{\fill}}lccc@{}}
\toprule
\textbf{Model}
& \textbf{Fine-Grained Spatial Reasoning Avg.}
& \textbf{Long-Horizon Procedural Planning Avg.}
& \textbf{Overall Avg.} \\
\midrule
RDT
& 6.1 [5.2, 7.1]
& 7.7 [6.8, 8.5]
& 6.9 [6.3, 7.5] \\

GO-1
& 10.6 [9.4, 11.8]
& 7.0 [6.2, 7.7]
& 8.8 [8.1, 9.5] \\

$\pi_{0.5}$
& 21.4 [19.8, 22.9]
& 23.1 [21.9, 24.3]
& 22.3 [21.3, 23.2] \\

X-VLA
& 23.9 [22.3, 25.5]
& 15.9 [14.8, 17.0]
& 19.9 [18.9, 20.8] \\
\bottomrule
\end{tabular*}
\end{table*}

%% file: EMNLP/tables/rebuttal.tex
\begin{table*}[t]
\centering
\caption{Multi-embodiment evaluation of X-VLA on Fine-Grained Spatial Reasoning tasks.}
\label{tab:multi_embodiment_fgsr}
\small
\setlength{\tabcolsep}{4pt}
\begin{tabular}{l*{5}{ccc}}
\toprule
\textbf{Embodiment}
& \multicolumn{3}{c}{\textbf{GAC-1}}
& \multicolumn{3}{c}{\textbf{SDE-5}}
& \multicolumn{3}{c}{\textbf{CPI-4}}
& \multicolumn{3}{c}{\textbf{RRR-3}}
& \multicolumn{3}{c}{\textbf{CVR-4}} \\
\cmidrule(lr){2-4}
\cmidrule(lr){5-7}
\cmidrule(lr){8-10}
\cmidrule(lr){11-13}
\cmidrule(lr){14-16}
& L1 & L3 & L5
& L1 & L3 & L5
& L1 & L3 & L5
& L1 & L3 & L5
& L1 & L3 & L5 \\
\midrule
Aloha-AgileX
& 48 & 22 & 16
& 48 & 31 & 27
& 29 & 13 & 7
& 43 & 31 & 14
& 41 & 30 & 22 \\

ARX-X5
& 51 & 18 & 11
& 36 & 29 & 26
& 32 & 14 & 7
& 50 & 24 & 13
& 28 & 11 & 13 \\

Piper
& 48 & 34 & 16
& 18 & 20 & 17
& 40 & 18 & 10
& 40 & 18 & 20
& 50 & 44 & 39 \\

Franka
& 22 & 12 & 8
& 14 & 10 & 10
& 38 & 18 & 8
& 48 & 18 & 14
& 12 & 8 & 7 \\

UR5
& 74 & 43 & 25
& 52 & 41 & 34
& 67 & 35 & 26
& 58 & 45 & 38
& 59 & 55 & 49 \\
\bottomrule
\end{tabular}
\end{table*}

\begin{table*}[t]
\centering
\caption{Multi-embodiment evaluation of X-VLA on Long-Horizon Procedural Planning tasks.}
\label{tab:multi_embodiment_lhpp}
\small
\setlength{\tabcolsep}{4pt}
\begin{tabular}{l*{5}{ccc}}
\toprule
\textbf{Embodiment}
& \multicolumn{3}{c}{\textbf{RPF-1}}
& \multicolumn{3}{c}{\textbf{OFE-1}}
& \multicolumn{3}{c}{\textbf{OCE-5}}
& \multicolumn{3}{c}{\textbf{CAC-6}}
& \multicolumn{3}{c}{\textbf{MIP-3}} \\
\cmidrule(lr){2-4}
\cmidrule(lr){5-7}
\cmidrule(lr){8-10}
\cmidrule(lr){11-13}
\cmidrule(lr){14-16}
& L1 & L3 & L5
& L1 & L3 & L5
& L1 & L3 & L5
& L1 & L3 & L5
& L1 & L3 & L5 \\
\midrule
Aloha-AgileX
& 49 & 30 & 27
& 79 & 55 & 28
& 99 & 67 & 18
& 54 & 37 & 24
& 100 & 8 & 0 \\

ARX-X5
& 54 & 46 & 40
& 64 & 59 & 27
& 100 & 54 & 22
& 99 & 53 & 29
& 99 & 12 & 0 \\

Piper
& 74 & 66 & 40
& 96 & 9 & 0
& 100 & 16 & 9
& 52 & 31 & 19
& 99 & 0 & 0 \\

Franka
& 66 & 40 & 24
& 26 & 20 & 11
& 92 & 18 & 0
& 98 & 40 & 0
& 100 & 0 & 0 \\

UR5
& 83 & 58 & 49
& 99 & 24 & 5
& 93 & 33 & 17
& 100 & 51 & 48
& 100 & 16 & 0 \\
\bottomrule
\end{tabular}
\end{table*}

%% file: EMNLP/tables/model_results_SR.tex
\begin{table*}[!t]
\centering
\caption{RDT per-task \textit{Success Rate} across difficulty levels.}
\label{tab:appendix-rdt-success}
\scriptsize
\setlength{\tabcolsep}{2.2pt}
\renewcommand{\arraystretch}{0.9}
\resizebox{\textwidth}{!}{%
\begin{tabular}{@{}lrrrrr@{\hspace{1.2em}}lrrrrr@{}}
\toprule
Task & L1 & L2 & L3 & L4 & L5 & Task & L1 & L2 & L3 & L4 & L5 \\
\midrule
Pick Blocks Size & 3 & 1 & 3 & 6 & 1 & Put Bottles Dustbin & 51 & 37 & 22 & 8 & 2 \\
Pick Blocks Height & 28 & 21 & 20 & 15 & 12 & Move Blocks Apart & 6 & 2 & 1 & 0 & 0 \\
Pick Blocks Length & 6 & 1 & 1 & 4 & 3 & Move Playing Cards Away & 8 & 5 & 0 & 0 & 0 \\
Pick Blocks Area & 3 & 1 & 1 & 5 & 3 & Place Bowls Plates & 48 & 15 & 12 & 6 & 2 \\
Pick Blocks Distance & 1 & 6 & 1 & 3 & 4 & Separate Fries Bread & 24 & 9 & 6 & 1 & 2 \\
Pick Mugs Distance & 15 & 17 & 12 & 13 & 10 & Rank Blocks Height & 6 & 1 & 0 & 1 & 0 \\
Pick Pill Bottles Distance & 25 & 15 & 14 & 13 & 13 & Rank Blocks Color & 14 & 6 & 1 & 0 & 0 \\
Pick Mixed Objects Distance A & 15 & 16 & 14 & 6 & 10 & Rank Blocks Size & 5 & 0 & 0 & 0 & 0 \\
Pick Mixed Objects Distance B & 20 & 13 & 14 & 13 & 10 & Hit Blocks Hammer Order & 40 & 4 & 0 & 0 & 0 \\
Pick Blocks Canonical & 3 & 5 & 4 & 1 & 5 & Click Objects Order & 20 & 0 & 0 & 0 & 0 \\
Pick Cups Canonical & 35 & 18 & 21 & 17 & 22 & Stamp Seals Order & 3 & 0 & 1 & 1 & 1 \\
Pick Rubik Cubes Canonical & 7 & 6 & 6 & 7 & 7 & Click Bell Clockwise Order & 5 & 0 & 0 & 0 & 0 \\
Pick Mixed Objects Canonical A & 11 & 4 & 5 & 1 & 5 & Stack Blocks Color Order & 11 & 0 & 0 & 0 & 0 \\
Pick Mixed Objects Canonical B & 19 & 6 & 5 & 6 & 4 & Stack Blocks Size Order & 9 & 0 & 0 & 0 & 0 \\
Pick Tea Box Referential & 7 & 2 & 5 & 6 & 3 & Stack Blocks Length Order & 7 & 0 & 0 & 0 & 0 \\
Pick Cans Referential & 2 & 1 & 7 & 5 & 6 & Place Phone Press Stapler & 18 & 6 & 1 & 0 & 0 \\
Pick Soaps Referential & 8 & 12 & 9 & 5 & 7 & Place Bottle Cup & 56 & 29 & 26 & 6 & 0 \\
Pick Mixed Objects Referential A & 2 & 2 & 7 & 7 & 4 & Hang Mug Stack Blocks & 10 & 0 & 0 & 0 & 0 \\
Pick Mixed Objects Referential B & 3 & 2 & 3 & 4 & 2 & Place Object Scale Click & 39 & 33 & 26 & 9 & 17 \\
Pick Sauce Can Multi View & 3 & 5 & 4 & 3 & 7 & Place Burger Fries Click Bell & 33 & 20 & 4 & 0 & 0 \\
Pick Seals Multi View & 3 & 5 & 5 & 4 & 4 & Click Can Place Items & 62 & 24 & 34 & 0 & 2 \\
Pick Breads Multi View & 3 & 5 & 2 & 3 & 0 & Stamp Seals Press Stapler & 2 & 3 & 0 & 0 & 0 \\
Pick Mixed Objects Multi View A & 1 & 5 & 2 & 3 & 4 & Click Open Place & 0 & 1 & 2 & 0 & 0 \\
Pick Mixed Objects Multi View B & 5 & 1 & 4 & 3 & 2 & Observe Blocks Move Memory & 0 & 0 & 0 & 0 & 0 \\
Press Stapler Repeat & 33 & 37 & 37 & 34 & 32 & Observe Objects Click Memory & 14 & 0 & 0 & 0 & 0 \\
Lift Pot Repeat & 76 & 68 & 63 & 56 & 68 & Remember Color Cover & 47 & 14 & 6 & 0 & 0 \\
Lift Fan Repeat & 39 & 33 & 26 & 9 & 17 & Remember Orientation Restore & 0 & 0 & 0 & 0 & 0 \\
Click Bell Repeat & 15 & 17 & 22 & 22 & 23 & Press Stapler Memory & 30 & 3 & 0 & 0 & 0 \\
\bottomrule
\end{tabular}%
}
\end{table*}

\clearpage
\begin{table*}[!t]
\centering
\caption{GO-1 per-task \textit{Success Rate} across difficulty levels.}
\label{tab:appendix-go1-success}
\scriptsize
\setlength{\tabcolsep}{2.2pt}
\renewcommand{\arraystretch}{0.9}
\resizebox{\textwidth}{!}{%
\begin{tabular}{@{}lrrrrr@{\hspace{1.2em}}lrrrrr@{}}
\toprule
Task & L1 & L2 & L3 & L4 & L5 & Task & L1 & L2 & L3 & L4 & L5 \\
\midrule
Pick Blocks Size & 33 & 13 & 12 & 10 & 9 & Put Bottles Dustbin & 59 & 77 & 42 & 33 & 11 \\
Pick Blocks Height & 71 & 62 & 50 & 47 & 32 & Move Blocks Apart & 21 & 10 & 3 & 1 & 0 \\
Pick Blocks Length & 3 & 11 & 15 & 7 & 5 & Move Playing Cards Away & 52 & 39 & 33 & 13 & 17 \\
Pick Blocks Area & 4 & 3 & 2 & 0 & 0 & Place Bowls Plates & 37 & 23 & 21 & 13 & 5 \\
Pick Blocks Distance & 3 & 2 & 6 & 4 & 3 & Separate Fries Bread & 44 & 32 & 19 & 21 & 7 \\
Pick Mugs Distance & 15 & 14 & 14 & 8 & 9 & Rank Blocks Height & 34 & 3 & 0 & 0 & 0 \\
Pick Pill Bottles Distance & 36 & 8 & 15 & 15 & 26 & Rank Blocks Color & 41 & 14 & 8 & 5 & 2 \\
Pick Mixed Objects Distance A & 26 & 19 & 16 & 19 & 12 & Rank Blocks Size & 41 & 22 & 13 & 9 & 3 \\
Pick Mixed Objects Distance B & 21 & 17 & 14 & 8 & 8 & Hit Blocks Hammer Order & 36 & 11 & 3 & 3 & 1 \\
Pick Blocks Canonical & 16 & 11 & 8 & 6 & 7 & Click Objects Order & 62 & 11 & 6 & 4 & 3 \\
Pick Cups Canonical & 24 & 17 & 13 & 8 & 7 & Stamp Seals Order & 12 & 15 & 10 & 9 & 6 \\
Pick Rubik Cubes Canonical & 26 & 17 & 14 & 13 & 15 & Click Bell Clockwise Order & 61 & 48 & 34 & 21 & 12 \\
Pick Mixed Objects Canonical A & 21 & 12 & 6 & 6 & 4 & Stack Blocks Color Order & 54 & 12 & 3 & 1 & 0 \\
Pick Mixed Objects Canonical B & 23 & 15 & 7 & 4 & 3 & Stack Blocks Size Order & 38 & 11 & 1 & 0 & 0 \\
Pick Tea Box Referential & 21 & 17 & 18 & 9 & 7 & Stack Blocks Length Order & 30 & 7 & 5 & 0 & 0 \\
Pick Cans Referential & 16 & 12 & 4 & 14 & 8 & Place Phone Press Stapler & 48 & 14 & 14 & 5 & 1 \\
Pick Soaps Referential & 12 & 8 & 4 & 3 & 3 & Place Bottle Cup & 16 & 12 & 34 & 23 & 4 \\
Pick Mixed Objects Referential A & 20 & 18 & 13 & 15 & 9 & Hang Mug Stack Blocks & 8 & 2 & 0 & 0 & 0 \\
Pick Mixed Objects Referential B & 17 & 15 & 15 & 14 & 6 & Place Object Scale Click & 12 & 14 & 10 & 2 & 3 \\
Pick Sauce Can Multi View & 28 & 33 & 31 & 29 & 27 & Place Burger Fries Click Bell & 37 & 29 & 11 & 3 & 5 \\
Pick Seals Multi View & 18 & 15 & 14 & 12 & 11 & Click Can Place Items & 77 & 11 & 14 & 5 & 3 \\
Pick Breads Multi View & 21 & 12 & 9 & 27 & 15 & Stamp Seals Press Stapler & 24 & 17 & 2 & 0 & 0 \\
Pick Mixed Objects Multi View A & 16 & 8 & 4 & 18 & 18 & Click Open Place & 12 & 13 & 7 & 0 & 2 \\
Pick Mixed Objects Multi View B & 4 & 4 & 12 & 12 & 10 & Observe Blocks Move Memory & 3 & 0 & 0 & 0 & 0 \\
Press Stapler Repeat & 27 & 21 & 12 & 8 & 7 & Observe Objects Click Memory & 19 & 1 & 0 & 0 & 0 \\
Lift Pot Repeat & 61 & 53 & 83 & 84 & 77 & Remember Color Cover & 5 & 0 & 0 & 0 & 0 \\
Lift Fan Repeat & 23 & 17 & 17 & 21 & 11 & Remember Orientation Restore & 2 & 0 & 0 & 0 & 0 \\
Click Bell Repeat & 16 & 12 & 3 & 1 & 0 & Press Stapler Memory & 9 & 2 & 0 & 0 & 0 \\
\bottomrule
\end{tabular}%
}
\end{table*}

\vspace{0.75em}

\begin{table*}[!t]
\centering
\caption{$\pi_{0.5}$ per-task \textit{Success Rate} across difficulty levels.}
\label{tab:appendix-pi05-success}
\scriptsize
\setlength{\tabcolsep}{2.2pt}
\renewcommand{\arraystretch}{0.9}
\resizebox{\textwidth}{!}{%
\begin{tabular}{@{}lrrrrr@{\hspace{1.2em}}lrrrrr@{}}
\toprule
Task & L1 & L2 & L3 & L4 & L5 & Task & L1 & L2 & L3 & L4 & L5 \\
\midrule
Pick Blocks Size & 58 & 27 & 20 & 17 & 13 & Put Bottles Dustbin & 95 & 85 & 81 & 74 & 61 \\
Pick Blocks Height & 67 & 54 & 48 & 30 & 24 & Move Blocks Apart & 73 & 51 & 24 & 35 & 18 \\
Pick Blocks Length & 22 & 5 & 11 & 4 & 3 & Move Playing Cards Away & 81 & 64 & 54 & 52 & 47 \\
Pick Blocks Area & 24 & 13 & 25 & 12 & 10 & Place Bowls Plates & 94 & 85 & 61 & 52 & 22 \\
Pick Blocks Distance & 16 & 16 & 16 & 10 & 11 & Separate Fries Bread & 92 & 84 & 58 & 59 & 36 \\
Pick Mugs Distance & 39 & 51 & 52 & 49 & 38 & Rank Blocks Height & 72 & 45 & 19 & 24 & 6 \\
Pick Pill Bottles Distance & 46 & 36 & 31 & 33 & 33 & Rank Blocks Color & 88 & 50 & 46 & 36 & 11 \\
Pick Mixed Objects Distance A & 33 & 37 & 27 & 29 & 23 & Rank Blocks Size & 91 & 66 & 37 & 15 & 8 \\
Pick Mixed Objects Distance B & 41 & 29 & 21 & 22 & 12 & Hit Blocks Hammer Order & 81 & 30 & 14 & 7 & 3 \\
Pick Blocks Canonical & 41 & 22 & 14 & 10 & 11 & Click Objects Order & 60 & 29 & 16 & 13 & 15 \\
Pick Cups Canonical & 61 & 37 & 40 & 41 & 45 & Stamp Seals Order & 40 & 40 & 30 & 36 & 35 \\
Pick Rubik Cubes Canonical & 51 & 35 & 33 & 36 & 44 & Click Bell Clockwise Order & 98 & 95 & 45 & 31 & 28 \\
Pick Mixed Objects Canonical A & 35 & 15 & 10 & 11 & 11 & Stack Blocks Color Order & 91 & 58 & 25 & 5 & 1 \\
Pick Mixed Objects Canonical B & 32 & 15 & 10 & 10 & 7 & Stack Blocks Size Order & 89 & 43 & 4 & 3 & 0 \\
Pick Tea Box Referential & 33 & 31 & 31 & 12 & 15 & Stack Blocks Length Order & 83 & 27 & 18 & 3 & 2 \\
Pick Cans Referential & 27 & 21 & 28 & 17 & 8 & Place Phone Press Stapler & 46 & 14 & 17 & 12 & 1 \\
Pick Soaps Referential & 53 & 44 & 38 & 19 & 10 & Place Bottle Cup & 97 & 79 & 69 & 67 & 30 \\
Pick Mixed Objects Referential A & 33 & 17 & 12 & 12 & 8 & Hang Mug Stack Blocks & 15 & 8 & 3 & 0 & 0 \\
Pick Mixed Objects Referential B & 20 & 21 & 18 & 18 & 8 & Place Object Scale Click & 19 & 36 & 56 & 6 & 5 \\
Pick Sauce Can Multi View & 47 & 48 & 43 & 42 & 42 & Place Burger Fries Click Bell & 87 & 89 & 57 & 14 & 34 \\
Pick Seals Multi View & 32 & 31 & 31 & 31 & 30 & Click Can Place Items & 78 & 46 & 47 & 14 & 18 \\
Pick Breads Multi View & 56 & 51 & 54 & 52 & 61 & Stamp Seals Press Stapler & 77 & 82 & 8 & 7 & 1 \\
Pick Mixed Objects Multi View A & 33 & 27 & 27 & 33 & 32 & Click Open Place & 50 & 22 & 37 & 5 & 11 \\
Pick Mixed Objects Multi View B & 38 & 29 & 30 & 22 & 23 & Observe Blocks Move Memory & 21 & 0 & 0 & 0 & 0 \\
Press Stapler Repeat & 79 & 41 & 42 & 36 & 45 & Observe Objects Click Memory & 97 & 35 & 3 & 0 & 0 \\
Lift Pot Repeat & 60 & 43 & 63 & 54 & 63 & Remember Color Cover & 100 & 48 & 11 & 2 & 0 \\
Lift Fan Repeat & 89 & 80 & 73 & 71 & 62 & Remember Orientation Restore & 20 & 0 & 1 & 0 & 0 \\
Click Bell Repeat & 93 & 84 & 87 & 95 & 90 & Press Stapler Memory & 32 & 3 & 0 & 0 & 0 \\
\bottomrule
\end{tabular}%
}
\end{table*}

\vspace{0.75em}

\begin{table*}[!t]

\centering
\caption{X-VLA per-task \textit{Success Rate} across difficulty levels.}
\label{tab:appendix-xvla-success}
\scriptsize
\setlength{\tabcolsep}{2.2pt}
\renewcommand{\arraystretch}{0.9}
\resizebox{\textwidth}{!}{%
\begin{tabular}{@{}lrrrrr@{\hspace{1.2em}}lrrrrr@{}}
\toprule
Task & L1 & L2 & L3 & L4 & L5 & Task & L1 & L2 & L3 & L4 & L5 \\
\midrule
Pick Blocks Size & 48 & 27 & 22 & 23 & 16 & Put Bottles Dustbin & 79 & 64 & 55 & 31 & 28 \\
Pick Blocks Height & 88 & 81 & 65 & 56 & 55 & Move Blocks Apart & 86 & 74 & 81 & 17 & 31 \\
Pick Blocks Length & 43 & 30 & 21 & 28 & 14 & Move Playing Cards Away & 97 & 96 & 92 & 87 & 83 \\
Pick Blocks Area & 52 & 27 & 25 & 28 & 16 & Place Bowls Plates & 67 & 44 & 29 & 15 & 8 \\
Pick Blocks Distance & 39 & 44 & 33 & 29 & 33 & Separate Fries Bread & 78 & 87 & 71 & 79 & 66 \\
Pick Mugs Distance & 44 & 42 & 25 & 29 & 23 & Rank Blocks Height & 91 & 95 & 84 & 72 & 46 \\
Pick Pill Bottles Distance & 34 & 42 & 42 & 34 & 64 & Rank Blocks Color & 93 & 92 & 91 & 82 & 63 \\
Pick Mixed Objects Distance A & 42 & 40 & 36 & 34 & 26 & Rank Blocks Size & 97 & 88 & 65 & 32 & 24 \\
Pick Mixed Objects Distance B & 48 & 40 & 31 & 37 & 27 & Hit Blocks Hammer Order & 84 & 36 & 16 & 7 & 4 \\
Pick Blocks Canonical & 36 & 23 & 10 & 10 & 8 & Click Objects Order & 39 & 6 & 0 & 0 & 0 \\
Pick Cups Canonical & 38 & 23 & 20 & 19 & 11 & Stamp Seals Order & 34 & 71 & 65 & 65 & 31 \\
Pick Rubik Cubes Canonical & 26 & 18 & 14 & 5 & 12 & Click Bell Clockwise Order & 57 & 15 & 1 & 0 & 0 \\
Pick Mixed Objects Canonical A & 29 & 10 & 13 & 7 & 7 & Stack Blocks Color Order & 99 & 93 & 67 & 33 & 18 \\
Pick Mixed Objects Canonical B & 30 & 15 & 15 & 13 & 6 & Stack Blocks Size Order & 78 & 87 & 29 & 2 & 1 \\
Pick Tea Box Referential & 43 & 38 & 27 & 23 & 14 & Stack Blocks Length Order & 99 & 74 & 59 & 14 & 0 \\
Pick Cans Referential & 36 & 22 & 19 & 24 & 13 & Place Phone Press Stapler & 51 & 28 & 25 & 22 & 7 \\
Pick Soaps Referential & 43 & 30 & 31 & 15 & 14 & Place Bottle Cup & 98 & 44 & 19 & 16 & 6 \\
Pick Mixed Objects Referential A & 36 & 21 & 26 & 14 & 17 & Hang Mug Stack Blocks & 23 & 5 & 3 & 2 & 0 \\
Pick Mixed Objects Referential B & 44 & 34 & 23 & 16 & 13 & Place Object Scale Click & 57 & 22 & 19 & 5 & 2 \\
Pick Sauce Can Multi View & 54 & 53 & 62 & 57 & 59 & Place Burger Fries Click Bell & 98 & 59 & 18 & 12 & 4 \\
Pick Seals Multi View & 30 & 38 & 32 & 30 & 35 & Click Can Place Items & 54 & 41 & 37 & 34 & 24 \\
Pick Breads Multi View & 37 & 37 & 33 & 41 & 34 & Stamp Seals Press Stapler & 75 & 86 & 4 & 2 & 1 \\
Pick Mixed Objects Multi View A & 41 & 28 & 30 & 28 & 22 & Click Open Place & 3 & 2 & 0 & 0 & 0 \\
Pick Mixed Objects Multi View B & 29 & 27 & 33 & 36 & 33 & Observe Blocks Move Memory & 46 & 11 & 5 & 3 & 0 \\
Press Stapler Repeat & 49 & 35 & 30 & 29 & 27 & Observe Objects Click Memory & 18 & 0 & 0 & 0 & 0 \\
Lift Pot Repeat & 0 & 0 & 0 & 0 & 0 & Remember Color Cover & 100 & 48 & 8 & 0 & 0 \\
Lift Fan Repeat & 61 & 70 & 59 & 51 & 43 & Remember Orientation Restore & 18 & 6 & 0 & 0 & 0 \\
Click Bell Repeat & 29 & 24 & 20 & 19 & 20 & Press Stapler Memory & 41 & 1 & 0 & 0 & 0 \\
\bottomrule
\end{tabular}%
}
\end{table*}
\vspace{0.75em}

%% file: EMNLP/tables/model_results_ONTA.tex
\begin{table*}[!t]
\centering
\caption{RDT per-task \textit{Object-Normalized Target Accuracy} ($\mathcal{ONTA}$) across difficulty levels. Fine-grained spatial reasoning tasks only.}
\label{tab:appendix-rdt-onta}
\scriptsize
\setlength{\tabcolsep}{2.2pt}
\renewcommand{\arraystretch}{0.9}
\resizebox{\textwidth}{!}{%
\begin{tabular}{@{}lrrrrr@{\hspace{1.2em}}lrrrrr@{}}
\toprule
Task & L1 & L2 & L3 & L4 & L5 & Task & L1 & L2 & L3 & L4 & L5 \\
\midrule
Pick Blocks Size & -94.0 & -48.5 & -29.3 & -17.5 & -18.8 & Pick Mixed Objects Canonical A & -33.5 & -15.2 & -8.6 & -11.4 & -3.6 \\
Pick Blocks Height & -44.0 & -18.5 & -6.7 & -6.3 & -5.6 & Pick Mixed Objects Canonical B & -21.5 & -12.8 & -8.6 & -5.8 & -4.7 \\
Pick Blocks Length & -88.0 & -48.5 & -32.0 & -20.0 & -16.4 & Pick Tea Box Referential & -86.0 & -47.0 & -26.7 & -12.8 & -9.1 \\
Pick Blocks Area & -94.0 & -48.5 & -32.0 & -18.8 & -16.4 & Pick Cans Referential & -96.0 & -48.5 & -24.0 & -14.0 & -5.8 \\
Pick Blocks Distance & -48.5 & -12.8 & -13.1 & -9.1 & -6.7 & Pick Soaps Referential & -84.0 & -32.0 & -21.3 & -14.0 & -4.6 \\
Pick Mugs Distance & -27.5 & -10.7 & -10.0 & -4.4 & -5.0 & Pick Mixed Objects Referential A & -96.0 & -47.0 & -24.0 & -11.6 & -8.0 \\
Pick Pill Bottles Distance & -12.5 & -13.3 & -7.5 & -4.4 & -1.5 & Pick Mixed Objects Referential B & -94.0 & -47.0 & -29.3 & -15.2 & -10.3 \\
Pick Mixed Objects Distance A & -27.5 & -12.0 & -7.5 & -12.8 & -5.0 & Pick Sauce Can Multi View & -45.5 & -26.7 & -28.0 & -29.3 & -24.0 \\
Pick Mixed Objects Distance B & -20.0 & -16.0 & -7.5 & -4.4 & -5.0 & Pick Seals Multi View & -45.5 & -26.7 & -26.7 & -28.0 & -28.0 \\
Pick Blocks Canonical & -45.5 & -14.0 & -9.7 & -11.4 & -3.6 & Pick Breads Multi View & -45.5 & -26.7 & -30.7 & -29.3 & -33.3 \\
Pick Cups Canonical & 2.5 & 1.6 & 9.7 & 6.6 & 14.9 & Pick Mixed Objects Multi View A & -48.5 & -26.7 & -30.7 & -29.3 & -28.0 \\
Pick Rubik Cubes Canonical & -39.5 & -12.8 & -7.4 & -4.6 & -1.5 & Pick Mixed Objects Multi View B & -42.5 & -32.0 & -28.0 & -29.3 & -30.7 \\
\bottomrule
\end{tabular}%
}
\end{table*}
\vspace{0.75em}

\clearpage

\begin{table*}[!t]
\centering
\caption{GO-1 per-task \textit{Object-Normalized Target Accuracy} ($\mathcal{ONTA}$) across difficulty levels. Fine-grained spatial reasoning tasks only.}
\label{tab:appendix-go1-onta}
\scriptsize
\setlength{\tabcolsep}{2.2pt}
\renewcommand{\arraystretch}{0.9}
\resizebox{\textwidth}{!}{%
\begin{tabular}{@{}lrrrrr@{\hspace{1.2em}}lrrrrr@{}}
\toprule
Task & L1 & L2 & L3 & L4 & L5 & Task & L1 & L2 & L3 & L4 & L5 \\
\midrule
Pick Blocks Size & -34.0 & -30.5 & -17.3 & -12.5 & -9.2 & Pick Mixed Objects Canonical A & -18.5 & -5.6 & -7.4 & -5.8 & -4.7 \\
Pick Blocks Height & 42.0 & 43.0 & 33.3 & 33.8 & 18.4 & Pick Mixed Objects Canonical B & -15.5 & -2.0 & -6.3 & -8.0 & -5.8 \\
Pick Blocks Length & -94.0 & -33.5 & -13.3 & -16.3 & -14.0 & Pick Tea Box Referential & -58.0 & -24.5 & -9.3 & -9.2 & -4.6 \\
Pick Blocks Area & -92.0 & -45.5 & -30.7 & -25.0 & -20.0 & Pick Cans Referential & -68.0 & -32.0 & -28.0 & -3.2 & -3.5 \\
Pick Blocks Distance & -45.5 & -17.6 & -7.4 & -8.0 & -7.8 & Pick Soaps Referential & -76.0 & -38.0 & -28.0 & -16.4 & -9.1 \\
Pick Mugs Distance & -27.5 & -14.7 & -7.5 & -10.4 & -6.2 & Pick Mixed Objects Referential A & -60.0 & -23.0 & -16.0 & -2.0 & -2.4 \\
Pick Pill Bottles Distance & 4.0 & -22.7 & -6.3 & -2.0 & 13.7 & Pick Mixed Objects Referential B & -66.0 & -27.5 & -13.3 & -3.2 & -5.8 \\
Pick Mixed Objects Distance A & -11.0 & -8.0 & -5.0 & 2.8 & -2.7 & Pick Sauce Can Multi View & -8.0 & 10.7 & 8.0 & 5.3 & 2.7 \\
Pick Mixed Objects Distance B & -18.5 & -10.7 & -7.5 & -10.4 & -7.3 & Pick Seals Multi View & -23.0 & -13.3 & -14.7 & -17.3 & -18.7 \\
Pick Blocks Canonical & -26.0 & -6.8 & -5.1 & -5.8 & -1.5 & Pick Breads Multi View & -18.5 & -17.3 & -21.3 & 2.7 & -13.3 \\
Pick Cups Canonical & -14.0 & 0.4 & 0.6 & -3.5 & -1.5 & Pick Mixed Objects Multi View A & -26.0 & -22.7 & -28.0 & -9.3 & -9.3 \\
Pick Rubik Cubes Canonical & -11.0 & 0.4 & 1.7 & 2.1 & 7.3 & Pick Mixed Objects Multi View B & -44.0 & -28.0 & -17.3 & -17.3 & -20.0 \\
\bottomrule
\end{tabular}%
}
\end{table*}

\vspace{0.75em}

\begin{table*}[!t]
\centering
\caption{$\pi_{0.5}$ per-task \textit{Object-Normalized Target Accuracy} ($\mathcal{ONTA}$) across difficulty levels. Fine-grained spatial reasoning tasks only.}
\label{tab:appendix-pi05-onta}
\scriptsize
\setlength{\tabcolsep}{2.2pt}
\renewcommand{\arraystretch}{0.9}
\resizebox{\textwidth}{!}{%
\begin{tabular}{@{}lrrrrr@{\hspace{1.2em}}lrrrrr@{}}
\toprule
Task & L1 & L2 & L3 & L4 & L5 & Task & L1 & L2 & L3 & L4 & L5 \\
\midrule
Pick Blocks Size & 16.0 & -9.5 & -6.7 & -3.8 & -4.4 & Pick Mixed Objects Canonical A & 2.5 & -2.0 & -2.9 & -0.1 & 2.9 \\
Pick Blocks Height & 34.0 & 31.0 & 30.7 & 12.5 & 8.8 & Pick Mixed Objects Canonical B & -2.0 & -2.0 & -2.9 & -1.2 & -1.5 \\
Pick Blocks Length & -56.0 & -42.5 & -18.7 & -20.0 & -16.4 & Pick Tea Box Referential & -34.0 & -3.5 & 8.0 & -5.6 & 4.4 \\
Pick Blocks Area & -52.0 & -30.5 & 0.0 & -10.0 & -8.0 & Pick Cans Referential & -46.0 & -18.5 & 4.0 & 0.4 & -3.5 \\
Pick Blocks Distance & -26.0 & -0.8 & 4.0 & -1.2 & 1.1 & Pick Soaps Referential & 6.0 & 16.0 & 17.3 & 2.8 & -1.2 \\
Pick Mugs Distance & 8.5 & 34.7 & 40.0 & 38.8 & 27.7 & Pick Mixed Objects Referential A & -34.0 & -24.5 & -17.3 & -5.6 & -3.5 \\
Pick Pill Bottles Distance & 19.0 & 14.7 & 13.8 & 19.6 & 21.8 & Pick Mixed Objects Referential B & -60.0 & -18.5 & -9.3 & 1.6 & -3.5 \\
Pick Mixed Objects Distance A & -0.5 & 16.0 & 8.8 & 14.8 & 10.2 & Pick Sauce Can Multi View & 20.5 & 30.7 & 24.0 & 22.7 & 22.7 \\
Pick Mixed Objects Distance B & 11.5 & 5.3 & 1.3 & 22.0 & -2.7 & Pick Seals Multi View & -2.0 & 8.0 & 8.0 & 8.0 & 6.7 \\
Pick Blocks Canonical & 11.5 & 6.4 & 1.7 & -1.2 & 2.9 & Pick Breads Multi View & 34.0 & 34.7 & 38.7 & 36.0 & 48.0 \\
Pick Cups Canonical & 41.5 & 24.4 & 31.4 & 33.6 & 40.0 & Pick Mixed Objects Multi View A & -0.5 & 2.7 & 2.7 & 10.7 & 9.3 \\
Pick Rubik Cubes Canonical & 26.5 & 22.0 & 23.4 & 28.0 & 38.9 & Pick Mixed Objects Multi View B & 7.0 & 5.3 & 6.7 & -4.0 & -2.7 \\
\bottomrule
\end{tabular}%
}
\end{table*}

\vspace{0.75em}
\begin{table*}[!t]

\centering
\caption{X-VLA per-task \textit{Object-Normalized Target Accuracy} ($\mathcal{ONTA}$) across difficulty levels. Fine-grained spatial reasoning tasks only.}
\label{tab:appendix-xvla-onta}
\scriptsize
\setlength{\tabcolsep}{2.2pt}
\renewcommand{\arraystretch}{0.9}
\resizebox{\textwidth}{!}{%
\begin{tabular}{@{}lrrrrr@{\hspace{1.2em}}lrrrrr@{}}
\toprule
Task & L1 & L2 & L3 & L4 & L5 & Task & L1 & L2 & L3 & L4 & L5 \\
\midrule
Pick Blocks Size & -4.0 & -9.5 & -4.0 & 3.8 & -0.8 & Pick Mixed Objects Canonical A & -6.5 & -8.0 & 0.6 & -4.6 & -1.5 \\
Pick Blocks Height & 76.0 & 71.5 & 53.3 & 45.0 & 46.0 & Pick Mixed Objects Canonical B & -5.0 & -2.0 & 2.9 & 2.1 & -2.5 \\
Pick Blocks Length & -14.0 & -5.0 & -5.3 & 10.0 & -3.2 & Pick Tea Box Referential & -14.0 & 7.0 & 2.7 & 7.6 & 3.3 \\
Pick Blocks Area & 4.0 & -9.5 & 0.0 & 10.0 & -0.8 & Pick Cans Referential & -28.0 & -17.0 & -8.0 & 8.8 & 2.1 \\
Pick Blocks Distance & 8.5 & 32.8 & 23.4 & 20.1 & 25.6 & Pick Soaps Referential & -14.0 & -5.0 & 8.0 & -2.0 & 3.3 \\
Pick Mugs Distance & 16.0 & 22.7 & 6.3 & 14.8 & 10.2 & Pick Mixed Objects Referential A & -28.0 & -18.5 & 1.3 & -3.2 & 6.6 \\
Pick Pill Bottles Distance & 1.0 & 22.7 & 27.5 & 20.8 & 58.0 & Pick Mixed Objects Referential B & -12.0 & 1.0 & -2.7 & -0.8 & 2.1 \\
Pick Mixed Objects Distance A & 13.0 & 20.0 & 20.0 & 20.8 & 13.7 & Pick Sauce Can Multi View & 31.0 & 37.3 & 49.3 & 42.7 & 45.3 \\
Pick Mixed Objects Distance B & 22.0 & 20.0 & 13.8 & 24.4 & 14.8 & Pick Seals Multi View & -5.0 & 17.3 & 9.3 & 6.7 & 13.3 \\
Pick Blocks Canonical & 4.0 & 7.6 & -2.9 & -1.2 & -0.4 & Pick Breads Multi View & 5.5 & 16.0 & 10.7 & 21.3 & 12.0 \\
Pick Cups Canonical & 7.0 & 7.6 & 8.6 & 8.9 & 2.9 & Pick Mixed Objects Multi View A & 11.5 & 4.0 & 6.7 & 4.0 & -4.0 \\
Pick Rubik Cubes Canonical & -11.0 & 1.6 & 1.7 & -6.9 & 4.0 & Pick Mixed Objects Multi View B & -6.5 & 2.7 & 10.7 & 14.7 & 10.7 \\
\bottomrule
\end{tabular}%
}
\end{table*}
\vspace{0.75em}

%% file: EMNLP/tables/model_results_PS.tex
\begin{table*}[!t]
\centering
\caption{RDT per-task step \textit{Progress Score} across difficulty levels. Step tables include long-horizon tasks only.}
\label{tab:appendix-rdt-step}
\scriptsize
\setlength{\tabcolsep}{2.2pt}
\renewcommand{\arraystretch}{0.9}
\resizebox{\textwidth}{!}{%
\begin{tabular}{@{}lrrrrr@{\hspace{1.2em}}lrrrrr@{}}
\toprule
Task & L1 & L2 & L3 & L4 & L5 & Task & L1 & L2 & L3 & L4 & L5 \\
\midrule
Press Stapler Repeat & 33.0 & 37.0 & 39.3 & 36.5 & 35.2 & Stack Blocks Color Order & 11.0 & 50.0 & 33.7 & 25.0 & 20.2 \\
Lift Pot Repeat & 76.0 & 75.5 & 75.7 & 72.0 & 80.0 & Stack Blocks Size Order & 9.0 & 0.0 & 33.7 & 25.3 & 20.4 \\
Lift Fan Repeat & 39.0 & 41.5 & 41.3 & 29.8 & 34.0 & Stack Blocks Length Order & 7.0 & 50.0 & 34.0 & 25.3 & 20.4 \\
Click Bell Repeat & 15.0 & 21.5 & 28.7 & 26.5 & 28.6 & Place Phone Press Stapler & 18.0 & 40.0 & 44.0 & 49.3 & 38.2 \\
Put Bottles Dustbin & 51.0 & 56.0 & 48.3 & 33.8 & 33.6 & Place Bottle Cup & 56.0 & 52.5 & 52.7 & 44.3 & 33.6 \\
Move Blocks Apart & 6.0 & 7.0 & 12.7 & 11.0 & 8.0 & Hang Mug Stack Blocks & 10.0 & 9.0 & 17.7 & 13.0 & 9.6 \\
Move Playing Cards Away & 8.0 & 13.0 & 8.0 & 9.0 & 10.6 & Place Object Scale Click & 39.0 & 41.5 & 41.3 & 29.8 & 34.0 \\
Place Bowls Plates & 48.0 & 37.5 & 39.0 & 32.3 & 20.6 & Place Burger Fries Click Bell & 33.0 & 23.5 & 17.7 & 15.3 & 11.4 \\
Separate Fries Bread & 24.0 & 26.0 & 33.0 & 28.0 & 35.4 & Click Can Place Items & 62.0 & 55.5 & 75.7 & 55.0 & 65.0 \\
Rank Blocks Height & 6.0 & 50.5 & 40.0 & 37.8 & 31.4 & Stamp Seals Press Stapler & 2.0 & 5.0 & 1.3 & 0.8 & 3.0 \\
Rank Blocks Color & 14.0 & 53.0 & 44.3 & 37.5 & 31.4 & Click Open Place & 0.0 & 16.5 & 10.0 & 22.8 & 17.2 \\
Rank Blocks Size & 5.0 & 50.0 & 38.3 & 33.8 & 28.0 & Observe Blocks Move Memory & 0.0 & 0.0 & 0.0 & 0.0 & 0.0 \\
Hit Blocks Hammer Order & 40.0 & 14.5 & 9.3 & 6.5 & 7.2 & Observe Objects Click Memory & 14.0 & 7.0 & 5.3 & 6.3 & 5.8 \\
Click Objects Order & 20.0 & 23.5 & 22.7 & 17.5 & 0.0 & Remember Color Cover & 47.0 & 23.5 & 15.0 & 5.5 & 6.0 \\
Stamp Seals Order & 3.0 & 3.0 & 5.7 & 6.3 & 6.6 & Remember Orientation Restore & 0.0 & 0.5 & 0.7 & 0.0 & 0.2 \\
Click Bell Clockwise Order & 5.0 & 50.0 & 38.3 & 33.8 & 28.0 & Press Stapler Memory & 30.0 & 7.0 & 4.0 & 3.3 & 5.8 \\
\bottomrule
\end{tabular}%
}
\end{table*}
\vspace{0.75em}


\begin{table*}[!t]
\centering
\caption{GO-1 per-task step \textit{Progress Score} across difficulty levels. Step tables include long-horizon tasks only.}
\label{tab:appendix-go1-step}
\scriptsize
\setlength{\tabcolsep}{2.2pt}
\renewcommand{\arraystretch}{0.9}
\resizebox{\textwidth}{!}{%
\begin{tabular}{@{}lrrrrr@{\hspace{1.2em}}lrrrrr@{}}
\toprule
Task & L1 & L2 & L3 & L4 & L5 & Task & L1 & L2 & L3 & L4 & L5 \\
\midrule
Press Stapler Repeat & 27.0 & 25.5 & 14.7 & 8.3 & 8.8 & Stack Blocks Color Order & 54.0 & 56.0 & 36.7 & 25.8 & 20.0 \\
Lift Pot Repeat & 61.0 & 54.0 & 92.0 & 92.8 & 91.6 & Stack Blocks Size Order & 38.0 & 16.5 & 5.0 & 4.3 & 4.2 \\
Lift Fan Repeat & 23.0 & 22.5 & 24.7 & 26.5 & 21.2 & Stack Blocks Length Order & 30.0 & 12.0 & 10.3 & 4.3 & 3.0 \\
Click Bell Repeat & 16.0 & 19.0 & 8.7 & 9.3 & 6.3 & Place Phone Press Stapler & 48.0 & 33.0 & 34.3 & 30.0 & 28.8 \\
Put Bottles Dustbin & 59.0 & 84.5 & 68.0 & 60.3 & 48.4 & Place Bottle Cup & 16.0 & 16.5 & 50.3 & 68.5 & 57.0 \\
Move Blocks Apart & 21.0 & 21.5 & 12.0 & 8.3 & 6.6 & Hang Mug Stack Blocks & 8.0 & 23.3 & 18.5 & 18.3 & 13.1 \\
Move Playing Cards Away & 52.0 & 43.5 & 45.0 & 20.5 & 21.0 & Place Object Scale Click & 12.0 & 19.0 & 14.3 & 6.8 & 8.2 \\
Place Bowls Plates & 37.0 & 36.0 & 35.0 & 26.8 & 20.8 & Place Burger Fries Click Bell & 37.0 & 32.5 & 13.7 & 5.0 & 6.6 \\
Separate Fries Bread & 44.0 & 37.5 & 26.7 & 27.8 & 14.6 & Click Can Place Items & 77.0 & 45.5 & 53.3 & 51.0 & 55.4 \\
Rank Blocks Height & 34.0 & 51.5 & 39.3 & 36.0 & 30.0 & Stamp Seals Press Stapler & 24.0 & 23.0 & 8.3 & 5.3 & 5.2 \\
Rank Blocks Color & 41.0 & 57.0 & 44.7 & 36.5 & 30.6 & Click Open Place & 12.0 & 13.0 & 7.0 & 0.0 & 5.8 \\
Rank Blocks Size & 41.0 & 61.0 & 47.3 & 40.5 & 34.0 & Observe Blocks Move Memory & 3.0 & 3.0 & 1.3 & 11.0 & 11.6 \\
Hit Blocks Hammer Order & 36.0 & 29.0 & 24.3 & 27.5 & 27.0 & Observe Objects Click Memory & 19.0 & 6.5 & 8.3 & 5.8 & 5.2 \\
Click Objects Order & 62.0 & 42.0 & 30.3 & 24.3 & 3.0 & Remember Color Cover & 5.0 & 0.0 & 0.0 & 0.3 & 0.0 \\
Stamp Seals Order & 12.0 & 29.5 & 24.0 & 23.3 & 21.4 & Remember Orientation Restore & 2.0 & 6.0 & 5.7 & 6.5 & 5.4 \\
Click Bell Clockwise Order & 61.0 & 52.0 & 38.3 & 30.8 & 25.2 & Press Stapler Memory & 9.0 & 6.5 & 9.3 & 9.5 & 7.6 \\
\bottomrule
\end{tabular}%
}
\end{table*}
\vspace{0.75em}

\begin{table*}[!t]

\centering
\caption{$\pi_{0.5}$ per-task step \textit{Progress Score} across difficulty levels. Step tables include long-horizon tasks only.}
\label{tab:appendix-pi05-step}
\scriptsize
\setlength{\tabcolsep}{2.2pt}
\renewcommand{\arraystretch}{0.9}
\resizebox{\textwidth}{!}{%
\begin{tabular}{@{}lrrrrr@{\hspace{1.2em}}lrrrrr@{}}
\toprule
Task & L1 & L2 & L3 & L4 & L5 & Task & L1 & L2 & L3 & L4 & L5 \\
\midrule
Press Stapler Repeat & 79.0 & 61.5 & 63.0 & 59.0 & 64.8 & Stack Blocks Color Order & 91.0 & 80.0 & 58.7 & 37.0 & 23.8 \\
Lift Pot Repeat & 60.0 & 64.5 & 77.0 & 71.8 & 74.8 & Stack Blocks Size Order & 89.0 & 75.0 & 49.3 & 36.3 & 25.0 \\
Lift Fan Repeat & 89.0 & 85.0 & 84.0 & 80.8 & 75.4 & Stack Blocks Length Order & 83.0 & 65.5 & 52.0 & 33.3 & 27.0 \\
Click Bell Repeat & 93.0 & 89.0 & 90.7 & 98.0 & 94.2 & Place Phone Press Stapler & 46.0 & 31.0 & 34.7 & 32.5 & 26.6 \\
Put Bottles Dustbin & 95.0 & 90.5 & 93.0 & 87.8 & 82.4 & Place Bottle Cup & 97.0 & 90.0 & 87.3 & 86.3 & 75.8 \\
Move Blocks Apart & 73.0 & 67.5 & 62.0 & 61.5 & 55.6 & Hang Mug Stack Blocks & 15.0 & 48.0 & 44.3 & 26.5 & 23.6 \\
Move Playing Cards Away & 81.0 & 78.5 & 77.3 & 81.5 & 77.8 & Place Object Scale Click & 19.0 & 50.5 & 56.0 & 22.8 & 30.2 \\
Place Bowls Plates & 94.0 & 91.5 & 78.0 & 69.5 & 50.4 & Place Burger Fries Click Bell & 87.0 & 90.0 & 78.0 & 61.8 & 72.0 \\
Separate Fries Bread & 92.0 & 88.5 & 83.0 & 86.8 & 74.4 & Click Can Place Items & 78.0 & 69.0 & 78.3 & 69.0 & 73.0 \\
Rank Blocks Height & 72.0 & 72.5 & 63.7 & 68.0 & 53.0 & Stamp Seals Press Stapler & 77.0 & 86.0 & 50.0 & 48.8 & 45.0 \\
Rank Blocks Color & 88.0 & 77.5 & 73.0 & 68.5 & 55.4 & Click Open Place & 50.0 & 48.5 & 55.0 & 30.8 & 37.4 \\
Rank Blocks Size & 91.0 & 84.0 & 69.7 & 57.5 & 47.8 & Observe Blocks Move Memory & 21.0 & 15.0 & 16.0 & 13.5 & 14.4 \\
Hit Blocks Hammer Order & 81.0 & 33.0 & 27.3 & 23.3 & 19.8 & Observe Objects Click Memory & 97.0 & 58.5 & 33.0 & 25.5 & 14.4 \\
Click Objects Order & 60.0 & 35.0 & 27.0 & 28.0 & 15.0 & Remember Color Cover & 100.0 & 50.5 & 20.3 & 12.0 & 7.0 \\
Stamp Seals Order & 40.0 & 50.5 & 43.0 & 43.6 & 46.6 & Remember Orientation Restore & 20.0 & 12.0 & 14.7 & 13.3 & 12.0 \\
Click Bell Clockwise Order & 98.0 & 96.5 & 52.7 & 40.8 & 38.6 & Press Stapler Memory & 32.0 & 10.0 & 9.7 & 8.8 & 9.4 \\
\bottomrule
\end{tabular}%
}
\end{table*}
\vspace{0.75em}




\begin{table*}[!t]
\centering
\caption{X-VLA per-task step \textit{Progress Score} across difficulty levels. Step tables include long-horizon tasks only.}
\label{tab:appendix-xvla-step}
\scriptsize
\setlength{\tabcolsep}{2.2pt}
\renewcommand{\arraystretch}{0.9}
\resizebox{\textwidth}{!}{%
\begin{tabular}{@{}lrrrrr@{\hspace{1.2em}}lrrrrr@{}}
\toprule
Task & L1 & L2 & L3 & L4 & L5 & Task & L1 & L2 & L3 & L4 & L5 \\
\midrule
Press Stapler Repeat & 49.0 & 42.5 & 40.3 & 38.8 & 36.6 & Stack Blocks Color Order & 99.0 & 96.5 & 79.0 & 54.5 & 37.8 \\
Lift Pot Repeat & 0.0 & 0.0 & 0.0 & 0.0 & 0.0 & Stack Blocks Size Order & 78.0 & 93.5 & 55.0 & 29.3 & 22.2 \\
Lift Fan Repeat & 61.0 & 79.5 & 72.0 & 71.8 & 66.2 & Stack Blocks Length Order & 99.0 & 87.5 & 73.7 & 41.8 & 24.2 \\
Click Bell Repeat & 29.0 & 28.5 & 27.0 & 23.3 & 24.8 & Place Phone Press Stapler & 51.0 & 45.5 & 46.7 & 43.5 & 43.8 \\
Put Bottles Dustbin & 79.0 & 80.0 & 78.3 & 71.0 & 74.0 & Place Bottle Cup & 98.0 & 71.0 & 61.3 & 49.0 & 40.0 \\
Move Blocks Apart & 86.0 & 80.5 & 89.7 & 65.0 & 71.2 & Hang Mug Stack Blocks & 23.0 & 51.0 & 51.7 & 41.3 & 31.4 \\
Move Playing Cards Away & 97.0 & 98.0 & 95.7 & 93.6 & 93.6 & Place Object Scale Click & 57.0 & 35.0 & 30.3 & 27.3 & 20.0 \\
Place Bowls Plates & 67.0 & 62.5 & 52.7 & 39.3 & 30.4 & Place Burger Fries Click Bell & 98.0 & 78.5 & 55.7 & 57.3 & 49.8 \\
Separate Fries Bread & 78.0 & 92.5 & 87.3 & 89.5 & 82.2 & Click Can Place Items & 54.0 & 64.5 & 67.7 & 69.3 & 64.2 \\
Rank Blocks Height & 91.0 & 97.5 & 93.3 & 91.0 & 83.0 & Stamp Seals Press Stapler & 75.0 & 89.0 & 61.0 & 49.3 & 42.8 \\
Rank Blocks Color & 93.0 & 96.5 & 95.0 & 92.5 & 90.0 & Click Open Place & 15.0 & 19.5 & 18.0 & 24.8 & 28.8 \\
Rank Blocks Size & 97.0 & 94.5 & 83.3 & 71.0 & 67.0 & Observe Blocks Move Memory & 46.0 & 29.5 & 27.7 & 27.3 & 19.4 \\
Hit Blocks Hammer Order & 84.0 & 57.5 & 46.7 & 40.5 & 30.8 & Observe Objects Click Memory & 18.0 & 4.5 & 5.0 & 2.8 & 2.6 \\
Click Objects Order & 39.0 & 20.0 & 10.3 & 6.5 & 0.0 & Remember Color Cover & 100.0 & 50.0 & 18.3 & 6.5 & 4.0 \\
Stamp Seals Order & 34.0 & 77.0 & 78.3 & 78.3 & 53.2 & Remember Orientation Restore & 18.0 & 19.5 & 7.0 & 6.3 & 5.6 \\
Click Bell Clockwise Order & 57.0 & 26.0 & 14.7 & 9.8 & 7.2 & Press Stapler Memory & 41.0 & 6.5 & 3.0 & 4.3 & 4.8 \\
\bottomrule
\end{tabular}%
}
\end{table*}
\vspace{0.75em}